\let\origvec\vec
\documentclass[twocolumn]{svjour3}

\let\vec\origvec
\smartqed  

\usepackage{graphicx}
\usepackage{amsmath}
\usepackage{amssymb}
\usepackage{tabularx}
\usepackage{multirow}
\usepackage{algorithm}
\usepackage{algorithmic}
\usepackage{stackengine}
\usepackage{xcolor}
\usepackage{adjustbox}
\usepackage{colortbl}
\usepackage{ragged2e}
\usepackage{rotating,soul}
\usepackage[caption=false,font=footnotesize]{subfig}
\usepackage[left=1.8cm,right=1.8cm,top=2.0cm,bottom=2.7cm]{geometry}
\usepackage[colorlinks,bookmarksnumbered,bookmarksdepth=3,bookmarksopen,citecolor=blue]{hyperref}

\usepackage{silence}
\usepackage[numbers,sort]{natbib}
\bibpunct[,]{(}{)}{;}{a}{}{,}
\renewcommand{\cite}[1]{\textcolor{blue}{\citep{#1}}}
\newcommand{\shortcite}[1]{\textcolor{blue}{(\citeyear{#1})}}

\graphicspath{{./Imgs/},{./Imgs/city/},{./Imgs/sbd/},{./Imgs/intro/},{./Imgs/feature/}}
\DeclareGraphicsExtensions{.pdf,.jpg,.png}
\newcommand{\figref}[1]{Fig.~\ref{#1}}
\newcommand{\tabref}[1]{Table~\ref{#1}}
\newcommand{\secref}[1]{Section~\ref{#1}}

\newcommand{\equref}[1]{Eq. (\ref{#1})}
\newcommand{\parahead}[1]{\noindent\textbf{#1}}
\newcommand{\myPara}[1]{\vspace{.14in}\noindent\textbf{#1}}
\def\ie{\textit{i.e.}}
\def\eg{\textit{e.g.}}

\def\etal{\textit{et al.~}}
\def\sArt{{state-of-the-art~}}
\def\conv{\textit{conv~}}

\newcommand{\revise}[1]{{\textcolor{black}{#1}}}
\newcommand{\myfbox}{\adjustbox{cfbox=white}}

\journalname{International Journal of Computer Vision}

\begin{document}

\title{Semantic Edge Detection with Diverse Deep Supervision}

\author{Yun Liu$^1$ \and
        Ming-Ming Cheng$^1$ \and
        Deng-Ping Fan$^1$ \and
        Le Zhang$^2$ \and
        Jia-Wang Bian$^3$ \and
        Dacheng Tao$^4$
}

\authorrunning{Yun Liu et al.} 

\institute{
This research was supported by the National Key Research and Development 
Program of China Grant No.2018AAA0100400 and
NSFC (NO. 61922046). \\
\rule{\linewidth}{0.25mm}
$^1$ Y. Liu, M.M. Cheng and D.P. Fan are with 
College of Computer Science, Nankai University. M.M. Cheng is the corresponding 
author (cmm@nankai.edu.cn). \\
$^2$ L. Zhang is with the University of Electronic Science and Technology 
of China. \\
$^3$ J.W. Bian is with the School of Computer Science, University of Adelaide \\
$^4$ D. Tao is with the JD Explore Academy in JD.com, China.
}

\date{Received: date / Accepted: date}

\maketitle

\begin{abstract}
Semantic edge detection (SED), which aims at jointly extracting 
edges as well as their category information, has far-reaching 
applications in domains such as semantic segmentation, object 
proposal generation, and object recognition.
SED naturally requires achieving two distinct supervision targets: 
locating fine detailed edges and identifying high-level semantics.
\revise{Our motivation comes from the hypothesis that such distinct targets
prevent \sArt SED methods from effectively using deep supervision 
to improve results.
To this end, we propose a novel fully convolutional neural 
network using diverse deep supervision 
(\textbf{DDS}) within a multi-task framework where bottom layers 
aim at generating category-agnostic edges, while top layers are 
responsible for the detection of category-aware semantic edges. 
To overcome the hypothesized supervision challenge, a novel 
information converter unit is introduced, whose effectiveness 
has been extensively evaluated on SBD and Cityscapes datasets.}

\keywords{
Semantic edge detection, diverse deep supervision, information converter}
\end{abstract}

\section{Introduction}\label{sec:introduction}
The aim of classical edge detection is to detect 
edges and object boundaries in natural images. 
It is \textbf{category-agnostic}, in that object categories 
need not be recognized.
Classical edge detection can be viewed as a pixel-wise binary 
classification problem, whose objective is to classify each pixel 
as belonging to either the class indicating an edge, 
or the class indicating a non-edge. 
In this paper, we consider more practical scenarios of 
semantic edge detection (\textbf{SED}), 
which jointly achieves edge detection and edge category 
recognition within an image.
SED \cite{hariharan2011semantic,yu2017casenet,maninis2017convolutional,bertasius2015high} 
is an active computer vision research topic due to 
its wide-ranging applications, including  
object proposal generation \cite{bertasius2015high},
occlusion and depth reasoning \cite{amer2015monocular,bian2021depth},
3D reconstruction \cite{shan2014occluding}, 
object detection \cite{ferrari2008groups,ferrari2010images},
and image-based localization \cite{ramalingam2010skyline2gps}.

In the past several years, deep convolutional neural networks 
(\textbf{DCNNs}) reign undisputed as the new de-facto method 
for category-agnostic edge detection 
\cite{xie2015holistically,xie2017holistically,liu2017richer,liu2019richer,hu2018learning}, 
where near human-level performance has been achieved. 
However, deep learning for \textbf{category-aware} SED, which jointly 
detects visually salient edges as well as recognizing their 
categories, has not yet witnessed such vast popularity. 
Hariharan \etal \shortcite{hariharan2011semantic} first combined generic 
object detectors with bottom-up edges to recognize semantic edges. 
Yang \etal \shortcite{yang2016object} proposed a fully convolutional 
encoder-decoder network to detect object contours but without 
recognizing specific categories. 
More recently, CASENet \cite{yu2017casenet} introduces a skip-layer 
structure to enrich the top-layer category-aware edge activation 
with bottom-layer features, improving previous \sArt methods 
with a significant margin.
\revise{However, CASENet imposes supervision only at the Side-5 
and final fused classification and uses feature maps from Side-1 
$\sim$ Side-3 without deep supervision.
After unsuccessfully trying various ways of adding deep supervision, 
CASENet claims that imposing deep supervision at bottom network sides 
(Side-1 $\sim$ Side-4) \textit{is unnecessary}.
This conclusion has also been widely accepted by recent SED works
\cite{yu2018simultaneous,acuna2019devil,hu2019dynamic}.}

\newcommand{\AddImg}[2]{\subfloat[#1]{\fbox{%
\includegraphics[width=.32\linewidth]{2011_003013#2}}}}
\newcommand{\AddColor}[2]{\subfloat[#1]{\myfbox{%
\includegraphics[width=.32\linewidth]{2011_003013#2}}}}
\begin{figure}[!tb]
    \centering
    \setlength{\fboxrule}{0.6pt}
    \setlength{\fboxsep}{0pt}
    \AddImg{Original image}{} \hfill
    \AddImg{Ground truth}{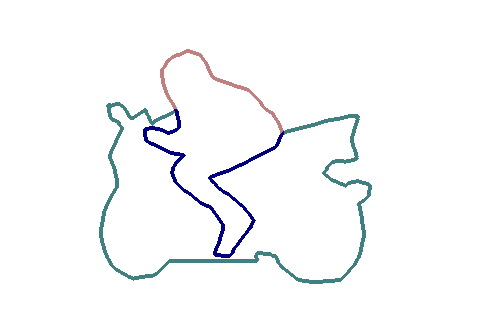} \hfill
    \AddColor{Color codes}{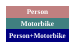}
    \\  \vspace{-0.12in}
    \AddImg{Side-1}{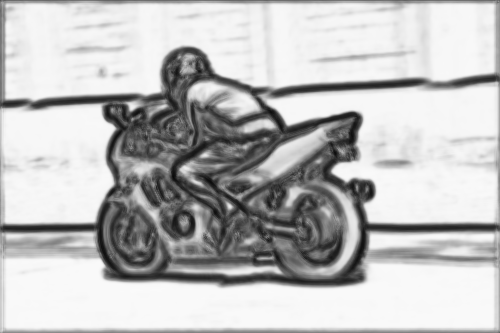} \hfill
    \AddImg{Side-2}{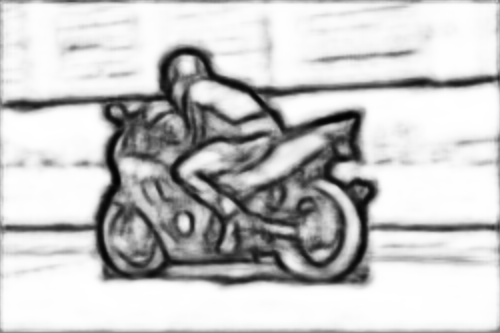} \hfill
    \AddImg{Side-3}{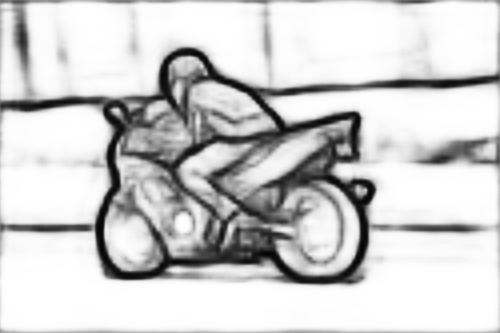}
    \\  \vspace{-0.12in}
    \AddImg{Side-4}{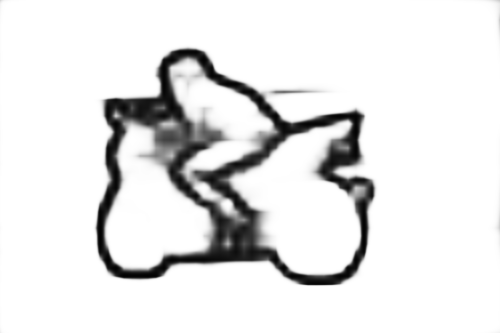} \hfill
    \AddImg{Side-5}{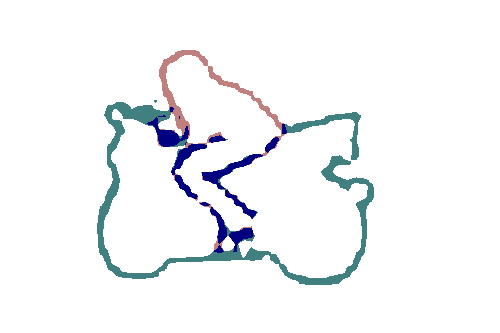} \hfill
    \AddImg{DDS}{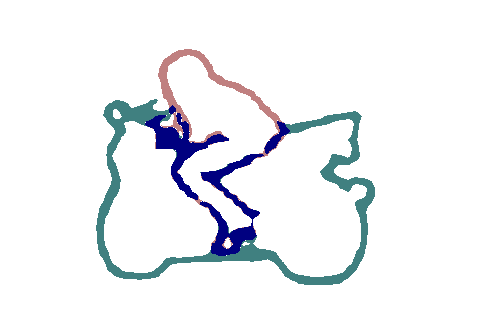}
    \\
    \caption{An example of our DDS algorithm. (a) shows the original 
    image from the SBD dataset \cite{hariharan2011semantic}. 
    (b)-(c) show its semantic edge map and corresponding color codes. 
    (d)-(g) display category-agnostic edges from Side-1 $\sim$ Side-4.
    (h)-(i) show semantic edges of Side-5 and DDS (DDS-R) output, 
    respectively.
    } \label{fig:motivation}
\end{figure}

\revise{SED naturally requires achieving two distinct supervision targets:
i) locating fine detailed edges by capturing discontinuity among 
image regions, mainly using low-level features;
and ii) identifying abstracted high-level semantics by summarizing 
different appearance variations of the target categories.
While it may be intuitive and straightforward to impose category-agnostic 
edge supervision at bottom network sides for low-level edge details 
and impose category-aware edge supervision at top sides for semantic 
classification, directly doing this as in CASENet \cite{yu2017casenet} 
even degrades the performance compared with directly learning semantic edges 
without deep supervision or category-agnostic edge guidance.
We hypothesize that distinct supervision targets prevent \sArt SED 
methods \cite{yu2017casenet,yu2018simultaneous,acuna2019devil,hu2019dynamic}
from successfully applying deep supervision \cite{lee2015deeply}.
Specifically, we observe that the success stories of deep supervision, 
including image categorization~\cite{szegedy2015going}, 
object detection~\cite{lin2020focal},
visual tracking~\cite{wang2015visual},
and category-agnostic edge detection \cite{xie2017holistically,liu2017richer},
usually adopt the same type of supervision for all network sides.
In contrast, CASENet \textit{directly} imposes \textit{distinct} supervision 
targets to bottom and top network sides.
Therefore, we consider achieving such distinct supervision using some buffers,
\ie, in an \textit{indirect} manner, to prevent the backbone network from
being directly influenced by distinct targets.}

\begin{figure*}[!tbh]
  \centering
  \includegraphics[width=1\linewidth]{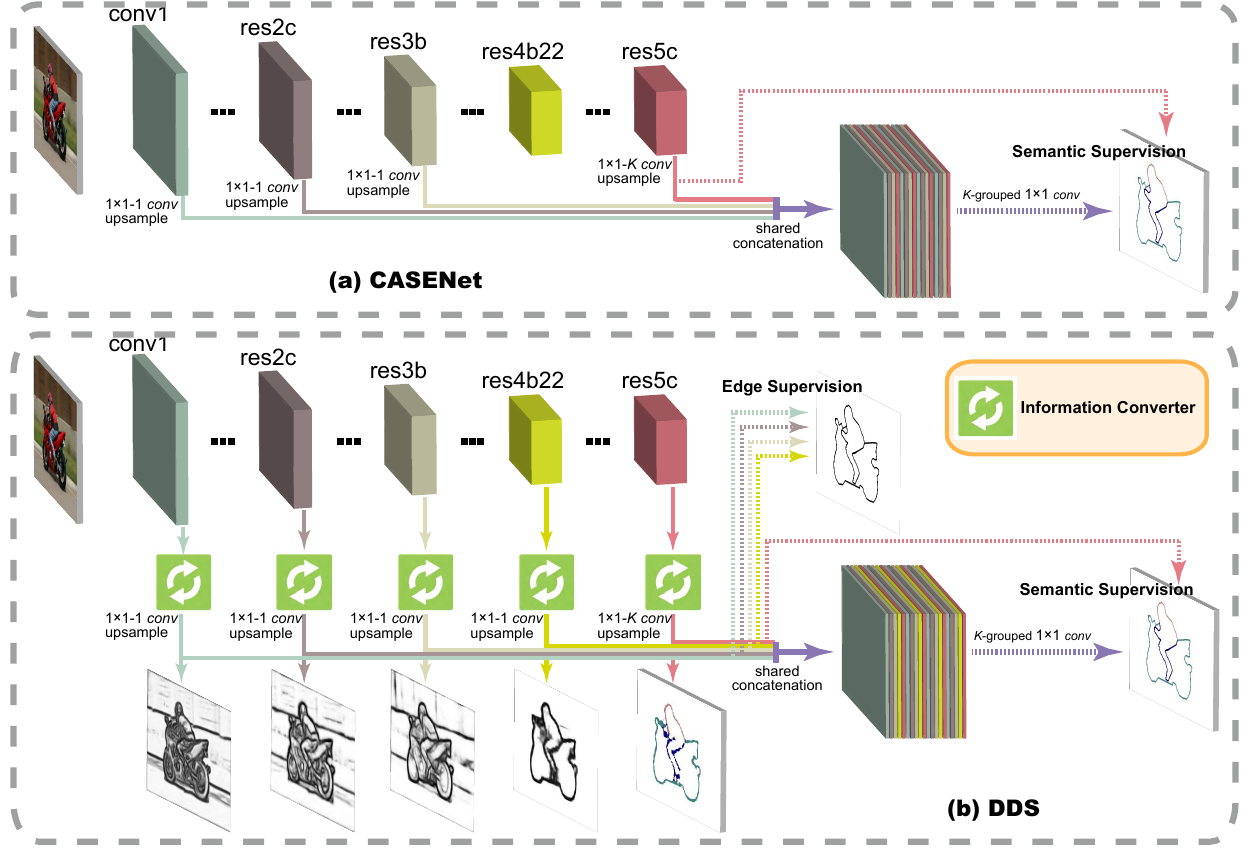} \\
  \caption{A comparison between two SED models:
    CASENet \cite{yu2017casenet} and our DDS.
    CASENet only adds top supervision on the Side-5 activation, 
    and the authors claimed that deep supervision was not necessary
    in their architecture. 
    \revise{However, our proposed DDS network adds deep supervision
    at all network sides. 
    Note that \textit{information converters} are crucial
    for resolving the distinct supervision targets of 
    category-agnostic and category-aware edges.}
  }\label{fig:netArchitecture}
\end{figure*}

\revise{In this paper, we propose a \textbf{diverse deep supervision} 
(\textbf{DDS}) method, which employs deep supervision with 
different loss functions for high-level and low-level feature 
learning, as shown in \figref{fig:netArchitecture}(b).
To this end, we propose an \textbf{\textit{information converter}} 
unit to change the backbone DCNN features into different representations,
for training category-agnostic or semantic edges, respectively.
Hence, \textit{information converters} act as buffers, 
making distinct supervision targets indirectly affect top and bottom
convolution (\ie, \textbf{\textit{conv}} layers.
The \textit{existence} of \textit{information converters} separates 
the information content in \conv layers by assigning unique sets of 
parameters and imposing separate losses to each network side.
This makes a single backbone network be effectively trained 
end-to-end towards different targets.}
An example of DDS is shown in \figref{fig:motivation}.
The bottom sides of the neural network help Side-5 to find fine 
details, thus making the final fused semantic edges 
(\figref{fig:motivation}(i)) smoother than those coming 
from Side-5 (\figref{fig:motivation}(h)).

\revise{In summary, our main contributions include:}
\begin{itemize}
\item \revise{We analyze the reason why \sArt SED methods cannot apply 
deep supervision to improve results, \ie, 
due to the distinct supervision targets in SED (\secref{sec:superv}).}
\item \revise{We propose a new SED method, called diverse deep supervision 
(DDS), which uses \textit{information converters} to separate
the information content in backbone \conv layers and thus achieve 
distinct supervision in an indirect manner (\secref{sec:approach}).}
\item \revise{We provide detailed ablation studies to further understand 
the proposed method (\secref{sec:ablation_study}).}
\end{itemize}
\revise{We extensively evaluate DDS on SBD \cite{hariharan2011semantic}
and Cityscapes \cite{cordts2016cityscapes} datasets.
DDS achieves \sArt performance, demonstrating the reasonability 
of our analyses and thus opening up a new path for future SED research.}

\section{Related Work}
An exhaustive review of the abundant literature on this topic is 
out of the scope of this paper. 
Instead, we first summarize the most important threads of research
to solve the problem of classical category-agnostic edge detection,
followed by the discussions of deep learning-based approaches, 
semantic edge detection (SED), and the technique of deep supervision.

\myPara{Classical category-agnostic edge detection.}
Edge detection is conventionally solved by designing various
filters (\eg, Sobel \cite{sobel1970camera} and 
Canny \cite{canny1986computational}) or complex models \cite{mafi2018robust,shui2017anti} to detect pixels with highest 
gradients in their local neighborhoods \cite{trahanias1993color,hardie1995gradient,henstock1996automatic}.
To the best of our knowledge, Konishi \etal 
\shortcite{konishi2003statistical} proposed the first data-driven edge 
detector in which, unlike previous model based approaches, 
edge detection was posed as statistical inferences.
\textit{Pb} features consisting of brightness, color and texture
are used in ~\cite{martin2004learning} to obtain the posterior
probability of each boundary point.
\textit{Pb} is further extended to \textit{gPb} \cite{arbelaez2011contour}
by computing local cues from multi-scale and globalizing them through
spectral clustering.
Sketch tokens are learned from hand-drawn sketches for contour detection~\cite{lim2013sketch}, while
random decision forests are employed in \cite{dollar2015fast} to learn
the local structure of edge patches, delivering competitive results among 
non-deep-learning approaches.

\myPara{Deep category-agnostic edge detection.}
The number of success stories of machine learning has seen an 
all-time rise across many computer vision tasks recently.
The unifying idea is deep learning which utilizes neural networks
with many hidden layers aimed at learning complex feature 
representations from raw data \cite{chan2015pcanet,tang2017learning,liu2018deep}.
Motivated by this, deep learning based methods have made vast
inroads into edge detection as well 
\cite{wang2019deep,deng2018learning,yang2017deep}.
Ganin \etal \shortcite{ganin2014n} applied deep neural network 
for edge detection using a dictionary learning and nearest 
neighbor algorithm.
DeepEdge \cite{bertasius2015deepedge} first extracts candidate 
contour points and then classifies these candidates.
HFL \cite{bertasius2015high} uses SE \cite{dollar2015fast} to 
generate candidate edge points in contrast to 
Canny \cite{canny1986computational} used in DeepEdge.
Compared with DeepEdge which has to process input patches for 
every candidate point, HFL turns out to be more computationally 
feasible as the input image is only fed into the network once.
DeepContour \cite{shen2015deepcontour} partitions edge data into
subclasses and fits each subclass using different model parameters.
Xie \etal \shortcite{xie2015holistically,xie2017holistically} 
leveraged deeply-supervised nets to build a fully convolutional 
network for image-to-image prediction.
Their deep model, known as HED, fuses the information from the 
bottom and top \conv layers.
Kokkinos \revise{\shortcite{kokkinos2016pushing}} proposed some training strategies
to retrain HED.
Liu \etal \shortcite{liu2017richer,liu2019richer} introduced the 
first real-time edge detector, which achieves higher F-measure 
scores than average human annotators on the popular BSDS500 
dataset \cite{arbelaez2011contour}.

\myPara{Semantic edge detection.}
By virtue of their strong capacity for semantic representation 
learning, DCNNs based edge detectors tend to generate 
high responses at object boundary locations, 
\eg, \figref{fig:motivation} (d)-(g).
This has inspired research on simultaneously detecting edge 
pixels and classifying them based on associations with one 
or more object categories.
This so-called ``category-aware'' edge detection is highly 
beneficial to a wide range of vision tasks including object 
recognition, stereo vision, semantic segmentation, 
and object proposal generation.

Hariharan \etal \shortcite{hariharan2011semantic}
proposed the first principled way of combining generic object detectors 
with bottom-up contours to detect semantic edges.
Yang \etal \shortcite{yang2016object} proposed a fully convolutional
encoder-decoder network for object contour detection.
HFL \cite{bertasius2015high} produces category-agnostic binary edges 
and assigns class labels to all boundary points 
using deep semantic segmentation networks.
Maninis \etal \shortcite{maninis2017convolutional} coupled their 
convolutional oriented boundaries (COB) with semantic segmentation 
generated by dilated convolutions \cite{yu2015multi} to obtain 
semantic edges.
A weakly supervised learning strategy is introduced
in \cite{khoreva2016weakly}, where bounding box annotations 
alone are sufficient to produce high-quality object boundaries 
without any object-specific annotations.
\revise{Gated-SCNN \cite{takikawa2019gated} converts the semantic edge 
representation from different ResNet layers to a representation suitable 
for segmentation, improving semantic segmentation substantially.}

Yu \etal \shortcite{yu2017casenet} proposed a novel network, CASENet,
which has pushed SED performance to a new state-of-the-art.
In their architecture, low-level features are only used to augment
top classifications.
After several failed experiments, they reported that 
\revise{imposing deep supervision at bottom sides} 
\textit{is unnecessary} for SED.
More recently, Yu \etal \shortcite{yu2018simultaneous} introduced a new 
training approach, SEAL, to train CASENet \cite{yu2017casenet}. 
This approach can simultaneously align ground-truth edges and 
learn semantic edge detectors. 
\revise{However, the training of SEAL is very time-consuming due to the heavy 
CPU computation load.
For example, it needs over 16 days to train CASENet on the SBD 
dataset \cite{hariharan2011semantic}, despite that we have used a powerful 
CPU (Intel Xeon(R) CPU E5-2683 v3 @ 2.00GHz $\times$ 56).}
Hu \etal \shortcite{hu2019dynamic} proposed a novel dynamic 
feature fusion (DFF) strategy to assign different fusion weights for 
different input images and locations adptively in the fusion of 
multi-scale DCNN features.
Acuna \etal \shortcite{acuna2019devil} focused on semantic thinning 
edge alignment learning (STEAL).
They presented a simple new layer and loss to train CASENet
\cite{yu2017casenet}, so that they can learn sharp and precise 
semantic boundaries.
However, all above methods give up applying deep supervision to 
bottom layers due to the distinct supervision targets in SED.
In this work, we aim to solve this problem, 
so our method is compatible with previous methods,
including SEAL \cite{yu2018simultaneous}, DFF \cite{hu2019dynamic},
and STEAL \cite{acuna2019devil}.

\myPara{Deep supervision.}
Deep supervision has been demonstrated to be effective in many
vision and learning tasks such as
image classification~\cite{lee2015deeply,szegedy2015going},
object detection~\cite{lin2020focal,lin2017feature,liu2016ssd},
visual tracking~\cite{wang2015visual}, category-agnostic 
edge detection~\cite{xie2017holistically,liu2017richer},
salient object detection~\cite{hou2019deeply}, and so on.
Theoretically, the bottom layers of deep networks can learn 
discriminative features so that classification/regression 
at top layers is easier.  
In practice, one can explicitly influence the hidden layer 
weight/filter update process to favor highly discriminative 
feature maps using deep supervision.
\revise{However, traditional deep supervision usually adopts the same type
of supervision at all layers, so it may be suboptimal for SED to directly
apply distinct supervision of category-agnostic and category-aware edges
to bottom and top network sides, respectively.
In the following sections, we will first analyze the problem of distinct 
supervision targets of SED and then introduce a new semantic edge detector 
with successful diverse deep supervision.}

\section{\revise{Distinct Supervision Targets in SED}}\label{sec:superv}
\revise{Before expounding the proposed method, we first analyze
the problem caused by the distinct supervision targets of SED.}

\subsection{A Typical Deep Model for SED}
To introduce previous attempts for using deep supervision in SED, 
without loss of generality, we take a typical deep model 
as an example, \ie, CASENet \cite{yu2017casenet}.
As shown in \figref{fig:netArchitecture}(a), this typical model is 
built on the well-known backbone network of ResNet \cite{he2016deep}.
It connects a $1 \times 1$ \conv layer after each of Side-1 
$\sim$ Side-3 to produce a single-channel feature map $F^{(m)}$. 
The top Side-5 is connected to a $1 \times 1$ \conv layer to 
output $K$-channel class activation map
$A^{(5)}=\{A_1^{(5)},A_2^{(5)},\cdots,A_K^{(5)}\}$, 
where $K$ is the number of categories.
Then, the \textit{shared concatenation} replicates bottom features 
$F^{(m)}$ to separately concatenate each channel of the class 
activation map:
\begin{equation}
F^f = \{F^{(1)},F^{(2)},F^{(3)},A_1^{(5)},\cdots,
F^{(1)},F^{(2)},F^{(3)},A_K^{(5)}\}.
\end{equation}
Next, a $K$-grouped $1 \times 1$ \conv is performed on $F^f$ to 
generate a semantic edge map with $K$ channels, in which the 
$k$-th channel represents the edge map for the $k$-th category.
Other SED models \cite{yu2018simultaneous,hu2019dynamic} 
have similar network designs.

\subsection{Discussion}
Previous SED models 
\cite{yu2017casenet,yu2018simultaneous,bertasius2015high,hu2019dynamic}
only impose supervision on Side-5 and the final fused activation.
In CASENet, the authors have tried several deeply supervised architectures.
They first separately used all of Side-1 $\sim$ Side-5 for SED, 
with each side connected with a semantic classification loss.
The evaluation results are even worse than the basic architecture 
that directly applies $1 \times 1$ convolution at Side-5 to 
obtain semantic edges.
It is widely accepted that the bottom layers of DCNNs
contain low-level, less-semantic features such as local edges, 
which are less effective for semantic classification because 
semantic category recognition needs abstracted high-level 
features that mainly appear in the top layers of neural networks.
Thus, they would obtain poor classification results at bottom sides.
Unsurprisingly, simply connecting each low-level feature layer and 
high-level feature layer with a classification loss and deep 
supervision for SED results in a clear performance drop.

Yu \etal \shortcite{yu2017casenet} also attempted to impose deep 
supervision of binary edges at Side-1 $\sim$ Side-3 in CASENet 
but observed divergence in the semantic classification at Side-5.
\revise{Here, we provide an intuitive and reasonable explanation 
for this phenomenon.}
With the top supervision of semantic edges, the top layers of 
the network will be supervised to learn abstracted high-level 
semantics that can summarize different appearance variations 
of object categories.
Since bottom layers are the bases of top layers for 
the representation power of DCNNs, bottom layers will 
be supervised to serve top layers for 
obtaining high-level semantics through back propagation.
Conversely, with bottom supervision of category-agnostic 
edges, bottom layers are taught to focus on distinction 
between edges and non-edges, rather than visual representations 
for semantic classification.
\revise{Hence, bottom layers have two conflict supervision targets.
Compared to traditional deep supervision applications that 
usually adopt the same type of supervision, we believe that
such distinct supervision targets of SED lead to the failure 
of previous attempts to apply deep supervision for SED.
Our motivation of this work comes from this hypothesis
by trying to resolve such distinct supervision targets.}

Note that Side-4 is not used in CASENet. 
\revise{We think that} it is a naive way to \textbf{alleviate} the supervision 
conflicts by regarding the whole \textit{res4} block as a buffer 
unit between bottom and top sides.
Indeed, when adding Side-4 to CASENet (see \secref{sec:ablation_study}),
the new model (\textit{CASENet +S4}) achieves a 70.9\% mean 
F-measure, compared to 71.4\% of original CASENet.
\revise{This suggests that our hypothesis about the buffer function 
of \textit{res4} block may be reasonable.
Moreover, the classical $1 \times 1$ \conv layer after each side
\cite{xie2017holistically,yu2017casenet} is too weak to buffer 
the conflicts.
We therefore propose an information converter unit to try to 
separate the information content in the backbone layers 
by assigning unique sets of parameters and imposing separate losses 
to each network side.
In this way, we tackle the distinct supervision targets of SED
in an indirect manner, rather than the previous direct manner.}

\section{\revise{Methodology}} \label{sec:approach}
Intuitively, by employing different but ``appropriate'' 
ground truths for bottom and top sides, the learned 
intermediate representations of the different levels may 
contain complementary information.
However, directly imposing deep supervision does not seem 
to be beneficial.
In this section, we propose a new network architecture for the 
complementary learning of bottom and top sides for SED.

\subsection{\revise{Diverse Deep Supervision}}
Based on the above discussion, we hypothesize that the bottom sides 
of neural networks may not be directly beneficial to SED. 
However, we still believe that bottom sides encode
fine details complementary to the top side (Side-5). 
\revise{With appropriate architecture re-design, maybe they 
can be used for category-agnostic edge detection to improve the 
localization accuracy of semantic edges generated by the top side. 
To this end, we design a novel \textit{information converter} 
to assist low-level feature learning, making it consistent 
with high-level feature learning.
This is essential as this enables bottom layers to learn find-grained
details and serve top layers to favor highly discriminative features
simultaneously.}

Our proposed network architecture is presented in 
\figref{fig:netArchitecture}(b).
We follow CASENet to use ResNet \cite{he2016deep} as our 
backbone network.
After each \textit{information converter} 
(\secref{sec:converter}) in Side-1 $\sim$ Side-4, we connect 
a $1 \times 1$ \conv layer with a single output channel to 
produce an edge response map.
These predicted maps are then upsampled to the original image 
size using bilinear interpolation.
These side-outputs are supervised by binary category-agnostic edges.
We perform $K$-channel $1 \times 1$ convolution on Side-5 to 
obtain semantic edges, where each channel represents the binary 
edge map of one category. 
We adopt the same upsampling operation as for Side-1 $\sim$ Side-4.
Semantic edges are used to supervise the training of Side-5.

We denote the produced binary edge maps from Side-1 $\sim$ 
Side-4 as $E=\{E^{(1)},E^{(2)},E^{(3)},E^{(4)}\}$.
The semantic edge map from Side-5 is still represented by $A^{(5)}$.
A shared concatenation is then performed to obtain the stacked edge
activation map:
\begin{equation} \label{equ:fuse_map}
E^f = \{E,A_1^{(5)},E,A_2^{(5)},E,A_3^{(5)},\cdots,E,A_K^{(5)}\}.
\end{equation}
Note that $E^f$ is a stacked edge activation map, while $F^f$ 
in CASENet is a stacked feature map.
Finally, we apply $K$-grouped $1 \times 1$ convolution on $E^f$ 
to generate the fused semantic edges.
The fused edges are supervised by the ground truth of 
the semantic edges.
As shown in HED \cite{xie2017holistically}, the $1 \times 1$ 
convolution can fuse the edges from bottom and top sides well.

\subsection{Information Converter} \label{sec:converter}
\begin{figure}[!tb]
    \centering
    \includegraphics[width=.85\linewidth]{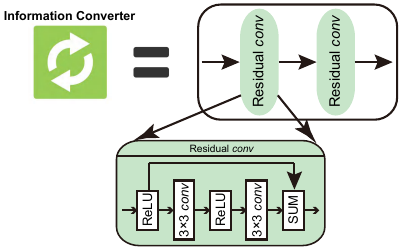} \\
    \caption{Schematic of our \textit{information converter} 
    unit (illustrated in the orange box in 
    \figref{fig:netArchitecture}).}
    \label{fig:converter}
\end{figure}

From the above analyses, the core for improving SED is the existence
of the \textit{information converter}.
In this paper, we try a simple design for \textit{information converter} 
to validate our hypothesis.
Recently, residual networks have been proved to be easier 
to optimize than plain networks \cite{he2016deep}. 
The residual learning operation is embodied by a shortcut 
connection and element-wise addition. 
We describe a residual \conv block in \figref{fig:converter},
which consists of two alternatively connected ReLU and 
\conv layers, and the output of the first ReLU layer is added 
to the output of the last \conv layer.
Our proposed \textit{information converter} combines two residual 
modules and is connected to each side of the DDS network to 
transform the learned representation into the proper form. 
This operation is expected to avoid the conflicts caused by the 
discrepancy in different losses.

\revise{The top supervision of semantic edges will guide top layers 
in learning semantic features, while the bottom supervision 
of category-agnostic edges will guide bottom layers 
in learning category-agnostic features.
Hence, bottom layers would have two distinct supervision 
through back propagation if the distinct supervision 
is directly imposed as discussed in \secref{sec:superv}.
Our information converters can separate the information content 
in the backbone layers by assigning unique sets of parameters 
and imposing separate losses to each network side,
playing a buffering role.
In this way, the distinct supervision targets are imposed 
to the backbone network in an indirect manner, rather than
the previous direct manner
\cite{yu2017casenet,yu2018simultaneous,bertasius2015high,hu2019dynamic}.}
Note that this paper mainly claims the importance of the
existence of the \textit{information converter}, not its specific 
format, so we only adopt a simple design.
In the experimental part, we will demonstrate different designs
for the \textit{information converter} achieve similar performance.

Our proposed network can successfully combine the fine details from
bottom sides and the semantic information from top sides.
Our experimental results demonstrate that this method solves the
problem of conflicts caused by diverse deep supervision.
Unlike CASENet, our semantic classification at Side-5 can be well
optimized without any divergence.
The produced binary edges from bottom sides help Side-5 
make up fine details.
Thus, the final fused semantic edges can achieve better 
localization quality.

We use binary edges of single pixel width to supervise 
Side-1 $\sim$ Side-4 and thick semantic boundaries to 
supervise Side-5 and the final fused edges.
One pixel is viewed as a binary edge if it belongs to the semantic 
boundaries of any category.
We obtain thick semantic boundaries by seeking the difference between 
a pixel and its neighbors in ground-truth semantic segmentation,
as in CASENet \cite{yu2017casenet}.
A pixel with label $k$ is regarded as a boundary of class $k$ 
if at least one neighbor with a label $k'$ ($k'\neq k$) exists.

\subsection{Multi-task Loss}
Two different loss functions, which represent category-agnostic 
and semantic edge detection losses, respectively, 
are employed in our multi-task learning framework. 
We denote all layer parameters in the network as $W$.
Suppose an image $I$ has a corresponding binary edge map
$Y=\{y_i: i=1,2,\cdots,|I|\}$.
The reweighted \textit{sigmoid} cross-entropy loss function for 
Side-1 $\sim$ Side-4 can be formulated as 
\begin{equation}
\begin{aligned}
L_{side}^{(m)}(W) =& -\sum_{i \in I} [\beta \cdot (1-y_i)
\cdot log(1-P(E_i^{(m)};W))\\
& + (1-\beta) \cdot y_i \cdot log(P(E_i^{(m)};W))], \\
&\hspace{1.09in} (m=1,\cdots,4),
\end{aligned}
\end{equation}
where we have $\beta=|Y^+|/|Y|$ and $1-\beta=|Y^-|/|Y|$.
$Y^+$ and $Y^-$ represent edge and non-edge ground-truth label
sets, respectively.
$E_i^{(m)}$ is the produced activation value at pixel $i$ 
for the $m$-th side.
$P(\cdot)$ is the standard \textit{sigmoid} function.

For an image $I$, suppose the semantic ground-truth label is
$\{\bar{Y}^1,\bar{Y}^2,\cdots,\bar{Y}^K\}$, in which
$\bar{Y}^k=\{\bar{y}^k_i: i=1,2,\cdots,|I|\}$ is the binary edge
map for the $k$-th category.
Note that each pixel may belong to the boundaries of 
multiple categories.
We define the reweighted multi-label loss for Side-5 as 
\begin{equation}
\begin{aligned}
L_{side}^{(5)}(W) =& -\sum_k\sum_{i \in I}[\beta \cdot 
(1-\bar{y}^k_i) \cdot log(1-P(A_{k,i}^{(5)};W)) \\
&+(1-\beta) \cdot \bar{y}^k_i \cdot log(P(A_{k,i}^{(5)};W))],
\label{equ:multi_label_loss_side5}
\end{aligned}
\end{equation}
in which $A_{k,i}^{(5)}$ is the Side-5's activation value for 
the $k$-th category at pixel $i$.
The loss of the fused semantic activation map is denoted 
as $L_{fuse}(W)$, which can be similarly defined as 
\begin{equation}
\begin{aligned}
L_{fuse}(W) =& -\sum_k\sum_{i \in I}[\beta \cdot 
(1-\bar{y}^k_i) \cdot log(1-P(A_{k,i}^{f};W)) \\
&+(1-\beta) \cdot \bar{y}^k_i \cdot log(P(A_{k,i}^{f};W))],
\label{equ:multi_label_loss_fuse}
\end{aligned}
\end{equation}
where $A^{f}$ is the final fused semantic edge map.
The total loss is formulated as
\begin{equation}
L(W) = \sum_{m=1,\cdots,5} L_{side}^{(m)}(W) + L_{fuse}(W).
\label{equ:overallloss}
\end{equation}
Using this total loss function, we can optimize all parameters
in an end-to-end way.
We denote DDS trained using the reweighted loss $L(W)$ 
as \textbf{DDS-R}.

Recently, Yu \etal \shortcite{yu2018simultaneous} proposed to 
simultaneously align and learn semantic edges. 
They found that the unweighted (regular) \textit{sigmoid} 
cross-entropy loss performed better than reweighted loss 
with their alignment training strategy. 
Due to the heavy computational load on the CPU, their approach 
was very time-consuming (over 16 days for SBD dataset 
\cite{hariharan2011semantic} with 28 CPU kernels and an NVIDIA 
TITAN Xp GPU) to train a network.
We use their method (SEAL) to align ground-truth 
edges only once prior to training and apply unweighted 
\textit{sigmoid} cross-entropy loss to train the aligned edges. 
The loss function for Side-1 $\sim$ Side-4 can thus be formulated as 
\begin{equation} 
\begin{aligned}
{L'}_{side}^{(m)}(W) =& -\sum_{i \in I} [(1-y_i) \cdot 
log(1-P(E_i^{(m)};W)) \\
& + y_i \cdot log(P(E_i^{(m)};W))],\\
&\hspace{1.in}(m=1,\cdots,4).
\end{aligned}
\end{equation}
The unweighted multi-label loss for Side-5 is  
\begin{equation} 
\begin{aligned}
{L'}_{side}^{(5)}(W) =& -\sum_k\sum_{i \in I}[(1-\bar{y}^k_i) \cdot 
log(1-P(A_{k,i}^{(5)};W)) \\ 
&+\bar{y}^k_i \cdot log(P(A_{k,i}^{(5)};W))].
\end{aligned}
\end{equation}
${L'}_{fuse}(W)$ can be similarly defined as
\begin{equation}
\begin{aligned}
L'_{fuse}(W) =& -\sum_k\sum_{i \in I}[
(1-\bar{y}^k_i) \cdot log(1-P(A_{k,i}^{f};W)) \\
&+\bar{y}^k_i \cdot log(P(A_{k,i}^{f};W))].
\end{aligned}
\end{equation}
The total loss is the sum across all sides: 
\begin{equation} \label{equ:overallloss2}
{L'}(W) = \sum_{m=1,\cdots,5} {L'}_{side}^{(m)}(W) + {L'}_{fuse}(W).
\end{equation}
We denote DDS trained using the unweighted loss $L'(W)$ 
as \textbf{DDS-U}.

\subsection{Implementation Details}
\revise{We implement our method using the well-known deep learning 
framework of Caffe \cite{jia2014caffe}.
The proposed network is built on ResNet \cite{he2016deep}.
We follow CASENet \cite{yu2017casenet} to change the strides of 
the first and fifth convolution blocks from 2 to 1,
so that the output scales of five convolution blocks are 
$1$, $1/2$, $1/4$, $1/8$, and $1/8$ compared to the input image, respectively.
The \textit{atrous} algorithm is used to keep the receptive field 
sizes the same as original ResNet.
Specifically, from the second convolution block to the fourth,
we use dilated convolutions with a dilation rate of 2;
for the fifth block, we use a dilation rate of 4.}
We also follow CASENet to pre-train the convolution blocks on 
the COCO dataset \cite{lin2014microsoft}.

The network is optimized with stochastic gradient descent (SGD).
Each SGD iteration chooses 10 images at uniformly random
and crops a $352 \times 352$ patch from each of them.
The weight decay and momentum are set to 0.0005 and 0.9, respectively.
We use the learning rate policy of ``poly", where the current 
learning rate equals the base one multiplying 
$(1-curr\_iter/max\_iter)^{power}$.
The parameter of $power$ is set to 0.9.
We run 25k/80k iterations ($max\_iter$) of SGD for 
SBD \cite{hariharan2011semantic} and Cityscapes 
\cite{cordts2016cityscapes}, respectively.
For DDS-R training, the base learning rate is set to 
5e-7/2.5e-7 for SBD and Cityscapes, respectively.
For DDS-U training, the loss at the beginning of training 
is very large.
Therefore, for both SBD and Cityscapes, we first pre-train the 
network with a fixed learning rate of 1e-8 for 3k iterations 
and then use the base learning rate of 1e-7 to continue training 
with the same settings as described above.
The upsampling operation is implemented with deconvolution 
layers by fixing the parameters to perform bilinear interpolation.
All experiments are performed using an NVIDIA TITAN Xp GPU.

\definecolor{mycolor}{rgb}{.9,.9,.9}

\begin{table*}[!tbh]
\centering
\renewcommand{\tabcolsep}{0.8mm}
\caption{\revise{ODS F-measure (\%) of DDS-R/DDS-U and ablation methods 
    on the SBD dataset \cite{hariharan2011semantic} using the 
    original benchmark protocol in \cite{hariharan2011semantic}.
    The best performance of each column is highlighted in \textbf{bold}.}}
\label{tab:ablation_study}
\small
\resizebox{\linewidth}{!}{
\begin{tabular}{l||c|c|c|c|c|c|c|c|c|c|c|c|c|c|c|c|c|c|c|c||c} 
    \hline\rowcolor{mycolor}
	Methods & aer. & bike & bird & boat & bot. & bus & car 
	& cat & cha. & cow & tab. & dog & hor. & mot. & per. 
	& pot. & she. & sofa & train & tv & mean \\ \hline\hline
    Softmax & 74.0 & 64.1 & 64.8 & 52.5 & 52.1 & 73.2 & 68.1 
    & 73.2 & 43.1 & 56.2 & 37.3 & 67.4 & 68.4 & 67.6 & 76.7 
    & 42.7 & 64.3 & 37.5 & 64.6 & 56.3 & 60.2 \\
    Basic & 82.5 & 74.2 & 80.2 & 62.3 & 68.0 & 80.8 & 74.3
    & 82.9 & 52.9 & 73.1 & 46.1 & 79.6 & 78.9 & 76.0
    & 80.4 & 52.4 & 75.4 & 48.6 & 75.8 & 68.0 & 70.6 \\
    DSN & 81.6 & 75.6 & 78.4 & 61.3 & 67.6 & 82.3 & 74.6 
    & 82.6 & 52.4 & 71.9 & 45.9 & 79.2 & 78.3 & 76.2 & 80.1 
    & 51.9 & 74.9 & 48.0 & 76.5 & 66.8 & 70.3 \\
    CASENet+S4 & 84.1 & 76.4 & 80.7 & 63.7 & 70.3 & 81.3 
    & 73.4 & 79.4 & 56.9 & 70.7 & 47.6 & 77.5 & 81.0 & 74.5 
    & 79.9 & 54.5 & 74.8 & 48.3 & 72.6 & 69.4 & 70.9 \\
    DDS$\backslash$Convt & 83.3 & 77.1 & 81.7 & 63.6 & 70.6 
    & 81.2 & 73.9 & 79.5 & 56.8 & 71.9 & 48.0 & 78.3 & 81.2 
    & 75.2 & 79.7 & 54.3 & 76.8 & 48.9 & 75.1 & 68.7 & 71.3 \\
    DDS$\backslash$Convt$^\dag$ & 83.6 & 75.4 & 78.9 & 59.9
    & 69.7 & 79.7 & 71.9 & 77.2 & 54.7 & 72.0 & 42.8 & 75.5 
    & 77.1 & 71.9 & 79.1 & 53.4 & 76.4 & 46.9 & 72.6 & 66.9
    & 69.3 \\
    DDS$\backslash$DeSup & 82.5 & 77.4 & 81.5 & 62.4 & 70.8 
    & 81.6 & 73.8 & 80.5 & 56.9 & 72.4 & 46.6 & 77.9 & 80.1 
    & 73.4 & 79.9 & 54.8 & 76.6 & 47.5 & 73.3 & 67.8 & 70.9 \\
    CASENet & 83.3 & 76.0 & 80.7 & 63.4 & 69.2 & 81.3 & 74.9 
    & \textbf{83.2} & 54.3 & 74.8 & 46.4 & 80.3 & 80.2 & 76.6 
    & 80.8 & 53.3 & 77.2 & 50.1 & 75.9 & 66.8 & 71.4 
    \\ \hline 
    \textbf{DDS-R} & 85.4 & 78.3 & 83.3 & 65.6 & 71.4 & 83.0 
    & 75.5 & 81.3 & 59.1 & 75.7 & 50.7 & 80.2 & 82.7 & 77.0 
    & 81.6 & 58.2 & 79.5 & 50.2 & 76.5 & 71.2 & 73.3 
    \\ 
    \textbf{DDS-U} & \textbf{87.2} & \textbf{79.7} & \textbf{84.7} 
    & \textbf{68.3} & \textbf{73.0} & \textbf{83.7} & \textbf{76.7} 
    & 82.3 & \textbf{60.4} & \textbf{79.4} & \textbf{50.9} 
    & \textbf{81.2} & \textbf{83.6} & \textbf{78.3} & \textbf{82.0} 
    & \textbf{60.1} & \textbf{82.7} & \textbf{51.2} & \textbf{78.0} 
    & \textbf{72.7} & \textbf{74.8} 
    \\ \hline
\end{tabular}}
\end{table*}

\section{Experiments}
\subsection{Experimental Settings}
\parahead{Datasets.}
We evaluate our method on the SBD \cite{hariharan2011semantic} and 
Cityscapes \cite{cordts2016cityscapes} datasets.
SBD \cite{hariharan2011semantic} comprises 11,355 images and
corresponding labeled semantic edge maps for 20 object classes.
It is divided into 8498 training and 2857 testing images.
We follow \cite{yu2017casenet} to use the training set to train 
our network and the test set for evaluation.
The Cityscapes dataset \cite{cordts2016cityscapes} is 
a large-scale semantic segmentation dataset with stereo video 
sequences recorded in street scenarios from 50 different cities.
It consists of 5000 images divided into 2975 training, 
500 validation, and 1525 testing images.
The ground truth of the test set has not been published because
it is an online competition for semantic segmentation labeling 
and scene understanding.
Hence, we use the training set for training and the validation set 
for testing.

\myPara{Evaluation metrics.}
For performance evaluation, we adopt several standard metrics 
with the recommended parameter settings in the original papers.
The first metric is the benchmark protocol in \cite{hariharan2011semantic} 
It calculates the class-wise F-measure score that is the harmonic mean of 
the precision and recall.
We follow the default settings with the matching distance tolerance 
of 0.02 for all datasets.
The maximum F-measure at the optimal dataset scale (ODS) for 
each class and mean maximum F-measure across all classes are reported.

We also follow \cite{yu2018simultaneous} to evaluate semantic edges 
with stricter rules than the benchmark in \cite{hariharan2011semantic}.
The ground-truth maps are instance-sensitive edges for \cite{yu2018simultaneous}.
This differs from \cite{hariharan2011semantic} which uses instance-insensitive edges.
Besides, \cite{hariharan2011semantic} thins the prediction 
before matching by default.
\cite{yu2018simultaneous} further proposes to match the raw predictions 
with unthinned ground truths.
This mode and the above conventional mode are referred as ``Raw'' 
and ``Thin'', respectively.
In this paper, we report both the ``Thin'' and ``Raw''  scores for the 
benchmark protocol in \cite{yu2018simultaneous}.
We follow \cite{yu2018simultaneous} to set the matching distance tolerance of 
0.02 for the original SBD dataset \cite{hariharan2011semantic}, 
0.0075 for the re-annotated SBD dataset \cite{yu2018simultaneous},
and 0.0035 for the Cityscapes dataset \cite{cordts2016cityscapes}.
The image borders of 5-pixels width are ignored for the SBD 
dataset, while not for the Cityscapes dataset.

We follow \cite{yu2018simultaneous} to generate both ``Thin'' 
and ``Raw'' ground truths for both instance-sensitive and 
instance-insensitive edges.
The produced edges can be viewed as the boundaries of semantic objects 
or stuff in semantic segmentation.
We downsample the ground truths and predicted edge maps of Cityscapes 
dataset to half the original dimensions to speed up evaluation 
as in previous works 
\cite{yu2017casenet,yu2018simultaneous,hu2019dynamic,acuna2019devil}.
For the performance comparison with baseline methods, 
we use the default code and pre-trained models released by the 
original authors to produce edges.


\begin{table*}[!tbh]
\centering
\renewcommand{\tabcolsep}{0.8mm}
\caption{\revise{Ablation studies for the design of the \textit{information 
	converter} on the SBD dataset \cite{hariharan2011semantic}.
	The results are ODS F-measure (\%) scores using the 
    original benchmark protocol in \cite{hariharan2011semantic}.
    The best performance of each column is highlighted in \textbf{bold}.}}
\label{tab:study_convter}
\small
\resizebox{\linewidth}{!}{
\begin{tabular}{l||c|c|c|c|c|c|c|c|c|c|c|c|c|c|c|c|c|c|c|c||c} 
    \hline\rowcolor{mycolor}
	Methods & aer. & bike & bird & boat & bot. & bus & car 
	& cat & cha. & cow & tab. & dog & hor. & mot. & per. 
	& pot. & she. & sofa & train & tv & mean
	\\ \hline\hline
	1 \conv unit & 85.2 & 78.1 & 82.8 & \textbf{66.0} & \textbf{71.8} 
	& 83.2 & \textbf{75.6} & 80.9 & 58.7 & 75.5 & 49.8 & 79.9 & 82.4 
	& 76.6 & 81.2 & 57.5 & 79.2 & 49.9 & 76.2 & 71.2 & 73.1
	\\
	3 \conv unit & \textbf{85.8} & 78.7 & 83.5 & \textbf{66.0} 
	& \textbf{71.8} & \textbf{83.6} & 75.4 & \textbf{81.4} & 58.9 
	& \textbf{76.9} & 49.5 & \textbf{80.4} & \textbf{83.0} & 76.7
	& \textbf{81.7} & \textbf{58.3} & \textbf{80.2} & \textbf{51.3} 
	& 76.0 & \textbf{71.5} & \textbf{73.5}
	\\
	w/o residual & 85.3 & \textbf{79.0} & \textbf{83.7} & 65.5 
	& 70.9 & \textbf{83.6} & 75.2 & 81.1 & 58.6 & 75.5 & 49.9 & 79.3 
	& 82.3 & 76.8 & 81.3 & 57.7 & 79.3 & 50.6 & \textbf{76.6} 
	& 70.9 & 73.1
	\\
	$\text{DDS-R}\times 1/16$ & 85.7 & 77.9 & 83.9 & 65.2 & 72.0
	& 83.7 & 75.5 & 81.1 & 58.9 & 76.9 & 49.4 & 80.5 & 82.3 
	& 77.2 & 81.2 & 58.3 & 80.4 & 50.6 & 76.5 & 71.6 & 73.4
    \\
    $\text{DDS-R}\times 1/4$ & 85.4 & 78.1 & 83.5 & 65.1 & 71.7
    & 83.2 & 74.8 & 81.5 & 59.0 & 75.3 & 49.0 & 79.5 & 82.3 & 76.3
    & 81.2 & 57.8 & 80.3 & 50.3 & 76.6 & 70.5 & 73.1
    \\
    $\text{DDS-R}\times 4$ & 85.5 & 77.9 & 83.3 & 65.8 & 71.4 
    & 83.1 & 75.2 & 81.4 & 58.6 & 77.1 & 48.6 & 79.9 & 83.1
    & 76.6 & 81.3 & 57.2 & 80.9 & 51.0 & 76.2 & 70.6 & 73.2
    \\ \hline
    \textbf{DDS-R} & 85.4 & 78.3 & 83.3 & 65.6 & 71.4 & 83.0 
    & 75.5 & 81.3 & \textbf{59.1} & 75.7 & \textbf{50.7} & 80.2 
    & 82.7 & \textbf{77.0} & 81.6 & 58.2 & 79.5 & 50.2 & 76.5 
    & 71.2 & 73.3 
    \\ \hline
\end{tabular}}
\vspace{-0.1in}
\end{table*}

\begin{table}[!t]
\centering
\renewcommand{\tabcolsep}{4.mm}
\caption{Class-agnostic evaluation results on the SBD 
    dataset \cite{hariharan2011semantic}.
    The results are ODS F-measure (\%) scores using the 
    original benchmark protocol in \cite{hariharan2011semantic}.}
\label{tab:study_binedge}
\small
\begin{tabular}{c||c|c|c} 
    \hline
	Methods & DSN & CASENet & \textbf{DDS-R}
	\\ \hline\hline
	ODS & 76.6 & 76.4 & \textbf{79.3}
    \\ \hline
\end{tabular}
\vspace{-0.1in}
\end{table}

\subsection{Ablation Studies} \label{sec:ablation_study}
We first perform ablation studies on the SBD dataset \cite{yu2018simultaneous}
to investigate various aspects of the proposed DDS  
before comparing it with existing state-of-the-art methods.
To this end, we propose \revise{seven} DDS variants:
\begin{itemize}
\item \textit{\textbf{Softmax}}, which only adopts the top side 
(Side-5) with a 21-class \textit{softmax} loss function, such that 
the ground-truth edges of each category do not overlap 
and thus each pixel has one specific class label.
\item \textit{\textbf{Basic}}, which employs the top side (Side-5) 
for multi-label classification, meaning that we directly connect 
the loss function of $L_{side}^{(5)}(W)$ on \textit{res5c} 
to train the detector.
\item \textit{\textbf{DSN}}, which directly applies the deeply 
supervised network architecture, in which each side of the 
backbone network is connected to a $1 \times 1$ \conv layer with 
$K$ output channels for SED, and the resulting activation maps from 
all sides are fused to generate the final semantic edges.
\item \textit{\textbf{CASENet+S4}}, which is similar to CASENet 
but takes into consideration Side-4 by connecting it to
a $1 \times 1$ \conv layer to produce a single-channel feature 
map, while CASENet only uses Side-1 $\sim$ Side-3 and Side-5.
\item \textit{\textbf{DDS$\backslash$Convt}}, which removes 
the \textit{information converters} in DDS, such that deep 
supervision is directly imposed after each side.
\item \revise{\textit{DDS$\backslash$Convt$^\dag$}, which not only 
removes the \textit{information converters} in DDS but also applies 
a progressive training strategy, \ie, each block of the ResNet \cite{he2016deep} 
along with their corresponding side-outputs is trained 
separately and then frozen, to simulate the effect of 
\textit{information converters}.}
\item \textit{\textbf{DDS$\backslash$DeSup}}, which removes the 
deep supervision from Side-1 $\sim$ Side-4 of DDS but retains the 
\textit{information converters}.
\end{itemize}
All these variants are trained using the reweighted loss function 
\equref{equ:overallloss} (except \textit{Softmax}) and the 
original SBD dataset for a fair comparison.

We evaluate these variants and the original DDS and CASENet 
\cite{yu2017casenet} on the SBD dataset
using the original benchmark protocol in \cite{hariharan2011semantic}.
The evaluation results are shown in \tabref{tab:ablation_study}.
We can see that \textit{Softmax} suffers from significant 
performance degradation.
Because the predicted semantic edges of neural networks are usually 
thick and overlap with other classes, it is improper to assign 
a single label to each pixel.
Hence, we apply multi-label loss in this paper.
The \textit{Basic} variant achieves an ODS F-measure of 70.6\%, 
which is 0.3\% higher than \textit{DSN}.
This further verifies our hypothesis presented in 
\secref{sec:superv} that features from bottom layers 
are not sufficiently discriminative for semantic classification.
Furthermore, \textit{CASENet+S4} performs better than 
\textit{DSN}, demonstrating that bottom convolutional 
features are more suitable for binary edge detection. 
Moreover, the F-measure of \textit{CASENet+S4} is lower than 
original CASENet.

\myPara{Why does DDS work well?}
The improvement from \textit{DDS$\backslash$DeSup} 
to DDS-R shows that the success of DDS does not arise due to 
more parameters (\conv layers) but instead from the coordination 
between deep supervision and \textit{information converters}.
On the contrary, adding more \conv layers but without 
deep supervision may make network convergence more difficult.
Our conclusion is consistent with \cite{yu2017casenet}, when 
comparing \textit{DDS$\backslash$Convt} with the results of 
CASENet, namely that there is no value in directly adding 
binary edge supervision to bottom sides.

\begin{table*}[!tbh]
\centering
\renewcommand{\tabcolsep}{0.8mm}
\caption{ODS F-measure (\%) of DDS-R/DDS-U and other
    competitors on the SBD dataset \cite{hariharan2011semantic}. 
    The best performance of each column is highlighted in 
    \textbf{bold}.}
\label{tab:eval_sbd_fm}
\small
\resizebox{\linewidth}{!}{
\begin{tabular}{l||c|c|c|c|c|c|c|c|c|c|c|c|c|c|c|c|c|c|c|c||c} 
    \hline\rowcolor{mycolor}
	Methods & aer. & bike & bird & boat & bot. & bus & car 
	& cat & cha. & cow & tab. & dog & hor. & mot. & per. 
	& pot. & she. & sofa & train & tv & mean 
	\\ \hline\hline\rowcolor{mycolor}
	\multicolumn{22}{c}{With the evaluation metric 
	    in \cite{hariharan2011semantic}} 
	\\ \hline
    InvDet & 41.5 & 46.7 & 15.6 & 17.1 & 36.5 & 42.6 & 40.3 
    & 22.7 & 18.9 & 26.9 & 12.5 & 18.2 & 35.4 & 29.4 & 48.2 
    & 13.9 & 26.9 & 11.1 & 21.9 & 31.4 & 27.9 
    \\
    HFL-FC8 & 71.6 & 59.6 & 68.0 & 54.1 & 57.2 & 68.0 & 58.8 
    & 69.3 & 43.3 & 65.8 & 33.3 & 67.9 & 67.5 & 62.2 & 69.0 
    & 43.8 & 68.5 & 33.9 & 57.7 & 54.8 & 58.7 
    \\
    HFL-CRF & 73.9 & 61.4 & 74.6 & 57.2 & 58.8 & 70.4 & 61.6 
    & 71.9 & 46.5 & 72.3 & 36.2 & 71.1 & 73.0 & 68.1 & 70.3 
    & 44.4 & 73.2 & 42.6 & 62.4 & 60.1 & 62.5 
    \\
    BNF & 76.7 & 60.5 & 75.9 & 60.7 & 63.1 & 68.4 & 62.0 
    & 74.3 & 54.1 & 76.0 & 42.9 & 71.9 & 76.1 & 68.3 & 70.5 
    & 53.7 & 79.6 & \textbf{51.9} & 60.7 & 60.9 & 65.4 
    \\
    WS & 65.9 & 54.1 & 63.6 & 47.9 & 47.0 & 60.4 & 50.9 
    & 56.5 & 40.4 & 56.0 & 30.0 & 57.5 & 58.0 & 57.4 & 59.5 
    & 39.0 & 64.2 & 35.4 & 51.0 & 42.4 & 51.9 
    \\
    DilConv & 83.7 & 71.8 & 78.8 & 65.5 & 66.3 & 82.6 & 73.0 
    & 77.3 & 47.3 & 76.8 & 37.2 & 78.4 & 79.4 & 75.2 & 73.8 
    & 46.2 & 79.5 & 46.6 & 76.4 & 63.8 & 69.0 
    \\
    DSN & 81.6 & 75.6 & 78.4 & 61.3 & 67.6 & 82.3 & 74.6 
    & 82.6 & 52.4 & 71.9 & 45.9 & 79.2 & 78.3 & 76.2 
    & 80.1 & 51.9 & 74.9 & 48.0 & 76.5 & 66.8 & 70.3 
    \\
    COB & 84.2 & 72.3 & 81.0 & 64.2 & 68.8 & 81.7 & 71.5 
    & 79.4 & 55.2 & 79.1 & 40.8 & 79.9 & 80.4 & 75.6 & 77.3 
    & 54.4 & \textbf{82.8} & 51.7 & 72.1 & 62.4 & 70.7 
    \\ \hline
    CASENet & 83.3 & 76.0 & 80.7 & 63.4 & 69.2 & 81.3 
    & 74.9 & \textbf{83.2} & 54.3 & 74.8 & 46.4 & 80.3 & 80.2 
    & 76.6 & 80.8 & 53.3 & 77.2 & 50.1 & 75.9 & 66.8 & 71.4 
    \\ 
    SEAL & 85.2 & 77.7 & 83.4 & 66.3 & 70.6 & 82.4 & 75.2 
    & 82.3 & 58.5 & 76.5 & 50.4 & 80.9 & 82.2 & 76.8 
    & \textbf{82.2} & 57.1 & 78.9 & 50.4 & 75.8 & 70.1 & 73.1 
    \\
    \textbf{DDS-R} & 85.4 & 78.3 & 83.3 & 65.6 & 71.4 & 83.0 
    & 75.5 & 81.3 & 59.1 & 75.7 & 50.7 & 80.2 & 82.7 & 77.0 
    & 81.6 & 58.2 & 79.5 & 50.2 & 76.5 & 71.2 & 73.3 
    \\
    \textbf{DDS-U} & \textbf{87.2} & \textbf{79.7} & \textbf{84.7} 
    & \textbf{68.3} & \textbf{73.0} & \textbf{83.7} & \textbf{76.7} 
    & 82.3 & \textbf{60.4} & \textbf{79.4} & \textbf{50.9} 
    & \textbf{81.2} & \textbf{83.6} & \textbf{78.3} & 82.0 
    & \textbf{60.1} & 82.7 & 51.2 & \textbf{78.0} & \textbf{72.7} 
    & \textbf{74.8} 
    \\ \hline\hline\rowcolor{mycolor}
    \multicolumn{22}{c}{With the ``Thin'' evaluation metric 
        in \cite{yu2018simultaneous}} 
	\\ \hline
	CASENet & 83.6 & 75.3 & 82.3 & 63.1 & 70.5 & 83.5 & 76.5 & 82.6 
	& 56.8 & 76.3 & 47.5 & 80.8 & 80.9 & 75.6 & 80.7 & 54.1 & 77.7
	& 52.3 & 77.9 & 68.0 & 72.3
	\\
	SEAL & 84.5 & 76.5 & 83.7 & 64.9 & 71.7 & 83.8 & 78.1 & 85.0
	& 58.8 & 76.6 & 50.9 & 82.4 & 82.2 & 77.1 & 83.0 & 55.1 & 78.4
	& 54.4 & 79.3 & 69.6 & 73.8
	\\
	STEAL & 85.2 & 77.3 & 84.0 & 65.9 & 71.1 & 85.3 & 77.5 & 83.8
	& 59.2 & 76.4 & 50.0 & 81.9 & 82.2 & 77.3 & 81.7 & 55.7 & 79.5 
	& 52.3 & 79.2 & 69.8 & 73.8
	\\
	Gated-SCNN & 81.6 & 70.5 & 73.9 & 60.2 & 64.1 & 82.5 & 72.9
	& 78.0 & 51.8 & 67.3 & 42.2 & 74.6 & 74.3 & 71.4 & 77.6 
	& 49.3 & 72.3 & 46.6 & 73.7 & 57.0 & 67.1
	\\
	\textbf{DDS-R} & 85.6 & 77.1 & 82.8 & 64.0 & 73.5 & 85.4 & 78.8 
	& 84.4 & 57.7 & 77.6 & 51.9 & 81.2 & 82.4 & 77.1 & 82.5 & 56.3 
	& 79.5 & 54.5 & 80.3 & 70.4 & 74.1
	\\
	\textbf{DDS-U} & 86.5 & 78.4 & 84.4 & 67.0 & 74.3 & 85.8 & 80.2 
	& \textbf{85.9} & 60.4 & \textbf{80.8} & \textbf{53.9} & 83.0 
	& 84.4 & \textbf{78.8} & \textbf{83.9} & \textbf{58.7} 
	& \textbf{81.9} & \textbf{56.0} & \textbf{82.1} & \textbf{73.0} 
	& \textbf{76.0}
	\\ \hline
	DFF & 86.5 & 79.5 & 85.5 & \textbf{69.0} & 73.9 & 86.1 & 80.3 
	& 85.3 & 58.5 & 80.1 & 47.3 & 82.5 & \textbf{85.7} & 78.5 
	& 83.4 & 57.9 & 81.2 & 53.0 & 81.4 & 71.6 & 75.4
	\\
	\textbf{DDS-R} & \textbf{86.7} & \textbf{79.6} & \textbf{85.6} 
	& 68.4 & \textbf{74.5} & \textbf{86.5} & \textbf{81.1} 
	& \textbf{85.9} & \textbf{60.5} & 79.3 & 53.5 & \textbf{83.2} 
	& 85.2 & \textbf{78.8} & \textbf{83.9} & 58.4 
	& 80.8 & 54.4 & 81.8 & 72.2 & \textbf{76.0}
	\\ \hline\hline\rowcolor{mycolor}
    \multicolumn{22}{c}{With the ``Raw'' evaluation metric 
        in \cite{yu2018simultaneous}} 
    \\ \hline
    CASENet & 71.8 & 60.2 & 72.6 & 49.5 & 59.3 & 73.3 & 65.2
    & 70.8 & 51.9 & 64.9 & 41.2 & 67.9 & 72.5 & 64.1 & 71.2 
    & 44.0 & 71.7 & 45.7 & 65.4 & 55.8 & 62.0
    \\
    SEAL & 81.1 & 69.6 & 81.7 & 60.6 & 68.0 & 80.5 & 75.1 & 80.7
    & 57.0 & 73.1 & 48.1 & 78.2 & 80.3 & 72.1 & 79.8 & 50.0 
    & 78.2 & 51.8 & 74.6 & 65.0 & 70.3
    \\
    STEAL & 77.2 & 66.2 & 78.9 & 56.8 & 63.2 & 77.8 & 71.9 & 75.3
    & 55.0 & 69.4 & 43.8 & 73.1 & 76.9 & 69.8 & 75.5 & 48.3 & 76.2
    & 47.7 & 70.4 & 60.5 & 66.7
    \\
    Gated-SCNN & 70.4 & 56.9 & 64.8 & 49.6 & 54.7 & 70.5 & 61.9
    & 66.0 & 46.9 & 55.3 & 36.7 & 61.0 & 62.4 & 59.9 & 67.6 
    & 39.5 & 68.2 & 40.1 & 59.6 & 49.1 & 57.1
    \\
    \textbf{DDS-R} & 80.5 & 68.2 & 78.6 & 56.4 & 67.6 & 80.9 & 72.7 
    & 77.6 & 55.4 & 70.9 & 47.0 & 74.9 & 77.5 & 70.0 & 77.4 & 50.9 
    & 75.7 & 50.7 & 74.5 & 65.5 & 68.6
    \\
    \textbf{DDS-U} & \textbf{83.8} & \textbf{71.8} & \textbf{82.1} 
    & \textbf{61.7} & \textbf{70.4} & \textbf{82.9} & \textbf{76.9} 
    & \textbf{80.8} & \textbf{58.5} & \textbf{77.1} & \textbf{49.9} 
    & \textbf{77.8} & \textbf{81.5} & \textbf{73.5} & \textbf{81.0} 
    & \textbf{52.9} & \textbf{81.3} & \textbf{53.0} & \textbf{76.3} 
    & \textbf{69.1} & \textbf{72.1}
    \\ \hline
    DFF & 77.6 & 65.7 & 79.3 & 57.2 & 65.5 & 78.5 & 72.0 & 76.2
    & 53.7 & 71.9 & 42.5 & 72.0 & 77.0 & 68.8 & 75.1 & 50.6 
    & 76.6 & 46.9 & 71.9 & 63.6 & 67.1
    \\
    \textbf{DDS-R} & 79.2 & 67.6 & 77.7 & 58.7 & 65.9 & 81.0 & 72.9 
    & 76.6 & 55.8 & 70.3 & 47.6 & 74.0 & 76.9 & 68.8 & 76.5 & 52.5 
    & 77.0 & 48.8 & 72.8 & 65.7 & 68.3
	\\ \hline
\end{tabular}}
\vspace{-0.1in}
\end{table*}

\myPara{Discussion about the proposed DDS.}
Intuitively, employing different but ``appropriate'' 
ground truths to bottom and top sides may enhance 
the feature learning in different layers. 
Upon this, the learned intermediate representations of different 
levels will tend to contain complementary information. 
\revise{However, in our case, it may be useless to directly add 
deep supervision of category-agnostic edges to bottom sides, 
because bottom layers would receive two distinct supervision
in the loss function of \equref{equ:overallloss}, as discussed above.}
Instead, we show that with proper architecture re-design, 
we can employ deep supervision to significantly boost 
performance. 
The \textit{information converters} adopted in the proposed 
method play a central role in guiding bottom layers for 
category-agnostic edge detection. 
In this way, low-level edges from bottom layers encode more details, 
which then assist top layers to better localize semantic edges.
\revise{They serve as buffers to separate the information content 
in the backbone layers.
This is essential, as they enable bottom layers to serve as the basis 
of top layers to favor highly discriminative feature maps 
for correct semantic classification.}

The significant improvement provided by the proposed 
DDS-R/DDS-U over \textit{CASENet+S4} and \textit{DDS$\backslash$Convt} 
demonstrates the importance of our design, in which different sides 
use different supervision after the information format conversion.
We also note that DDS-U achieves better performance than DDS-R 
by applying the unweighted loss function and aligned edges 
\cite{yu2018simultaneous}.

\myPara{\revise{Progressive training.}}
\revise{\textit{DDS$\backslash$Convt$^\dag$} adopts progressive training 
to simulate the effect of \textit{information converters}, 
as suggested by a paper reviewer.
However, from \tabref{tab:ablation_study}, we can see that 
\textit{DDS$\backslash$Convt$^\dag$} performs significantly worse than other variants.
This is because progressive training cannot optimize DCNNs well.
As widely acknowledged, it is necessary for DCNNs to adopt end-to-end 
training for deriving optimal parameters.
In fact, progressive training is an old way for DCNN training \cite{hinton2006fast}.
After the invention of ReLU \cite{nair2010rectified}, 
batch normalization \cite{ioffe2015batch}, and dropout \cite{srivastava2014dropout},
progressive training has not been used in the deep learning community
due to its poor performance.
Hence, \textit{DDS$\backslash$Convt$^\dag$} cannot simulate 
\textit{information converters}.}

\myPara{Discussion about the design of the \textit{information converter}.}
\revise{This paper mainly discusses and resolves the distinct supervision 
targets in SED, and the core is the existence, not 
the specific format, of the \textit{information converter}.}
Hence we design a simple \textit{information converter} that consists 
of two sequential residual \conv units.
Here, we conduct ablation studies for this design.
Results are shown in \tabref{tab:study_convter}.
We experiment with three different converter designs: 
i) with only one \conv unit;
ii) with three \conv units; 
iii) without residual connections in the \conv units (plain \conv units).
It is easy to observe that the \textit{information converter} with 
three residual \conv units achieves the best performance, but it is only 
slightly better than that with two residual \conv units.
To make a trade-off between model complexity and performance,
we use two residual \conv units as the default setting.

\revise{We also evaluate the effect of the number of parameters of 
the \textit{information converter}.
Here, we change its size by simply multiplying a constant to the number 
of channels of the default \textit{information converter}.
In this way, the resulting variants have various model sizes but keep the same structure.
Specifically, we try three constants of $1/4$, $1/2$, and $2$,
leading to $1/16$, $1/4$, and $4$ times of the default model size, respectively.
The results are depicted in \tabref{tab:study_convter}.
It is interesting to find that the proposed method is quite robust to different
model sizes, demonstrating that the improvement mainly comes from
the existence of the \textit{information converter}, not its specific format.}

\myPara{Improvement of the edge localization.}
To demonstrate if the introduced information converter actually 
improves the localization of the semantic edges, we ignore the semantic 
labels and perform class-agnostic evaluation for the proposed DDS and 
previous baselines.
Given an input image, SED methods generate an edge probability map
for each class.
To generate a class-agnostic edge map for an image,
at each pixel, we view the maximum edge probability across all classes 
as the class-agnostic edge probability at this pixel.
For ground truth, at each pixel, if any class has an edge on this pixel, 
this pixel is viewed as a class-agnostic edge pixel.
Then, we use the standard benchmark in \cite{hariharan2011semantic}
for evaluation.
From \tabref{tab:study_binedge}, we find DDS can significantly improve
the edge localization accuracy, which demonstrates that imposing 
class-agnostic edge supervision at bottom network sides can well benefit
edge localization.
\revise{After exploring DDS with several variants and establishing the 
effectiveness of the approach, we summarize the results obtained 
by our method and compare it with previous state-of-the-art methods.}

\renewcommand{\AddImg}[1]{\fbox{\includegraphics%
[width=.315\linewidth]{2008_003379#1}}}
\renewcommand{\AddColor}[1]{\myfbox{\includegraphics%
[width=.315\linewidth]{2008_003379#1}}}
\begin{figure}[!t]
    \centering
    \setlength{\fboxrule}{0.6pt}
    \setlength{\fboxsep}{0pt}
    \AddImg{}
    \AddImg{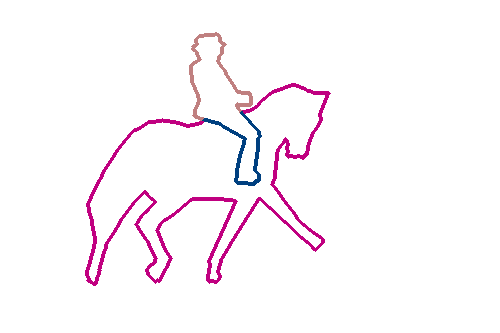}
    \AddColor{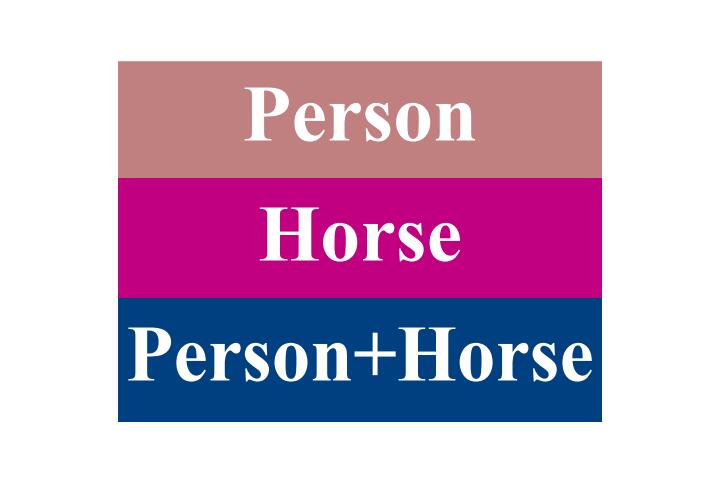}
    \\ \vspace{-0.08in} \rule{\linewidth}{0.5mm}
    \leftline{\scriptsize\hspace{0.405in} DSN \hspace{0.665in} 
    CASENet \hspace{0.59in} DDS-R}
    \\ \vspace{0.03in}
    \AddImg{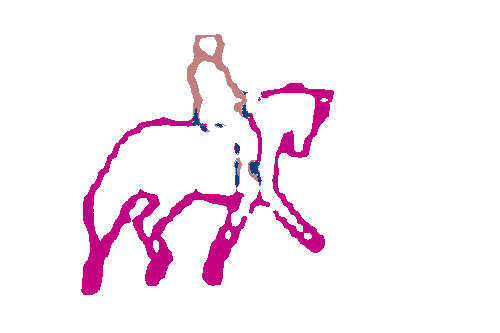}
    \AddImg{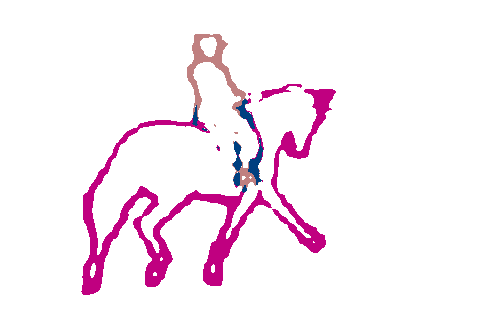}
    \AddImg{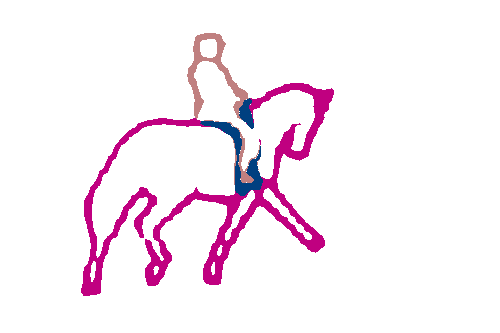}
    \\ \vspace{0.05in}
    \AddImg{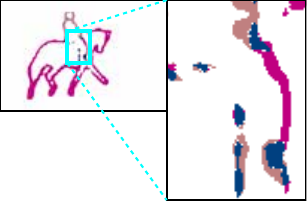}
    \AddImg{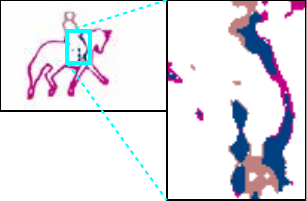}
    \AddImg{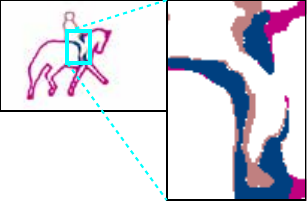}
    \\ \vspace{0.05in}
    \AddImg{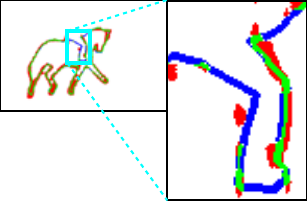}
    \AddImg{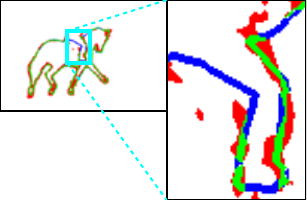}
    \AddImg{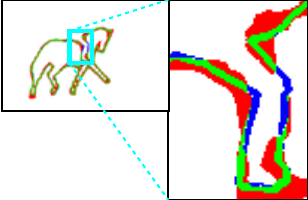}
    \\ \vspace{0.05in}
    \AddImg{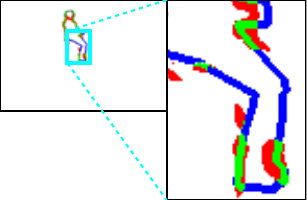}
    \AddImg{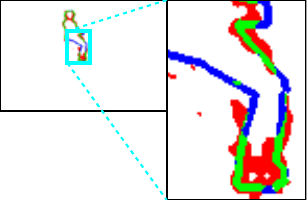}
    \AddImg{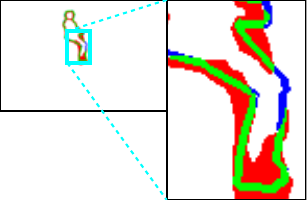}
    \\
    \caption{A qualitative comparison of DSN, CASENet and DDS-R. 
    First row: the original image, ground truth, and category 
    color codes.
    This image is taken from the SBD dataset \cite{hariharan2011semantic}.
    Second row: the semantic edges predicted by different methods.
    Third row: an enlarged area of predicted edges.
    Fourth row: the predicted horse boundaries only. 
    Last row: the predicted person boundaries only.
    Green, red, white, and blue pixels represent 
    true positive, false positive, true negative, and false 
    negative points, respectively, at the threshold of 0.5.
    Best viewed in color.
    } \label{fig:sbd_samples}
\end{figure}

\begin{table*}[!ht]
\centering
\renewcommand{\tabcolsep}{0.8mm}
\caption{\revise{ODS F-measure (\%) of DDS-R/DDS-U and other competitors on the
	re-annotated SBD dataset \cite{yu2018simultaneous}. The best 
	performance of each column is highlighted in \textbf{bold}.}}
\label{tab:eval_reanno_sbd}
\small
\resizebox{\linewidth}{!}{
\begin{tabular}{l||c|c|c|c|c|c|c|c|c|c|c|c|c|c|c|c|c|c|c|c||c} 
    \hline\rowcolor{mycolor}
	Methods & aer. & bike & bird & boat & bot. & bus & car 
	& cat & cha. & cow & tab. & dog & hor. & mot. & per. 
	& pot. & she. & sofa & train & tv & mean 
	\\ \hline\hline\rowcolor{mycolor}
    \multicolumn{22}{c}{With the ``Thin'' evaluation metric 
        in \cite{yu2018simultaneous}} 
	\\ \hline
	CASENet & 74.5 & 59.7 & 73.4 & 48.0 & 67.1 & 78.6 & 67.3 & 76.2
	& 47.5 & 69.7 & 36.2 & 75.7 & 72.7 & 61.3 & 74.8 & 42.6 & 71.8
	& 48.9 & 71.7 & 54.9 & 63.6
	\\
	SEAL & 78.0 & 65.8 & 76.6 & 52.4 & 68.6 & 80.0 & 70.4 & 79.4 
	& 50.0 & 72.8 & 41.4 & 78.1 & 75.0 & 65.5 & 78.5 & 49.4 & 73.3
	& 52.2 & 73.9 & 58.1 & 67.0
	\\
	STEAL & 77.1 & 63.6 & 76.2 & 51.1 & 68.0 & 80.4 & 70.0 & 76.8
	& 49.4 & 71.9 & 40.4 & 78.1 & 74.7 & 64.5 & 75.7 & 45.4 & 73.5
	& 47.5 & 73.5 & 58.7 & 65.8
	\\
	Gated-SCNN & 74.8 & 58.4 & 65.9 & 47.1 & 63.0 & 74.5 & 65.6 
    & 71.6 & 41.4 & 61.6 & 39.4 & 70.8 & 65.0 & 57.4 & 72.9 & 44.1
    & 69.3 & 44.0 & 64.5 & 50.5 & 60.1
	\\
	\textbf{DDS-R} & 79.7 & 65.2 & 74.6 & 51.8 & 71.9 & 81.3 & 72.5 
	& 79.4 & 49.2 & 75.1 & 43.9 & 77.8 & 75.3 & 65.2 & 78.9 & 51.1 
	& 74.9 & 54.1 & 75.1 & 61.7 & 67.9
	\\
	\textbf{DDS-U} & \textbf{81.4} & 67.6 & 77.8 & \textbf{55.7} 
	& 70.9 & 82.0 & 74.5 & \textbf{81.2} & \textbf{52.1} & 76.5 
	& \textbf{47.2} & \textbf{79.6} & 77.3 & \textbf{68.1} 
	& \textbf{80.2} & \textbf{53.4} & \textbf{78.5} & \textbf{56.1} 
	& 76.6 & 63.9 & \textbf{70.0}
	\\ \hline
	DFF & 78.6 & 66.2 & 77.9 & 53.2 & \textbf{72.3} & 81.3 & 73.3 
	& 79.0 & 50.7 & \textbf{76.8} & 38.7 & 77.2 & \textbf{78.6} 
	& 65.2 & 77.9 & 49.4 & 76.1 & 49.7 & 74.7 & 62.9 & 68.0
	\\
	\textbf{DDS-R} & 78.8 & \textbf{68.0} & \textbf{78.3} & 55.0 
	& 71.9 & \textbf{82.4} & \textbf{74.6} & 80.5 & 52.0 & 74.0 
	& 42.0 & 78.3 & 77.1 & 66.1 & 78.5 & 49.3 & 77.5 & 49.3 
	& \textbf{76.9} & \textbf{64.8} & 68.8
	\\ \hline\hline\rowcolor{mycolor}
    \multicolumn{22}{c}{With the ``Raw'' evaluation metric 
        in \cite{yu2018simultaneous}} 
    \\ \hline
    CASENet & 65.8 & 51.5 & 65.0 & 43.1 & 57.5 & 68.1 & 58.2 & 66.0
    & 45.4 & 59.8 & 32.9 & 64.2 & 65.8 & 52.6 & 65.7 & 40.9 & 65.0
    & 42.9 & 61.4 & 47.8 & 56.0
    \\
    SEAL & 75.3 & 60.5 & 75.1 & 51.2 & 65.4 & 76.1 & 67.9 & 75.9 
    & 49.7 & 69.5 & 39.9 & \textbf{74.8} & 72.7 & 62.1 & 74.2 
    & 48.4 & 72.3 & 49.3 & 70.6 & 56.7 & 64.4
    \\
    STEAL & 70.9 & 55.9 & 71.6 & 47.6 & 61.5 & 72.6 & 64.6 & 70.2 
    & 47.5 & 67.4 & 37.3 & 70.6 & 69.4 & 59.1 & 69.2 & 44.3 & 69.1
    & 42.6 & 67.7 & 53.5 & 60.6
    \\
    Gated-SCNN & 66.8 & 50.1 & 59.4 & 44.2 & 54.9 & 64.6 & 57.9 
    & 62.2 & 39.6 & 50.9 & 35.9 & 59.5 & 56.5 & 48.9 & 64.2 
    & 41.3 & 64.0 & 35.6 & 54.2 & 45.5 & 52.8
    \\
    \textbf{DDS-R} & 75.6 & 61.1 & 71.0 & 49.5 & 67.7 & 76.1 & 67.2 & 74.2 
    & 48.8 & 69.1 & 40.4 & 72.5 & 71.7 & 60.4 & 73.4 & 49.6
    & 70.6 & 49.5 & 71.9 & 59.4 & 64.0
    \\
    \textbf{DDS-U} & \textbf{78.4} & \textbf{62.7} & \textbf{75.6} 
    & \textbf{53.4} & \textbf{67.8} & \textbf{78.5} & \textbf{71.4} 
    & \textbf{77.4} & \textbf{51.3} & \textbf{72.8} & \textbf{44.5} 
    & 74.7 & \textbf{74.8} & \textbf{64.3} & \textbf{76.3} 
    & \textbf{51.9} & \textbf{77.3} & \textbf{51.9} & \textbf{73.7} 
    & \textbf{62.9} & \textbf{67.1}
    \\ \hline
    DFF & 72.3 & 58.4 & 73.4 & 48.7 & 65.4 & 74.8 & 66.4 & 72.5 
    & 47.8 & 70.1 & 34.7 & 69.2 & 71.5 & 58.7 & 70.2 & 47.5 
    & 71.2 & 43.7 & 69.5 & 59.1 & 62.3
    \\
    \textbf{DDS-R} & 74.2 & 61.2 & 71.3 & 51.9 & 65.5 & 77.3 & 68.0
    & 73.8 & 50.0 & 66.0 & 39.4 & 70.8 & 70.5 & 58.9 & 71.8 & 49.0
    & 72.6 & 44.7 & 71.6 & 62.2 & 63.5
	\\ \hline
\end{tabular}}
\end{table*}

\subsection{Evaluation on SBD}
In this part, we compare DDS-R/DDS-U on the SBD dataset \cite{hariharan2011semantic}
with previous state-of-the-art methods, including 
InvDet \cite{hariharan2011semantic},
HFL-FC8 \cite{bertasius2015high}, HFL-CRF \cite{bertasius2015high},
BNF \cite{bertasius2016semantic}, WS \cite{khoreva2016weakly},
DilConv \cite{yu2015multi}, DSN \cite{yu2017casenet}, 
COB \cite{maninis2017convolutional}, CASENet \cite{yu2017casenet},
SEAL \cite{yu2018simultaneous}, STEAL \cite{acuna2019devil},
DFF \cite{hu2019dynamic}, \revise{and Gated-SCNN} \cite{takikawa2019gated}.
Among them, DFF \cite{hu2019dynamic} shares the same distinct supervision 
problem as CASENet, so we also integrate DDS-R into DFF to demonstrate 
the generalizability of DDS-R.
We adopt the same code implementation and training strategies 
for DFF-based DDS-R as the original DFF.
\revise{Gated-SCNN \cite{takikawa2019gated} learns 
semantic edges for improving the training of semantic segmentation.
Hence, we retrain it for semantic edge detection by removing its 
segmentation loss and dual task loss, and the other settings 
are kept by default.}

Results are summarized in \tabref{tab:eval_sbd_fm}.
DDS-U achieves the \sArt performance across all competitors.
The ODS F-measure of the proposed DDS-U is 1.7\% higher than 
SEAL and 3.4\% higher than CASENet in terms of the metric 
in \cite{hariharan2011semantic}, so delivering a new state-of-the-art.
We can observe that DDS-R can also improve the performance of DFF \cite{hu2019dynamic}.
Therefore, the proposed DDS can be viewed as a general idea 
to improve SED.
The improvement from CASENet to DDS is also larger than the 
improvement of STEAL.
Moreover, InvDet \cite{hariharan2011semantic} is a non-deep 
learning based approach which shows competitive results among 
other conventional approaches.
COB \cite{maninis2017convolutional} is a \sArt category-agnostic 
edge detection method, and combining it with semantic segmentation 
of DilConv \cite{yu2015multi} produces a competitive semantic 
edge detector.
COB's superiority over DilConv reflects the effectiveness
of its fusion algorithm.
The fact that both CASENet and DDS-R/DDS-U outperform COB 
illustrates the importance of directly learning semantic edges, 
because the combination of binary edges and semantic 
segmentation is insufficient for SED.
The average runtime of DSN, CASENet, and DDS is shown 
in \tabref{tab:time_sbd}.
DDS can generate \sArt semantic edges with only a slight 
reduction in speed.

\begin{table}[!t]
\centering
\caption{Average runtime per image on the SBD dataset 
\cite{hariharan2011semantic}.}
\label{tab:time_sbd}
\small
\begin{tabular}{c|c|c|c|c} \hline
    Methods & DSN & CASENet & SEAL & DDS \\ \hline
    Time (s) & 0.171 & 0.166 & 0.166 & 0.175 \\ \hline
\end{tabular}
\end{table}

\begin{table*}[!ht]
\centering
\renewcommand{\tabcolsep}{0.8mm}
\caption{\revise{ODS F-measure (\%) of \revise{DDS-R/DDS-U and other} 
    competitors on the Cityscapes dataset \cite{cordts2016cityscapes}. 
    The best performance of each column is highlighted in \textbf{bold}.}}
\label{tab:eval_cityscapes}
\small
\resizebox{\linewidth}{!}{
\begin{tabular}{l|c|c|c|c|c|c|c|c|c|c|c|c|c|c|c|c|c|c|c||c} 
    \hline\rowcolor{mycolor}
	Methods & road & sid. & bui. & wall & fen. & pole & light 
	& sign & veg. & ter. & sky & per. & rider & car & tru. 
	& bus & tra. & mot. & bike & mean
    \\ \hline\hline\rowcolor{mycolor}
    \multicolumn{21}{c}{With the ``Thin'' evaluation metric 
        in \cite{yu2018simultaneous}} 
	\\ \hline
	CASENet & 86.2 & 74.9 & 74.5 & 47.6 & 46.5 & 72.8 & 70.0 & 73.3 
	& 79.3 & 57.0 & 86.5 & 80.4 & 66.8 & 88.3 & 49.3 & 64.6 & 47.8 
	& 55.8 & 71.9 & 68.1 
	\\
	SEAL & 87.6 & 77.5 & 75.9 & 47.6 & 46.3 & 75.5 & 71.2 & 75.4 
	& 80.9 & 60.1 & 87.4 & 81.5 & 68.9 & 88.9 & 50.2 & 67.8 & 44.1 
	& 52.7 & 73.0 & 69.1
	\\
	STEAL & 87.8 & 77.2 & 76.4 & 49.5 & 49.2 & 74.9 & 73.2 & 76.3 
	& 80.8 & 58.9 & 86.8 & 80.2 & 69.0 & 83.2 & 52.1 & 67.7 & 53.2
	& 55.8 & 72.8 & 69.7
	\\
	Gated-SCNN & 88.9 & 78.7 & 78.3 & 51.3 & 52.3 & 78.6 & 78.7 
	& 78.0 & 82.2 & 62.3 & 87.7 & 83.5 & 70.8 & 90.6 & 33.1 & 60.4
	& 31.6 & 50.9 & 74.4 & 69.1
	\\
	\textbf{DDS-R} & 86.1 & 76.5 & 76.1 & 49.8 & 49.9 & 74.6 & 76.4 
	& 76.8 & 80.4 & 58.9 & 87.2 & 83.5 & 70.7 & 89.6 & 52.9 & 71.5 
	& 50.4 & 61.8 & 74.4 & 70.9
	\\
	\textbf{DDS-U} & 89.2 & 79.2 & 79.0 & 51.9 & 52.9 & 77.5 & 79.4 
	& 80.3 & 82.6 & 61.4 & 88.8 & 85.0 & 74.1 & 91.1 & 59.0 
	& \textbf{76.0} & \textbf{55.7} & 63.6 & 76.3 & 73.8
	\\ \hline
	DFF & 89.4 & \textbf{80.1} & 79.6 & 51.3 & \textbf{54.5} & 81.3 
	& 81.3 & \textbf{81.2} & 83.6 & \textbf{62.9} & 89.0 & 85.4 
	& 75.8 & 91.6 & 54.9 & 73.9 & 51.9 & 64.3 & 76.4 & 74.1
	\\
	\textbf{DDS-R} & \textbf{89.7} & 79.4 & \textbf{80.4} 
	& \textbf{52.1} & 53.0 & \textbf{82.4} & \textbf{81.9}
	& 80.9 & \textbf{83.9} & 62.0 & \textbf{89.4} & \textbf{86.0} 
	& \textbf{77.8} & \textbf{92.3} & \textbf{59.8} 
	& 74.8 & 55.3 & \textbf{64.4} & \textbf{77.4} & \textbf{74.9}
	\\ \hline\hline\rowcolor{mycolor}
    \multicolumn{21}{c}{With the ``Raw'' evaluation metric 
        in \cite{yu2018simultaneous}} 
    \\ \hline
    CASENet & 66.8 & 64.6 & 66.8 & 39.4 & 40.6 & 71.7 & 64.2
    & 65.1 & 71.1 & 50.2 & 80.3 & 73.1 & 58.6 & 77.0 & 42.0
    & 53.2 & 39.1 & 46.1 & 62.2 & 59.6
    \\
    SEAL & \textbf{84.4} & 73.5 & 72.7 & \textbf{43.4} 
    & \textbf{43.2} & 76.1 & 68.5 & 69.8 & 77.2 & \textbf{57.5} 
    & \textbf{85.3} & 77.6 & 63.6 & 84.9 & \textbf{48.6} & 61.9 
    & \textbf{41.2} & 49.0 & 66.7 & 65.5
    \\
    STEAL & 75.8 & 68.5 & 69.8 & 34.9 & 36.1 & 73.4 & 66.7 
    & 67.7 & 73.5 & 49.7 & 78.7 & 72.9 & 59.1 & 76.5 & 35.3
    & 52.8 & 37.7 & 43.8 & 63.7 & 59.8
    \\
    Gated-SCNN & 77.3 & 69.7 & 74.8 & 38.2 & 40.1 & 79.7 & 72.6
    & 72.4 & 77.7 & 54.2 & 82.0 & 77.7 & 62.0 & 86.1 & 17.1 
    & 37.7 & 14.3 & 37.5 & 66.8 & 59.9
    \\
    \textbf{DDS-R} & 73.3 & 65.9 & 70.9 & 33.2 & 37.4 & 76.8 
    & 70.1 & 70.2 & 74.6 & 50.4 & 80.6 & 77.9 & 62.6 & 82.5
    & 37.1 & 55.0 & 32.0 & 49.4 & 66.1 & 61.4
    \\
    \textbf{DDS-U} & 83.5 & \textbf{74.2} & 76.0 & 37.5 & 40.7 
    & 79.5 & 75.6 & 75.3 & \textbf{79.3} & 55.7 & \textbf{85.3} 
    & 81.1 & 67.1 & \textbf{87.9} & 44.6 & \textbf{63.4} & 40.4 
    & 52.3 & 70.0 & \textbf{66.8}
    \\ \hline
    DFF & 72.8 & 68.3 & 72.6 & 37.2 & 42.2 & 79.6 & 75.0
    & 73.9 & 75.3 & 51.4 & 80.8 & 78.6 & 69.4 & 83.0 & 44.1
    & 56.7 & 38.4 & 52.0 & 68.8 & 64.2
    \\
    \textbf{DDS-R} & 80.8 & 70.8 & \textbf{76.4} & 38.9 & 41.1 
    & \textbf{80.0} & \textbf{78.2} & \textbf{76.3} & 79.2 & 53.2 
    & 82.5 & \textbf{81.8} & \textbf{72.2} & 86.2 & 44.8 & 59.5 
    & 37.6 & \textbf{55.7} & \textbf{71.3} & 66.7 
	\\ \hline
\end{tabular}}
\end{table*}

\renewcommand{\AddImg}[1]{\fbox{\includegraphics%
[width=.23\linewidth]{#1}}}
\begin{figure}[!t]
    \centering
     \setlength{\fboxrule}{0.6pt}
     \setlength{\fboxsep}{0pt}
     \scriptsize
     \leftline{\hspace{0.13in} DSN-Horse \hspace{0.175in} 
     DSN-Person \hspace{0.207in} CASENet \hspace{0.33in} DDS-R}
     \vspace{0.03in}
     \stackunder[2pt]{
       \AddImg{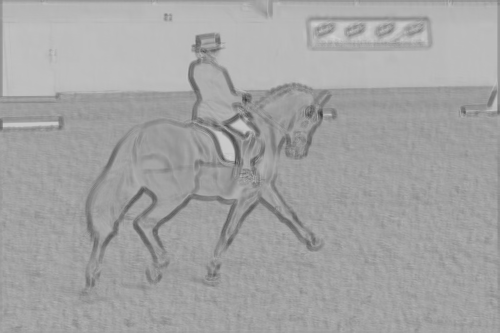}
       \AddImg{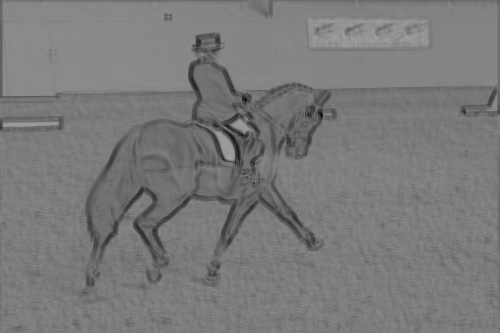}
       \AddImg{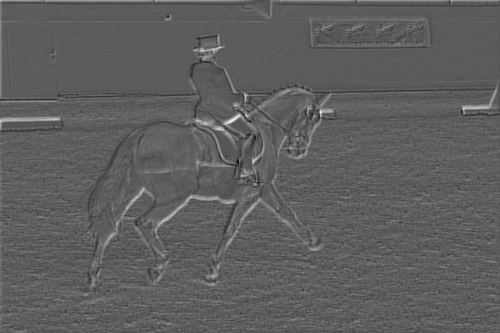}
       \AddImg{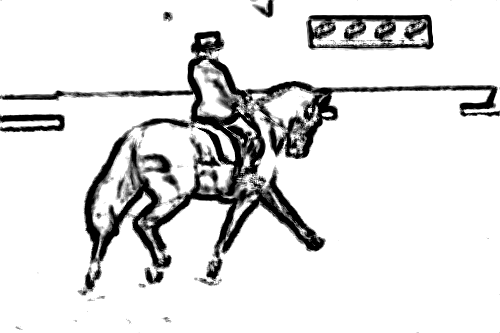}
     }{Side-1} \\ \vspace{0.02in}
     \stackunder[2pt]{
       \AddImg{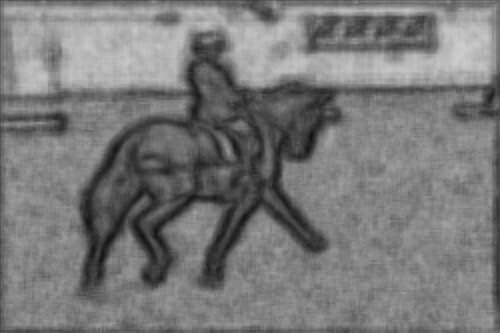}
       \AddImg{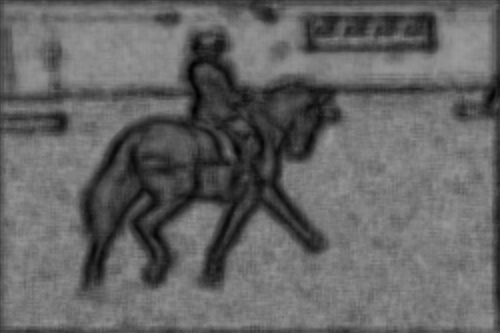}
       \AddImg{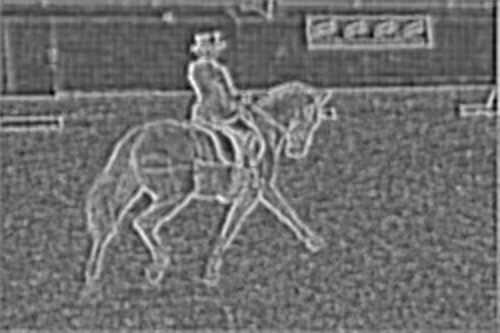}
       \AddImg{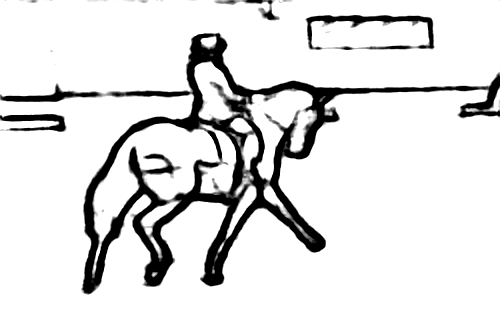}
     }{Side-2} \\ \vspace{0.02in}
     \stackunder[2pt]{
       \AddImg{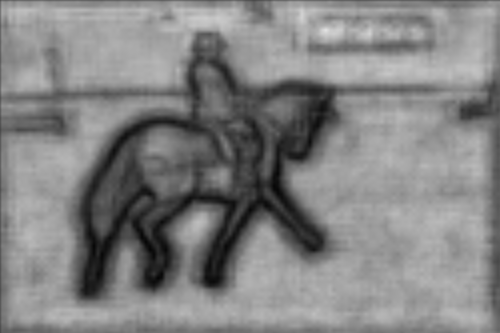}
       \AddImg{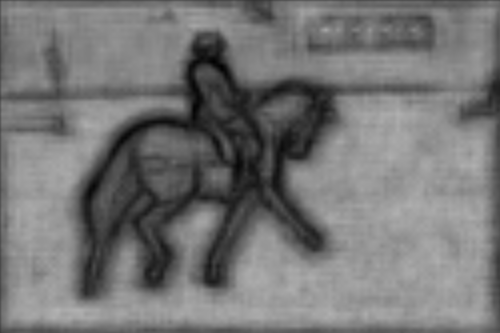}
       \AddImg{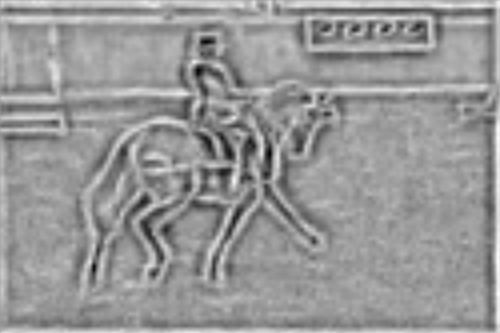}
       \AddImg{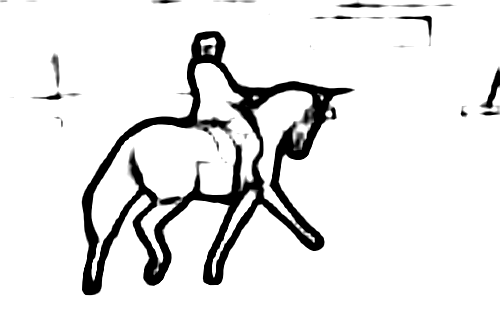}
     }{Side-3} \\ \vspace{0.02in}
     \stackunder[2pt]{
       \AddImg{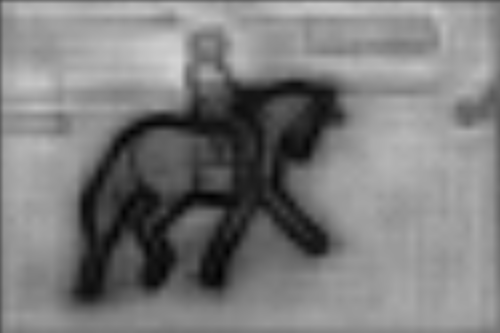}
       \AddImg{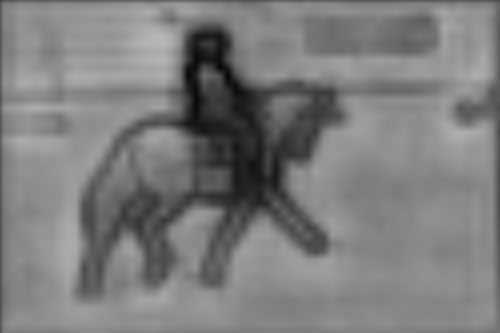}
     }{Side-4} \hspace{-0.13in}
     \stackunder[2pt]{
       \AddImg{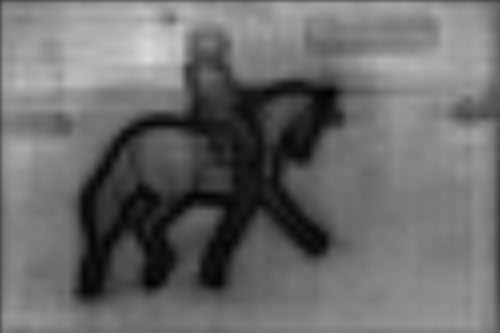}
       \AddImg{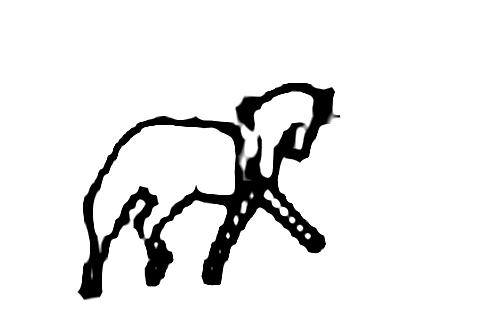}
     }{Side5-Horse} \\ \vspace{0.02in}
     \stackunder[2pt]{
       \AddImg{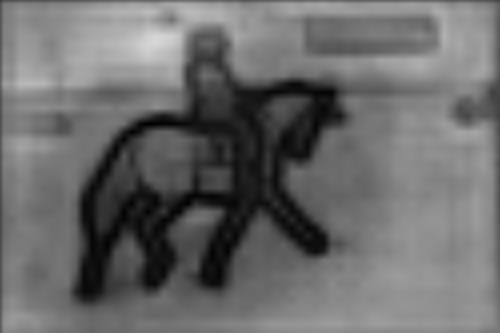}
       \AddImg{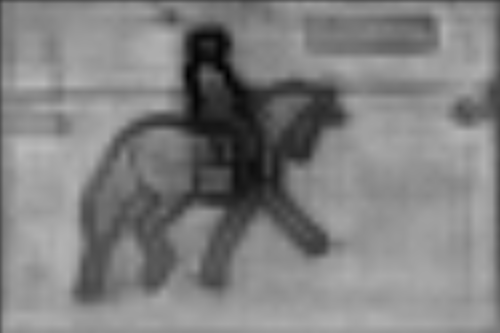}
     }{Side-5} \hspace{-0.13in}
     \stackunder[2pt]{
       \AddImg{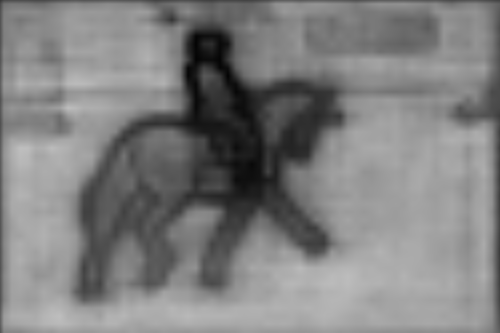}
       \AddImg{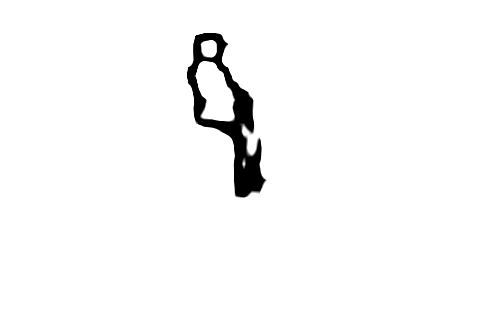}
     }{Side5-Person} \\
    \caption{Side activation maps on the input image of 
    \figref{fig:sbd_samples}.
    The first two columns display DSN's side class classification 
    activation for the classes of horse and person, respectively.
    The last two columns show the side features of Side-1 $\sim$ 
    Side-3 and class classification activation of Side-5 for 
    CASENet and our DDS-R, respectively.
    These images are obtained by normalizing the activation to $[0, 255]$.
    Note that all activations are directly outputted without any 
    non-linearization, \eg, \textit{sigmoid} function.
    } \label{fig:sbd_activation}
\end{figure}

\renewcommand{\AddImg}[1]{\fbox{\includegraphics[width=.189\linewidth,height=.123\linewidth]{#1}}}
\newcommand{\white}[1]{{\textcolor{white}{#1}}}
\begin{figure*}[!ht]
    \centering
    \small 
    \renewcommand{\tabcolsep}{2.75mm}
    \hspace{0.09in}
    \begin{tabular}{cccccccccc}
      \cellcolor[rgb]{0.5,0,0} \white{aeroplane} &
      \cellcolor[rgb]{0,0.5,0} \white{bicycle} &
      \cellcolor[rgb]{0.5,0.5,0} \white{bird} &
      \cellcolor[rgb]{0,0,0.5} \white{boat} &
      \cellcolor[rgb]{0.5,0,0.5} \white{bottle} &
      \cellcolor[rgb]{0,0.5,0.5} \white{bus} &
      \cellcolor[rgb]{0.5,0.5,0.5} \white{car} & 
      \cellcolor[rgb]{0.25,0,0} \white{cat} &
      \cellcolor[rgb]{0.75,0,0} \white{chair} &
      \cellcolor[rgb]{0.25,0.5,0} \white{cow} \\
      \cellcolor[rgb]{0.75,0.5,0} \white{dining table} &
      \cellcolor[rgb]{0.25,0,0.5} \white{dog} &
      \cellcolor[rgb]{0.75,0,0.5} \white{horse} &
      \cellcolor[rgb]{0.25,0.5,0.5} \white{motorbike} &
      \cellcolor[rgb]{0.75,0.5,0.5} \white{person} &
      \cellcolor[rgb]{0,0.25,0} \white{potted plant} &
      \cellcolor[rgb]{0.5,0.25,0} \white{sheep} & 
      \cellcolor[rgb]{0,0.75,0} \white{sofa} &
      \cellcolor[rgb]{0.5,0.75,0} \white{train} &
      \cellcolor[rgb]{0,0.25,0.5} \white{tv monitor} \\
    \end{tabular}
    \\ \vspace{0.08in}
    \setlength{\fboxrule}{0.6pt}
    \setlength{\fboxsep}{0pt}
    \begin{sideways} \scriptsize \hspace{0.02in} Original Images \end{sideways}
    \hspace{-0.062in}
    \AddImg{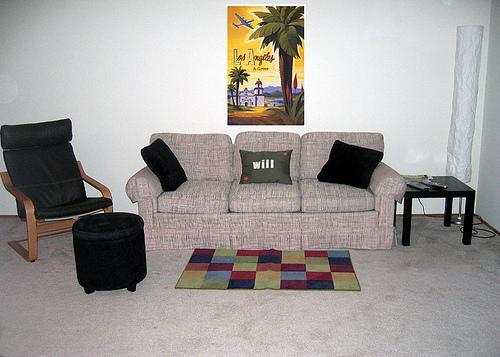} \AddImg{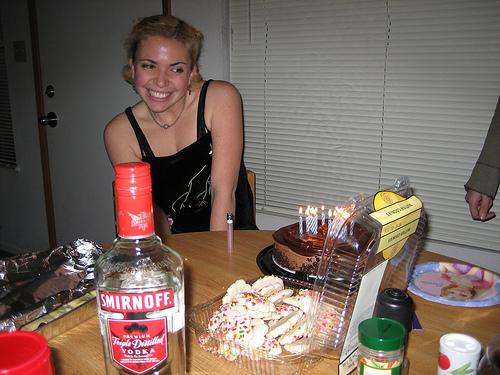} \AddImg{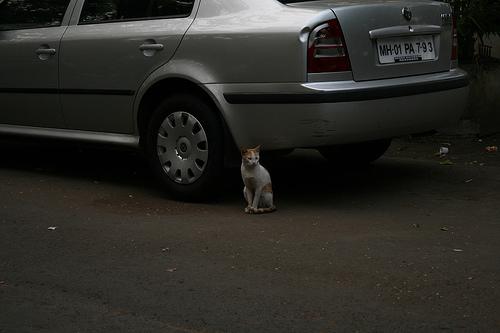}
    \AddImg{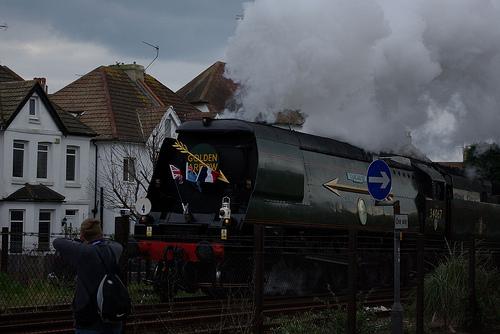} \AddImg{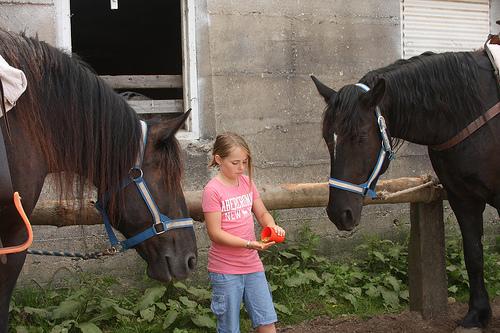}
    \\ \vspace{0.02in}
    \begin{sideways} \scriptsize \hspace{0.06in} Ground Truth \end{sideways}
    \AddImg{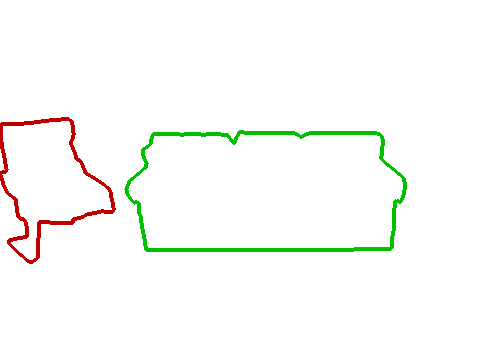} \AddImg{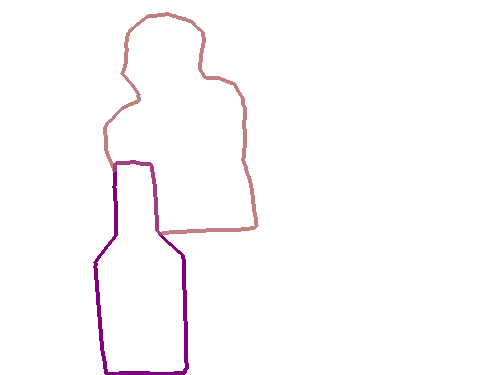}
    \AddImg{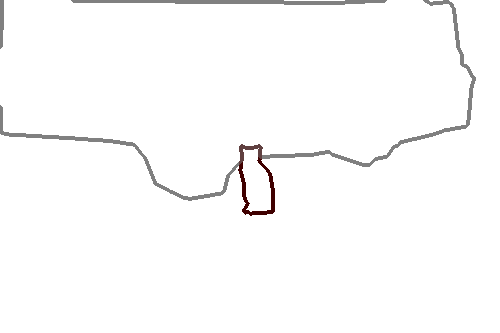} \AddImg{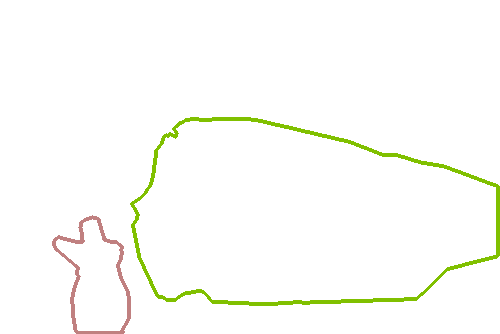}
    \AddImg{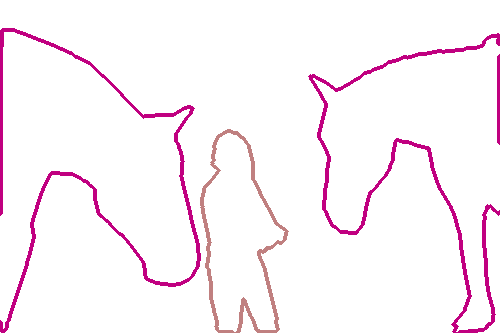}
    \\ \vspace{0.02in}
    \begin{sideways} \scriptsize \hspace{0.30in} DSN \end{sideways}
    \AddImg{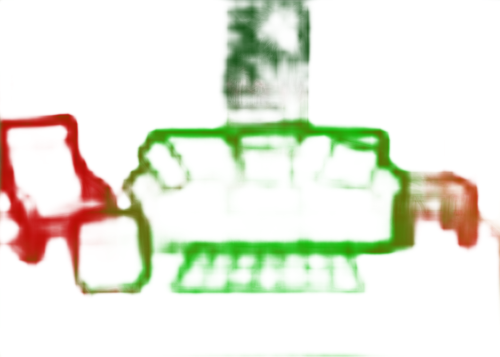} \AddImg{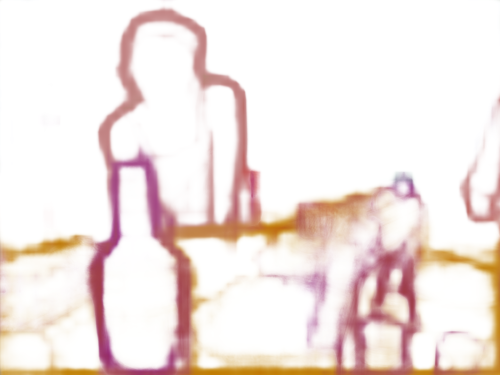}
    \AddImg{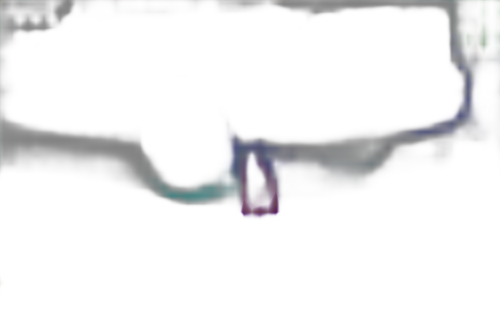} \AddImg{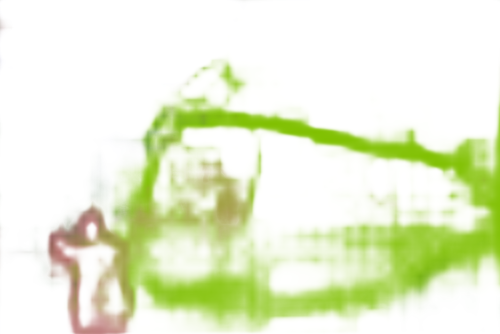}
    \AddImg{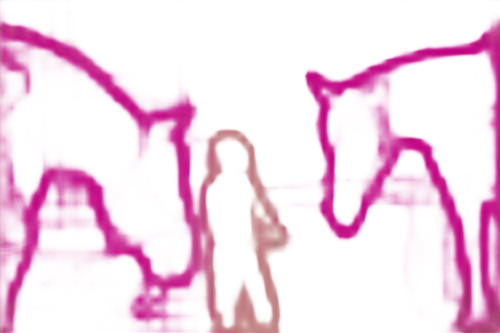}
    \\ \vspace{0.02in}
    \begin{sideways} \scriptsize \hspace{0.18in} CASENet \end{sideways}
    \AddImg{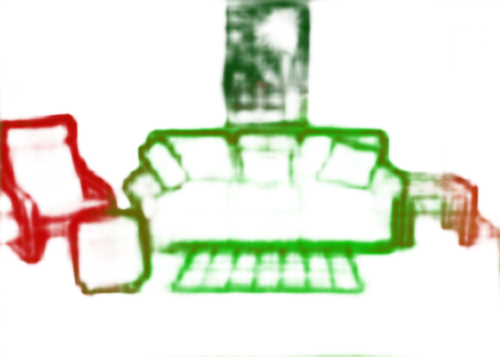} \AddImg{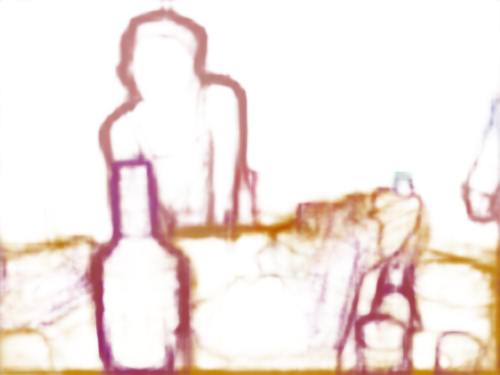}
    \AddImg{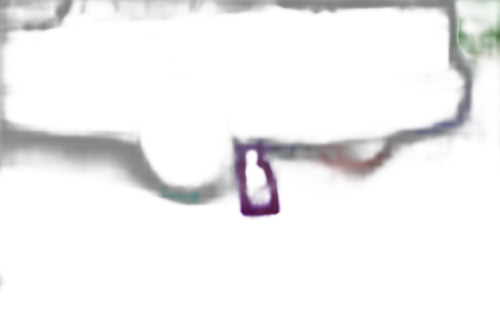} \AddImg{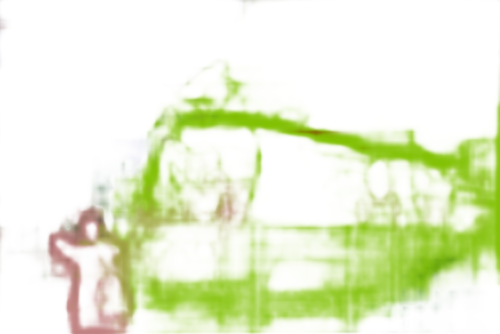}
    \AddImg{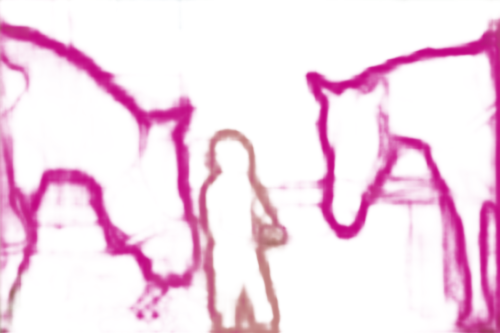}
    \\ \vspace{0.02in}
    \begin{sideways} \scriptsize \hspace{0.27in} SEAL \end{sideways}
    \AddImg{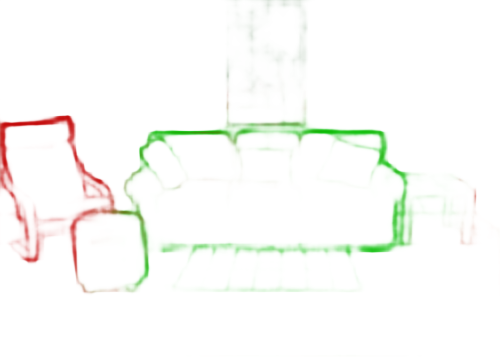} \AddImg{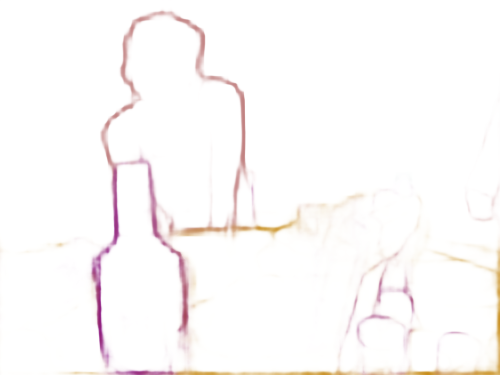}
    \AddImg{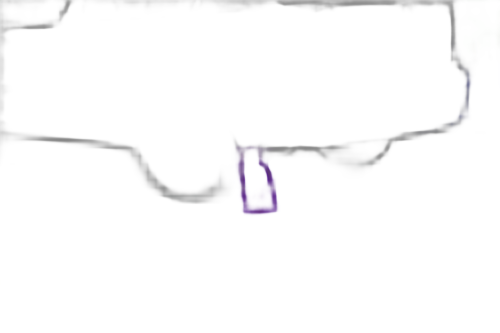} \AddImg{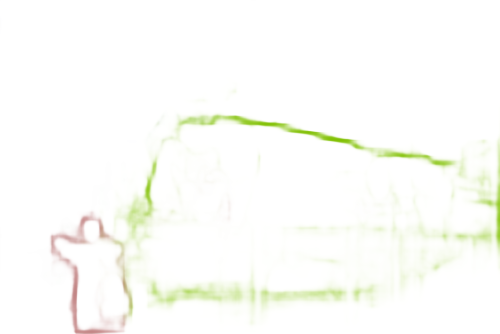}
    \AddImg{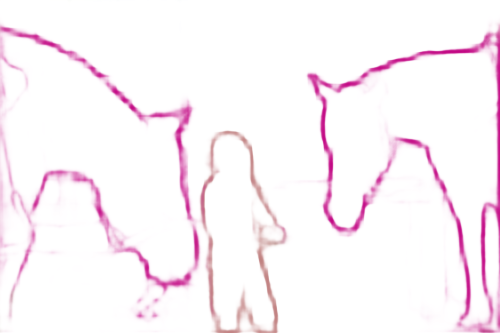}
    \\ \vspace{0.02in}
    \begin{sideways} \scriptsize \hspace{0.24in} STEAL \end{sideways}
    \AddImg{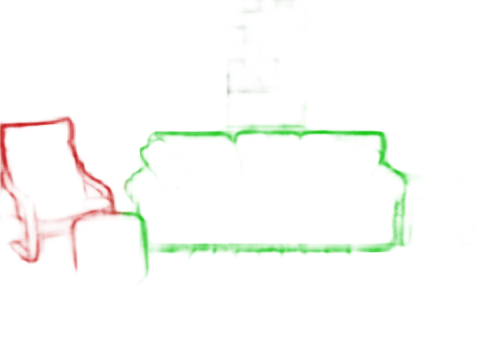} \AddImg{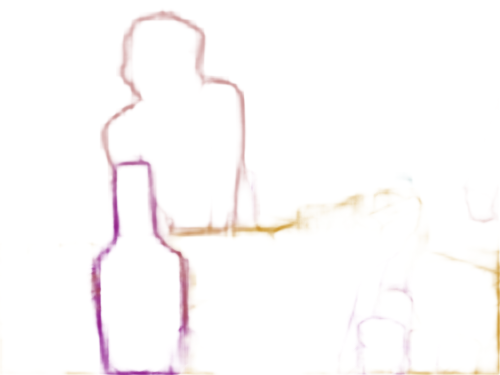}
    \AddImg{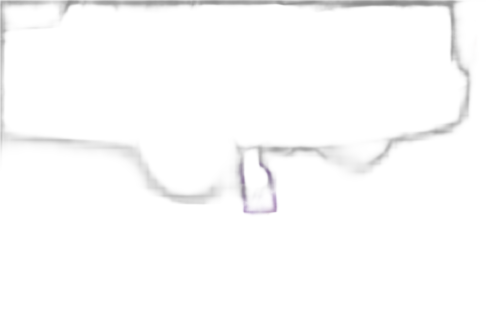} \AddImg{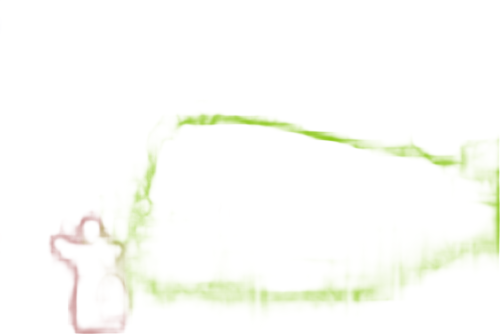}
    \AddImg{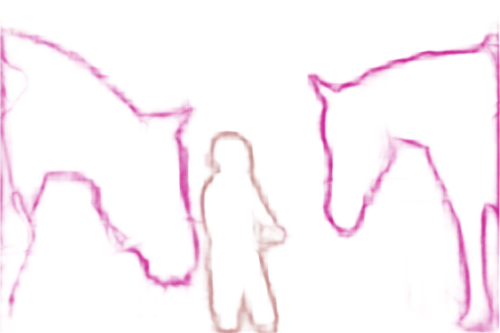}
    \\ \vspace{0.02in}
    \begin{sideways} \scriptsize \hspace{0.30in} DFF \end{sideways}
    \AddImg{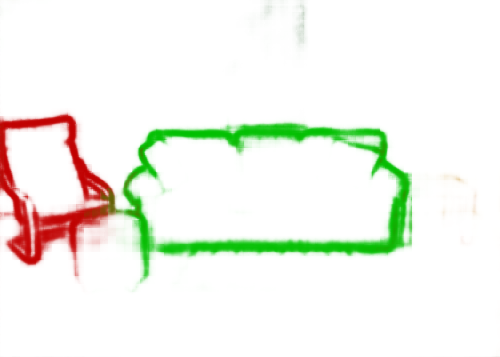} \AddImg{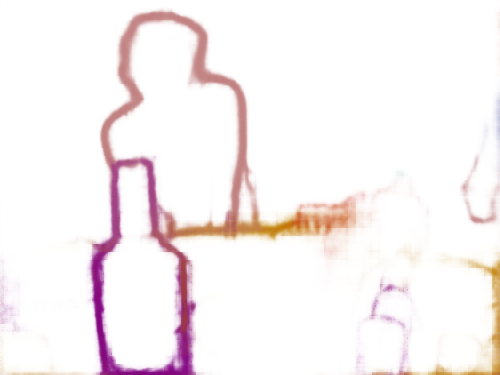}
    \AddImg{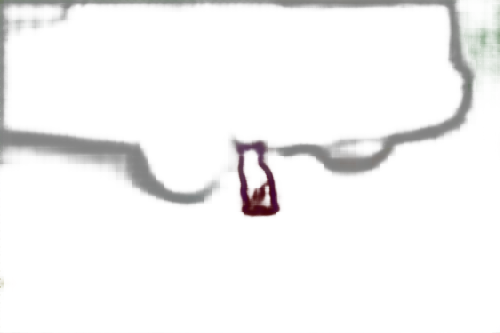} \AddImg{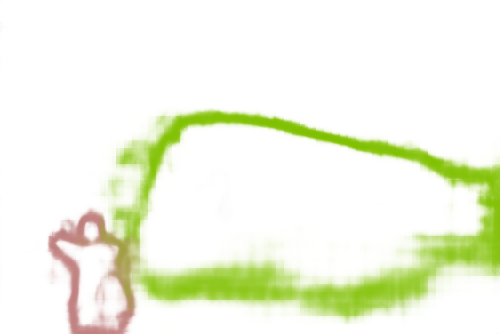}
    \AddImg{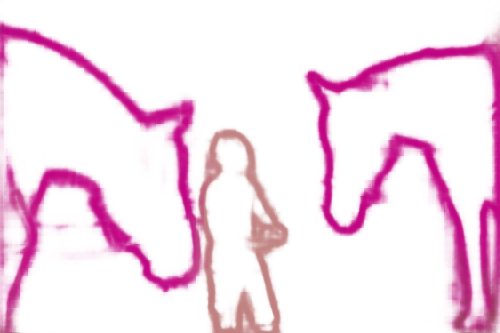}
    \\ \vspace{0.02in}
    \begin{sideways} \scriptsize \hspace{0.24in} DDS-R \end{sideways}
    \AddImg{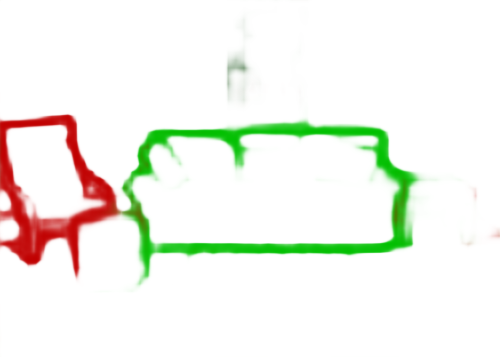} \AddImg{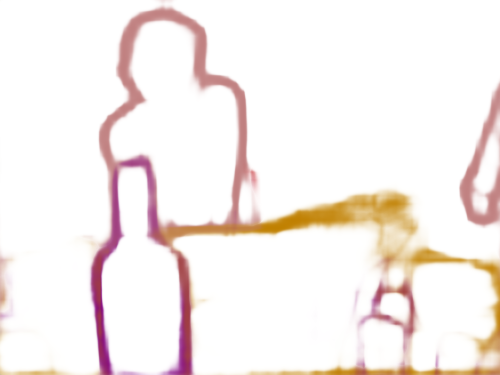}
    \AddImg{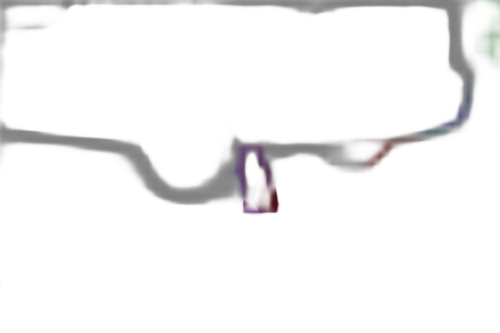} \AddImg{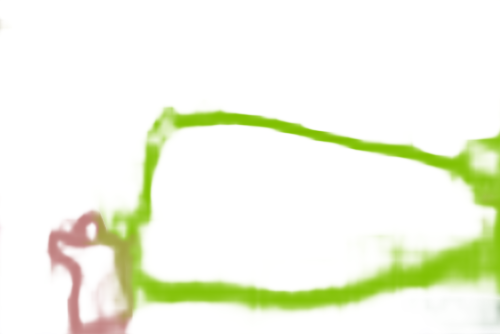}
    \AddImg{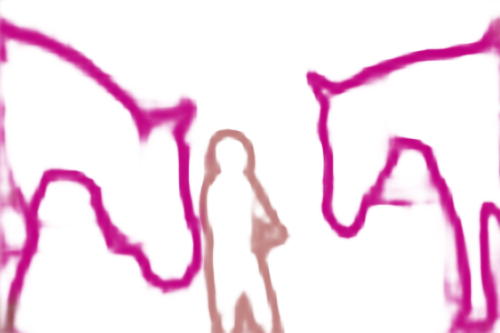}
    \\ \vspace{0.02in}
    \begin{sideways} \scriptsize \hspace{0.24in} DDS-U \end{sideways}
    \AddImg{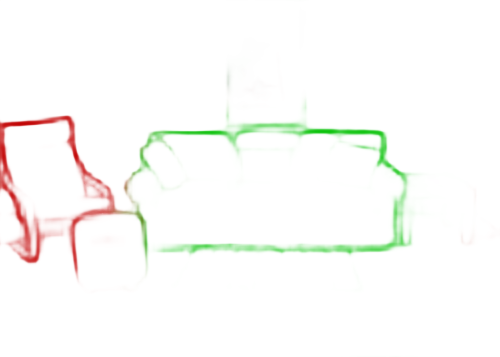} \AddImg{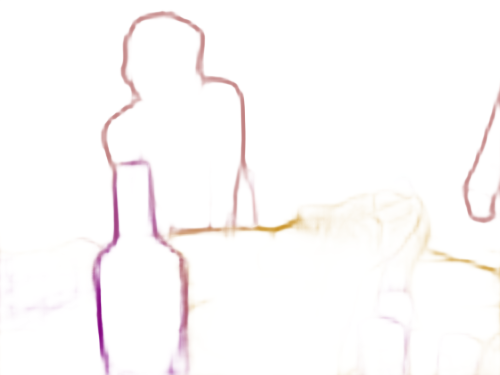}
    \AddImg{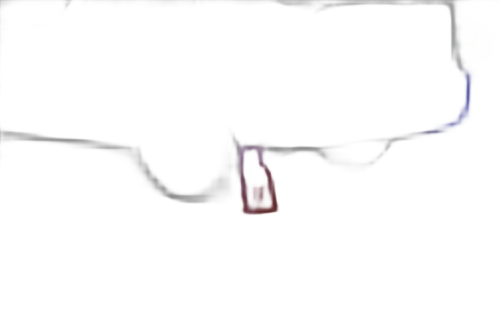} \AddImg{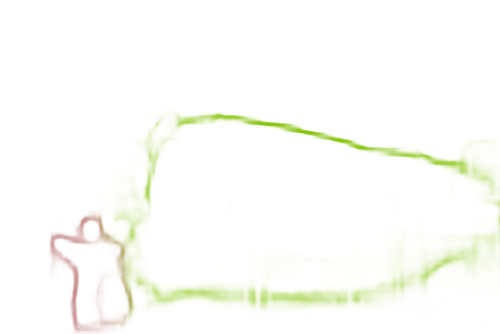}
    \AddImg{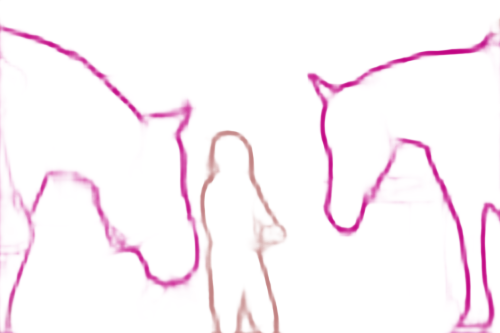}
    \\ 
    \caption{\revise{Some examples from SBD dataset \cite{hariharan2011semantic}.
    \textbf{From top to bottom:} color codes, original images, 
    ground truth, DSN, CASENet \cite{yu2017casenet}, 
    SEAL \cite{yu2018simultaneous}, STEAL \cite{acuna2019devil}, 
    DFF \cite{hu2019dynamic}, our DDS-R and DDS-U. 
    We follow the color coding protocol in \cite{yu2018simultaneous}.}}
    \label{fig:sbd_samps}
\end{figure*}

\renewcommand{\AddImg}[1]{\fbox{\includegraphics[width=.314\linewidth,height=.140\linewidth]{#1}}}

\begin{figure*}[!ht]
    \centering
    \small
    \renewcommand{\tabcolsep}{3.08mm}
    \hspace{0.08in}
    \begin{tabular}{cccccccccc}
      \cellcolor[rgb]{0.500,0.750,0.500} {road} &
      \cellcolor[rgb]{0.043,0.863,0.090} {sidewalk} &
      \cellcolor[rgb]{0.726,0.726,0.726} {building} &
      \cellcolor[rgb]{0.600,0.600,0.388} {wall} &
      \cellcolor[rgb]{0.255,0.400,0.400} \white{fence} &
      \cellcolor[rgb]{0.400,0.400,0.400} \white{pole} &
      \cellcolor[rgb]{0.020,0.333,0.882} \white{traffic light} & 
      \cellcolor[rgb]{0.137,0.137,1.000} \white{traffic sign} &
      \cellcolor[rgb]{0.580,0.443,0.863} {vegetation} &
      \cellcolor[rgb]{0.404,0.016,0.404} \white{terrain} \\
      \cellcolor[rgb]{0.726,0.490,0.294} {sky} &
      \cellcolor[rgb]{0.137,0.922,0.765} {person} &
      \cellcolor[rgb]{0.000,1.000,1.000} {rider} &
      \cellcolor[rgb]{1.000,1.000,0.443} {car} &
      \cellcolor[rgb]{1.000,1.000,0.726} {truck} &
      \cellcolor[rgb]{1.000,0.765,0.608} {bus} &
      \cellcolor[rgb]{1.000,0.686,0.608} {train} & 
      \cellcolor[rgb]{1.000,1.000,0.098} {motorcycle} &
      \cellcolor[rgb]{0.533,0.957,0.875} {bicycle} \\
    \end{tabular}
    \\ \vspace{0.02in}
    \setlength{\fboxrule}{0.6pt}
    \setlength{\fboxsep}{0pt}
    \begin{sideways} \footnotesize \hspace{0.035in} Original Images \end{sideways}
    \hspace{-0.062in}
    \AddImg{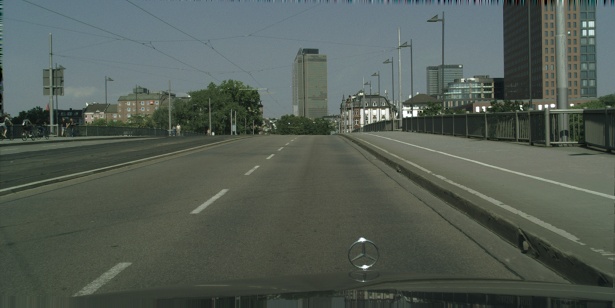}
    \AddImg{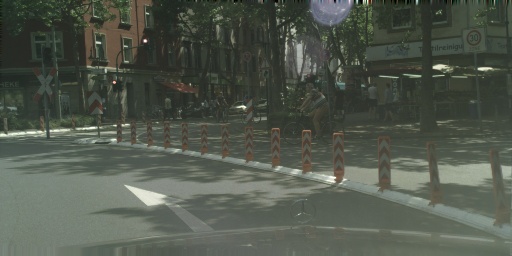}
    \AddImg{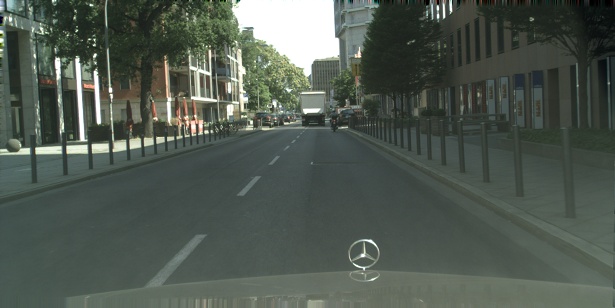}
    \\ \vspace{0.02in}
    \begin{sideways} \footnotesize \hspace{0.075in} Ground Truth \end{sideways}
    \AddImg{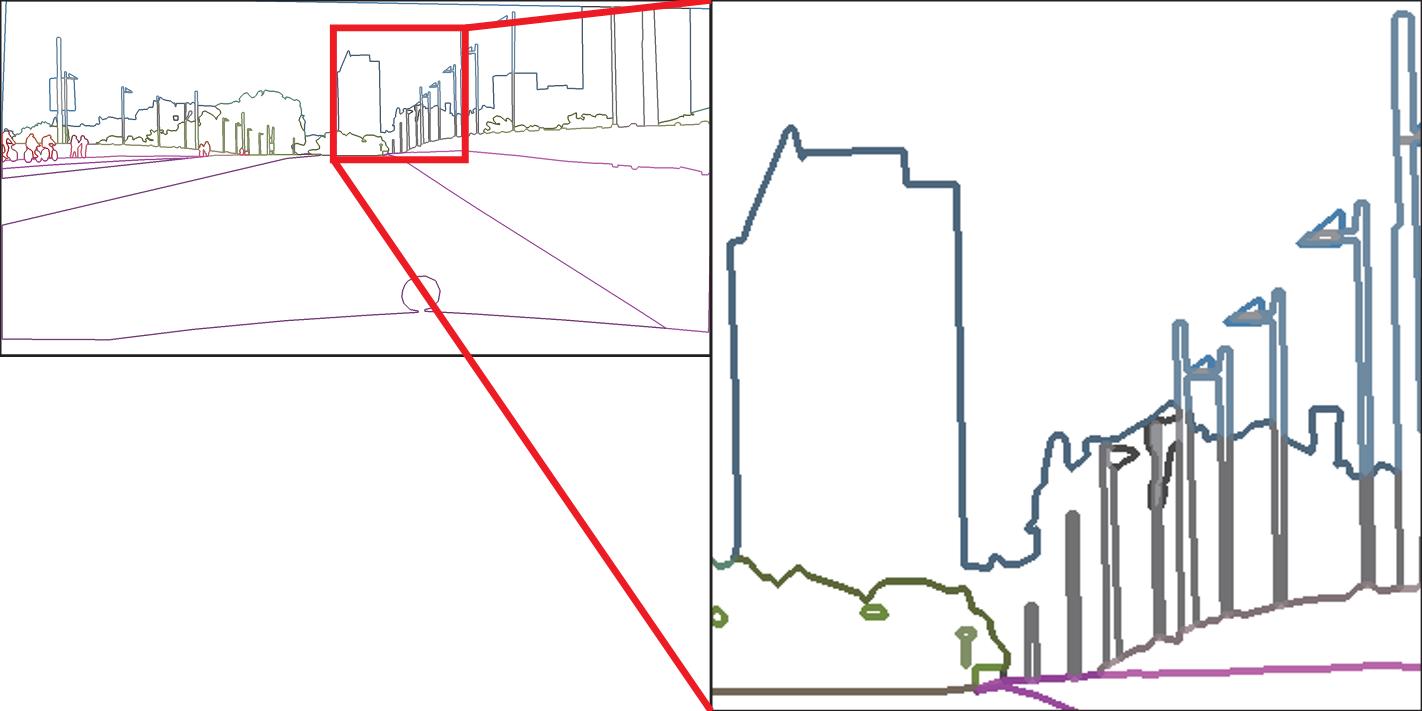}
    \AddImg{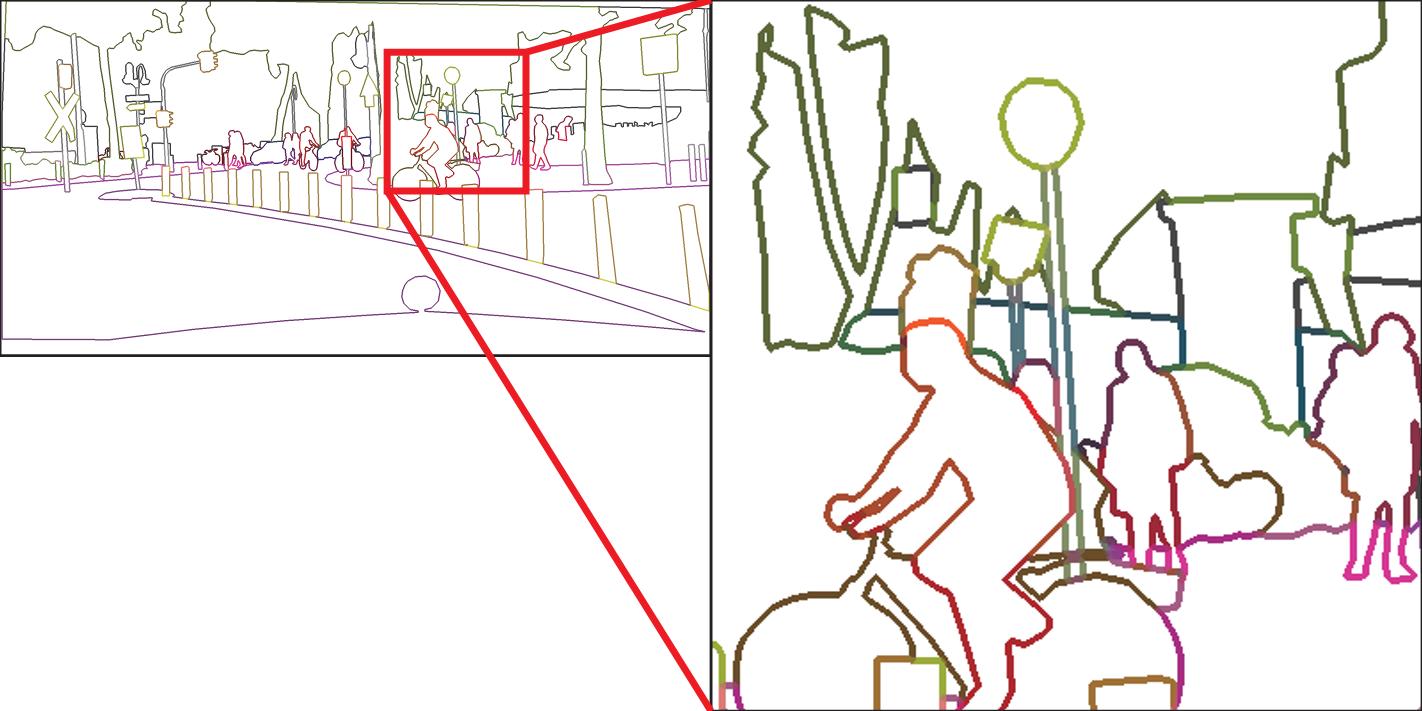}
    \AddImg{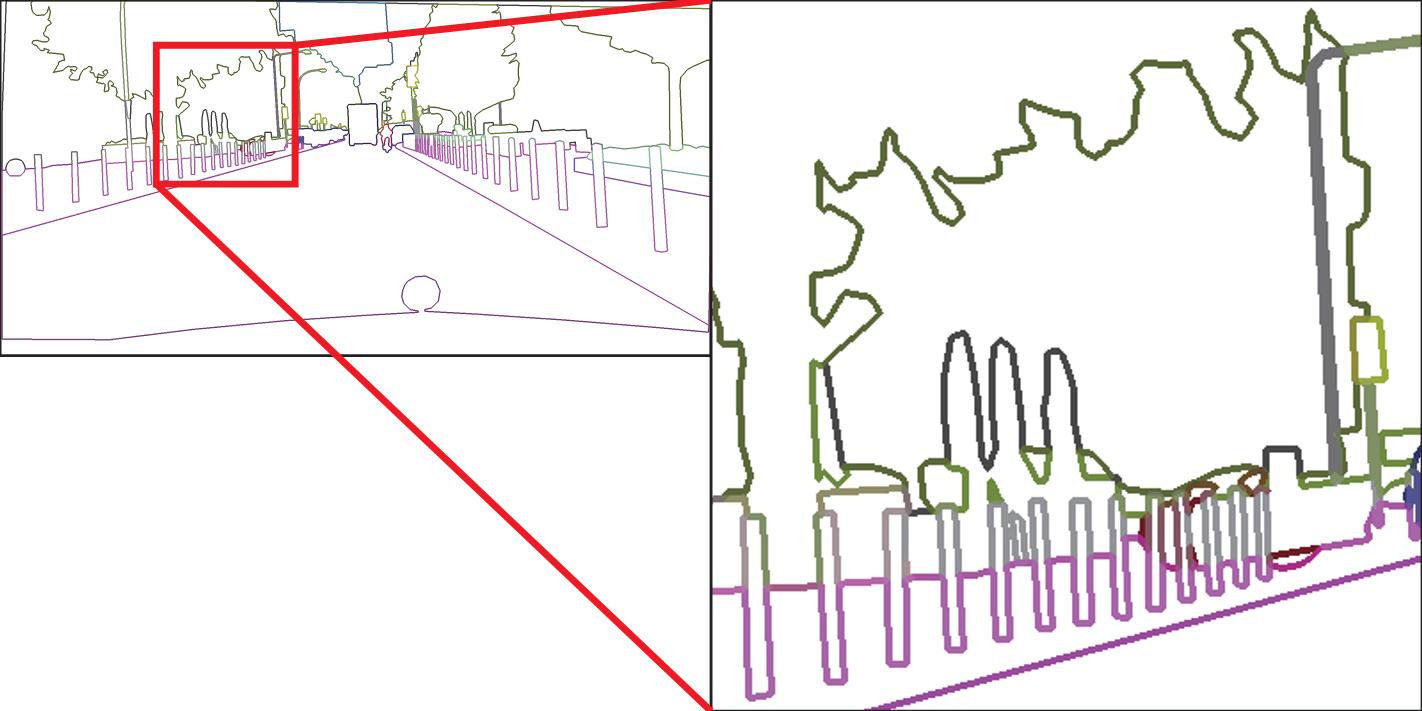}
    \\ \vspace{0.02in}
    \begin{sideways} \footnotesize \hspace{0.190in} CASENet \end{sideways}
    \AddImg{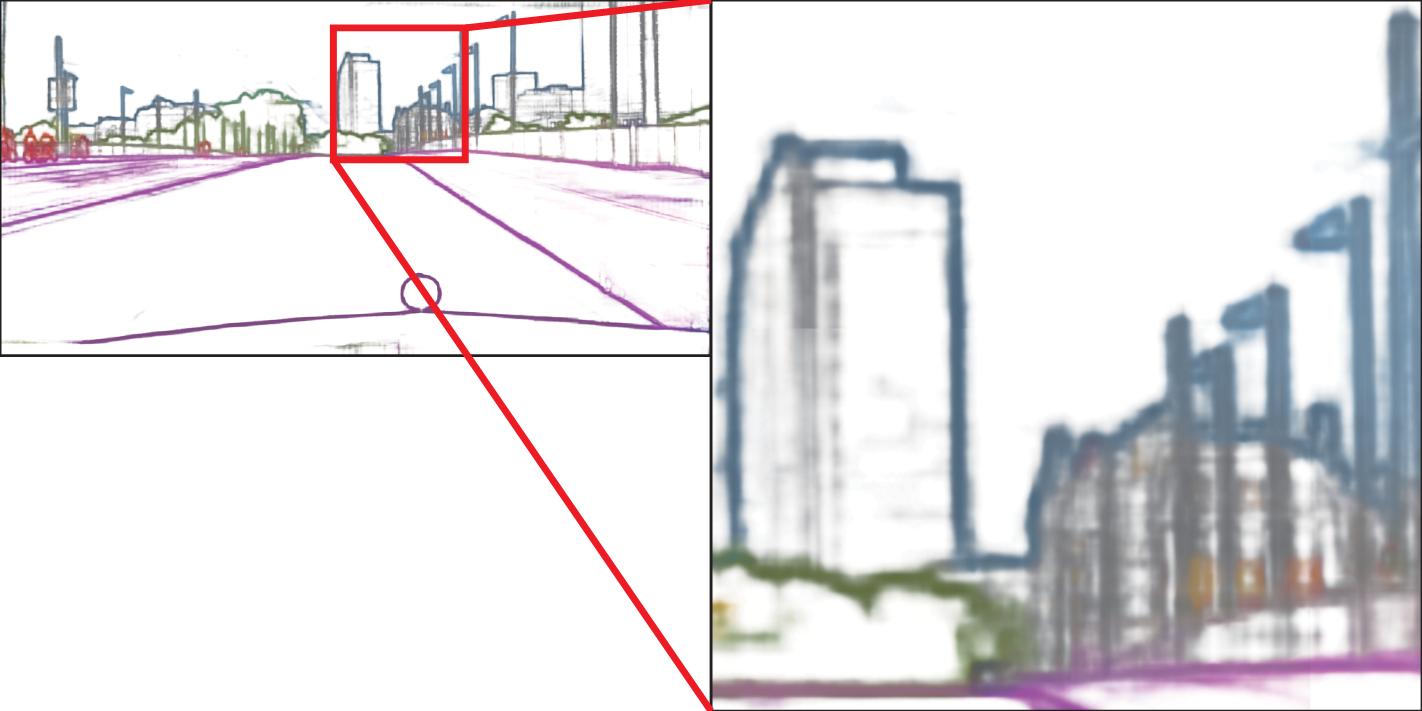}
    \AddImg{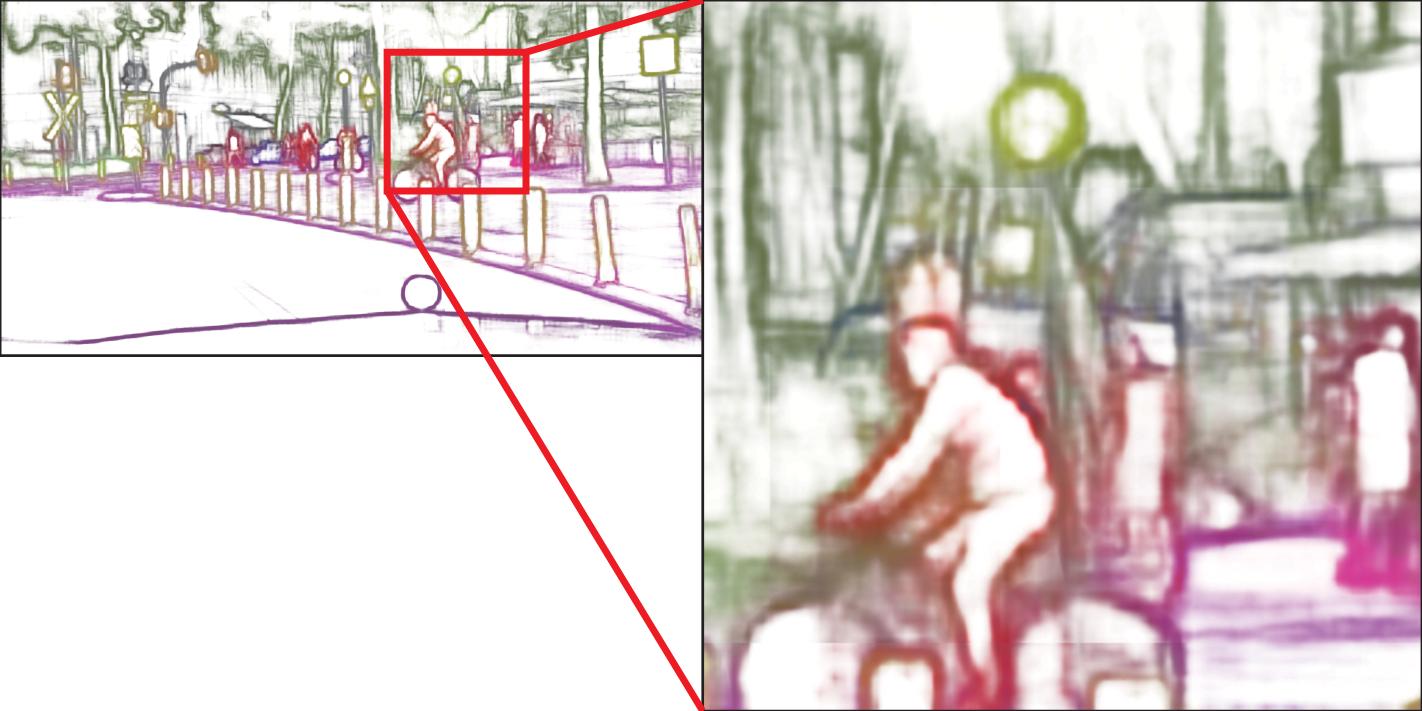}
    \AddImg{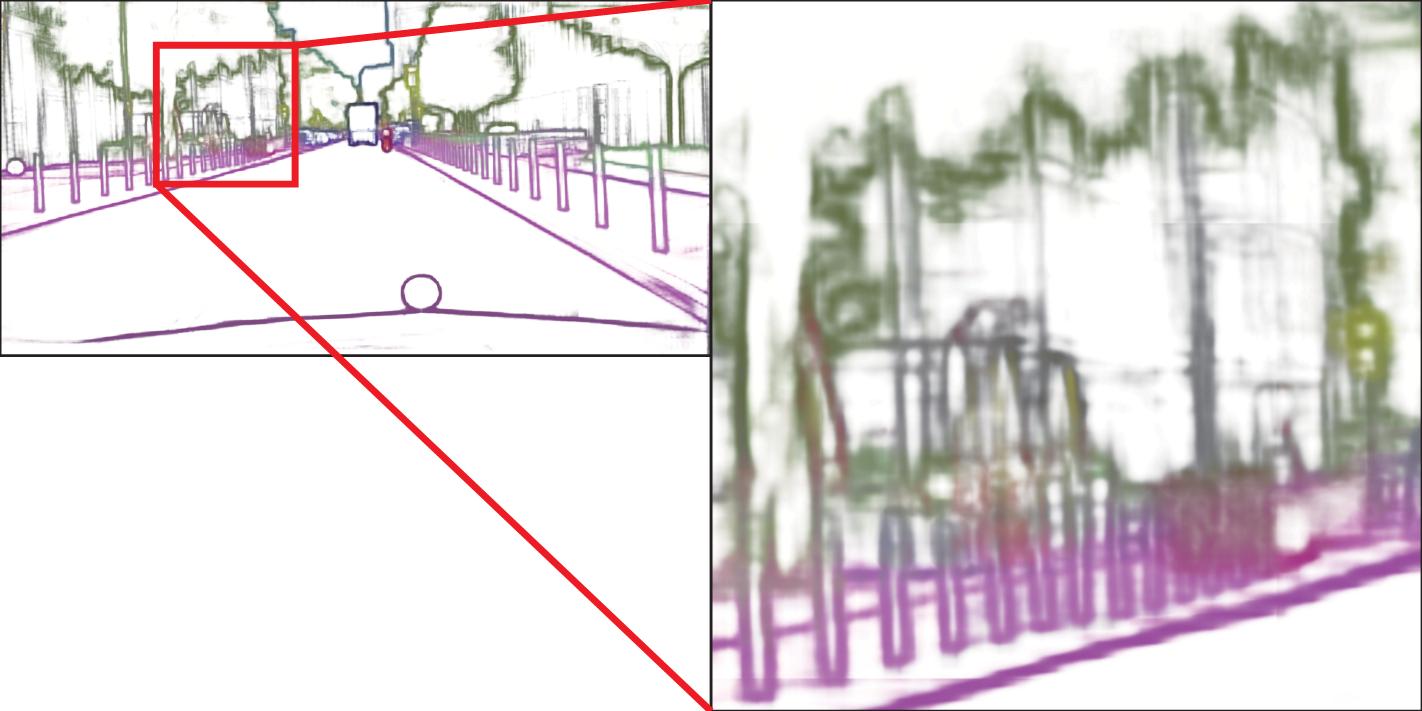}
    \\ \vspace{0.02in}
    \begin{sideways} \footnotesize \hspace{0.310in} SEAL \end{sideways}
    \AddImg{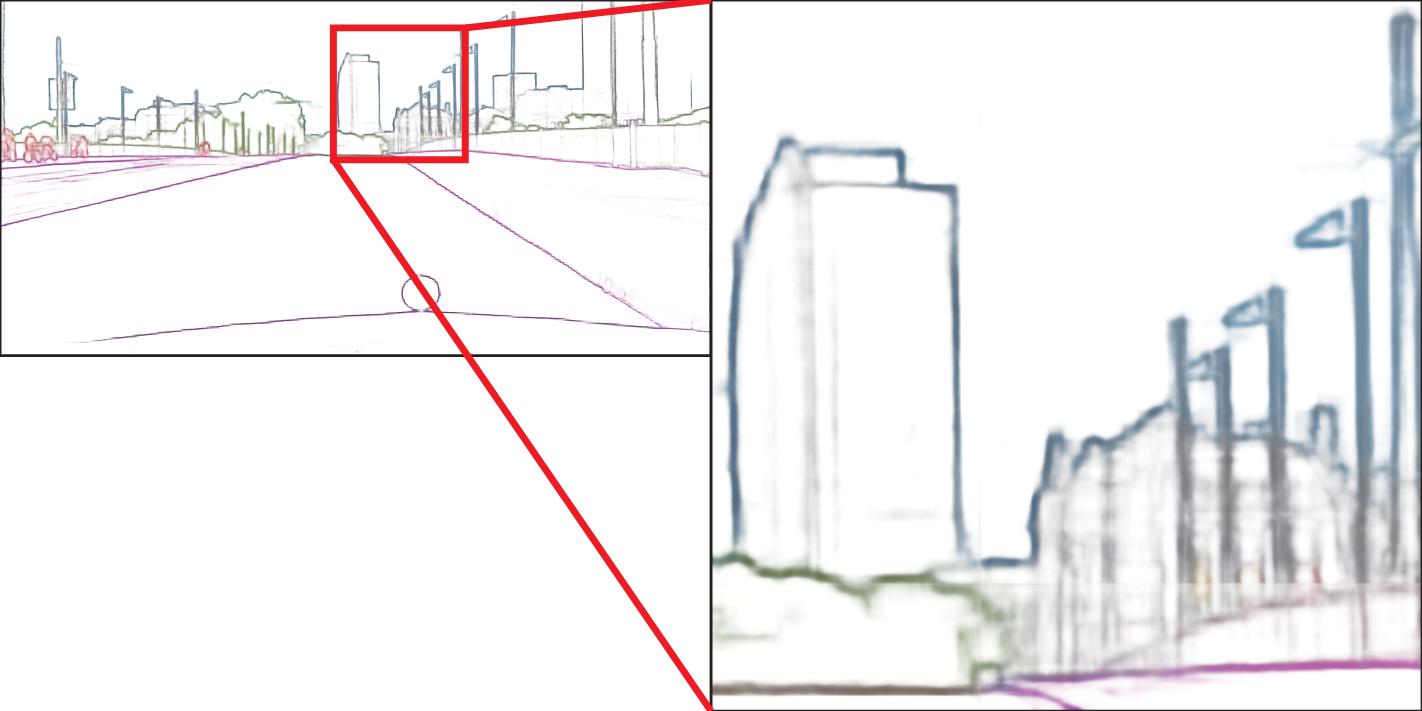}
    \AddImg{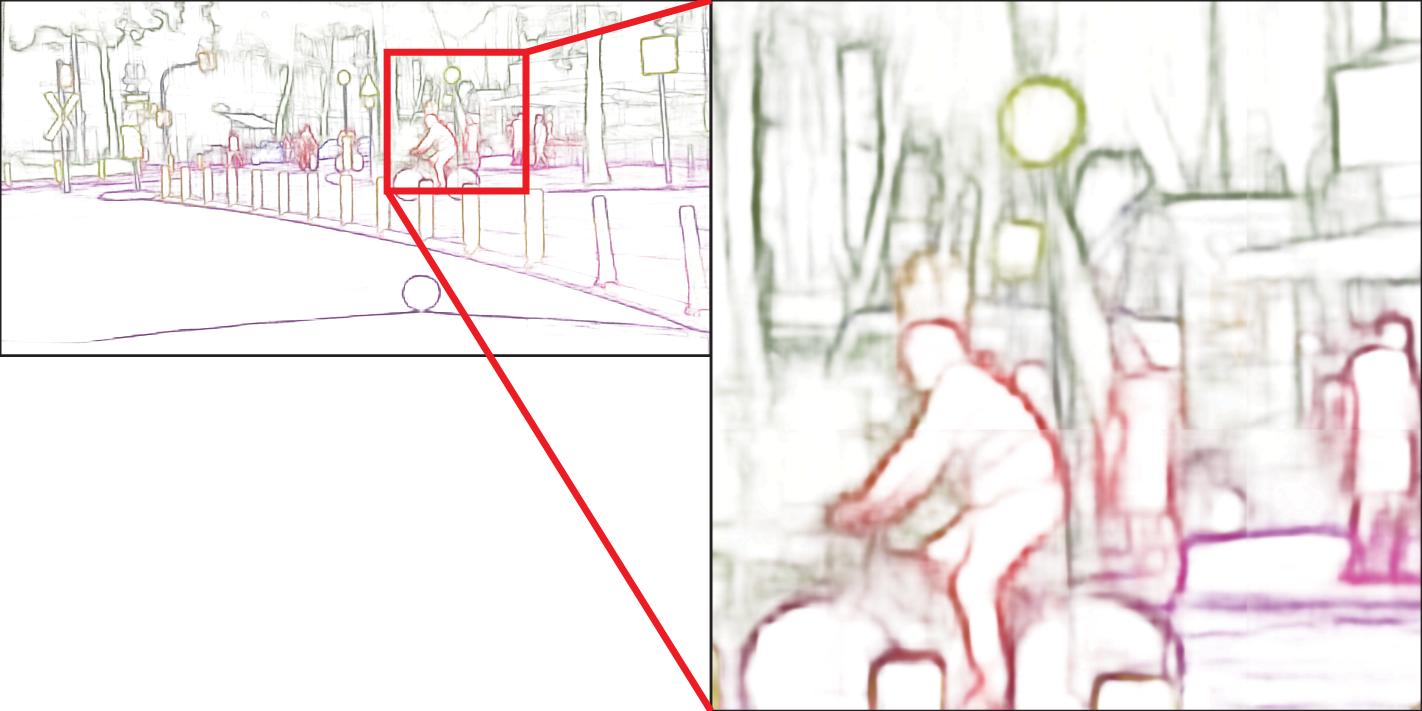}
    \AddImg{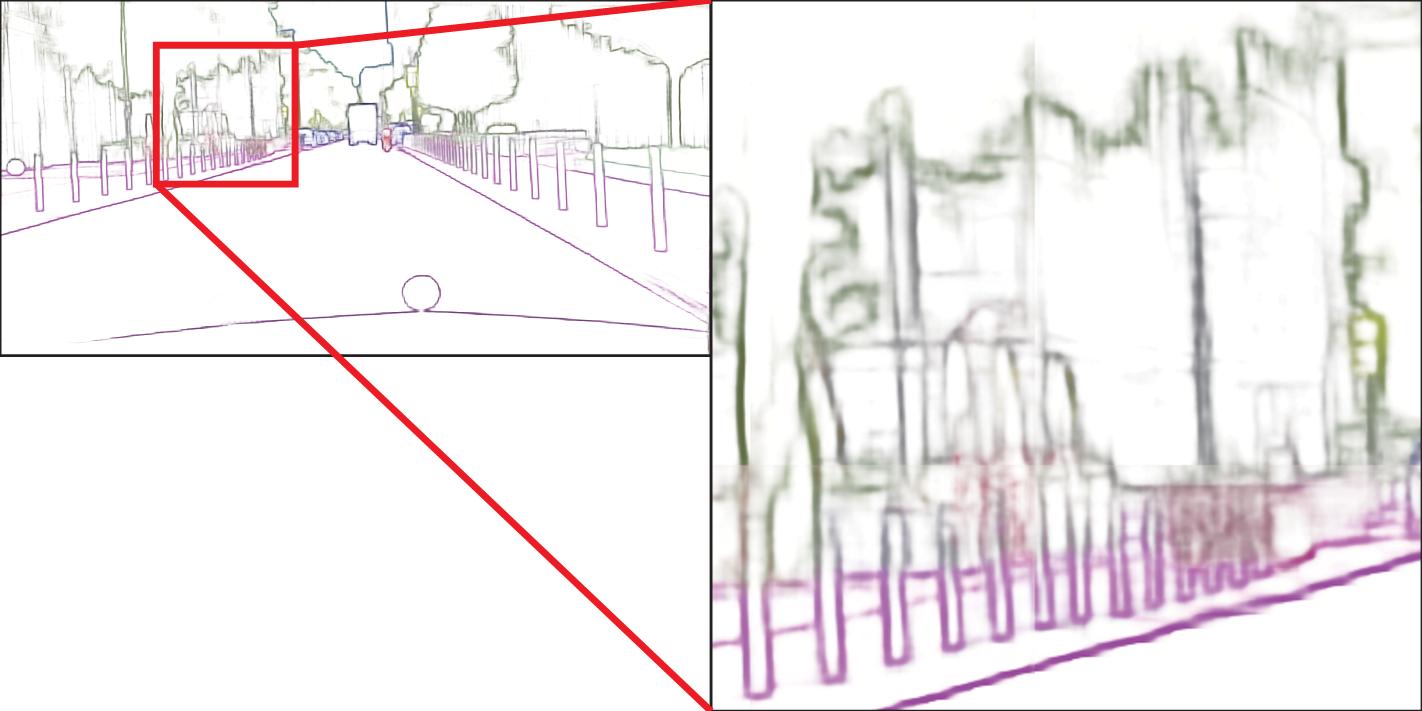}
    \\ \vspace{0.02in}
    \begin{sideways} \footnotesize \hspace{0.270in} STEAL \end{sideways}
    \AddImg{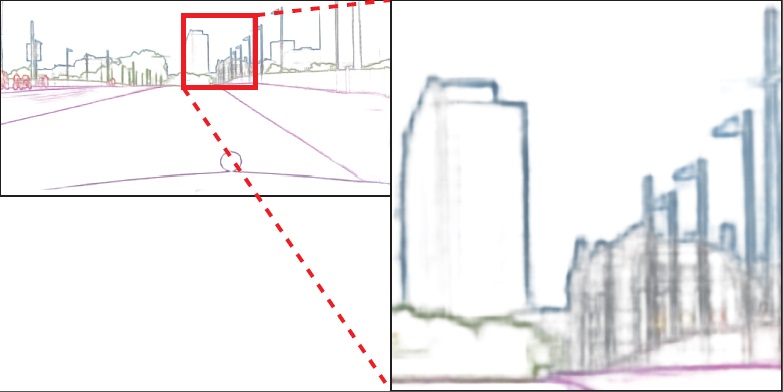}
    \AddImg{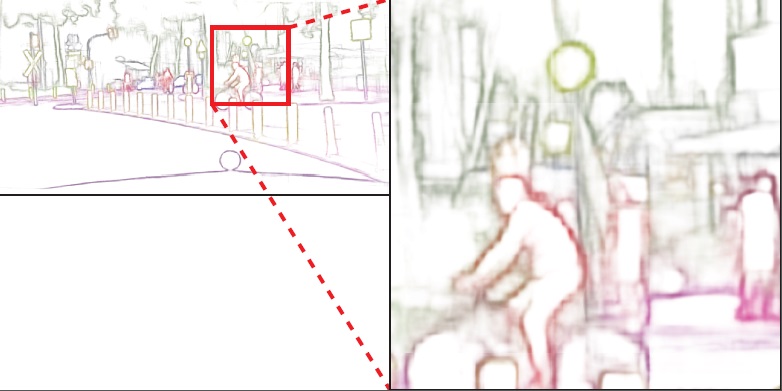}
    \AddImg{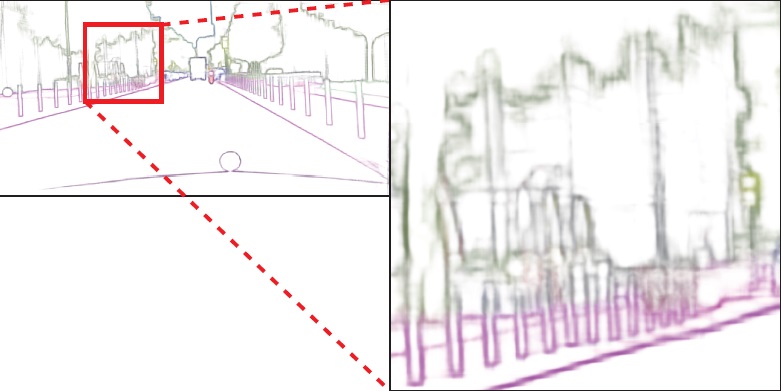}
    \\ \vspace{0.02in}
    \begin{sideways} \footnotesize \hspace{0.340in} DFF \end{sideways}
    \AddImg{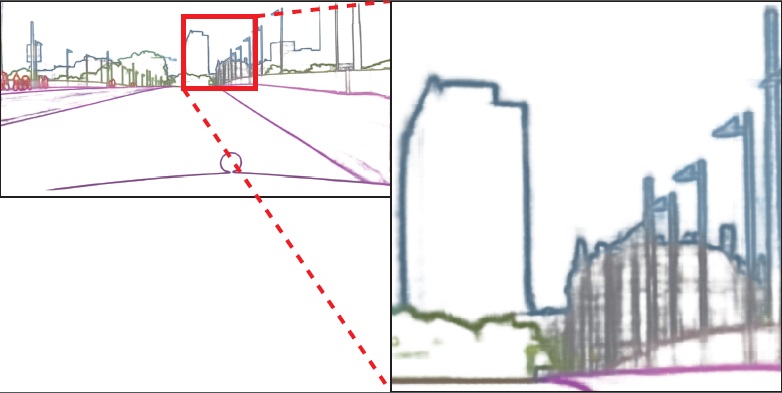}
    \AddImg{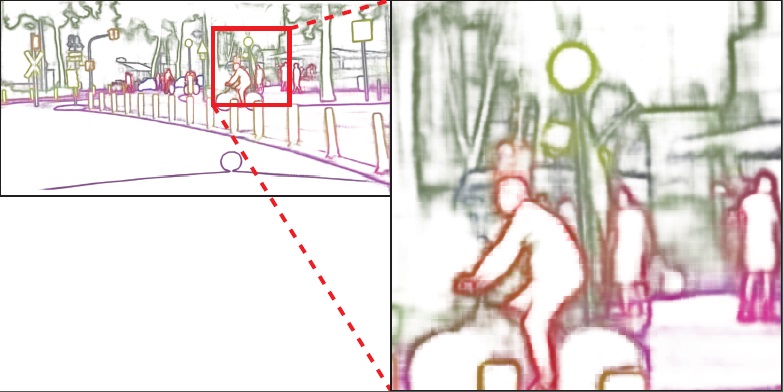}
    \AddImg{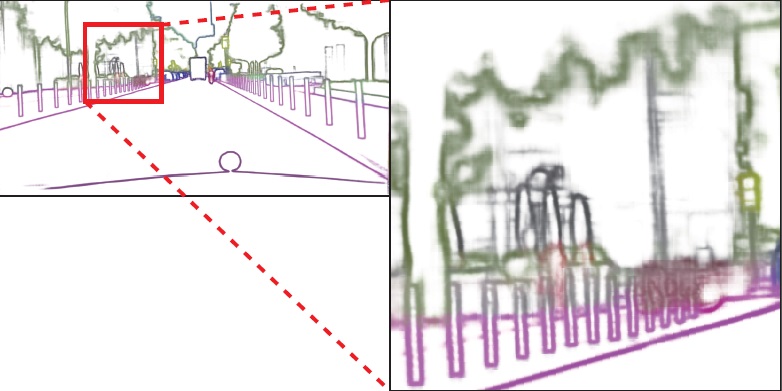}
    \\ \vspace{0.02in}
    \begin{sideways} \footnotesize \hspace{0.290in} DDS-R \end{sideways}
    \AddImg{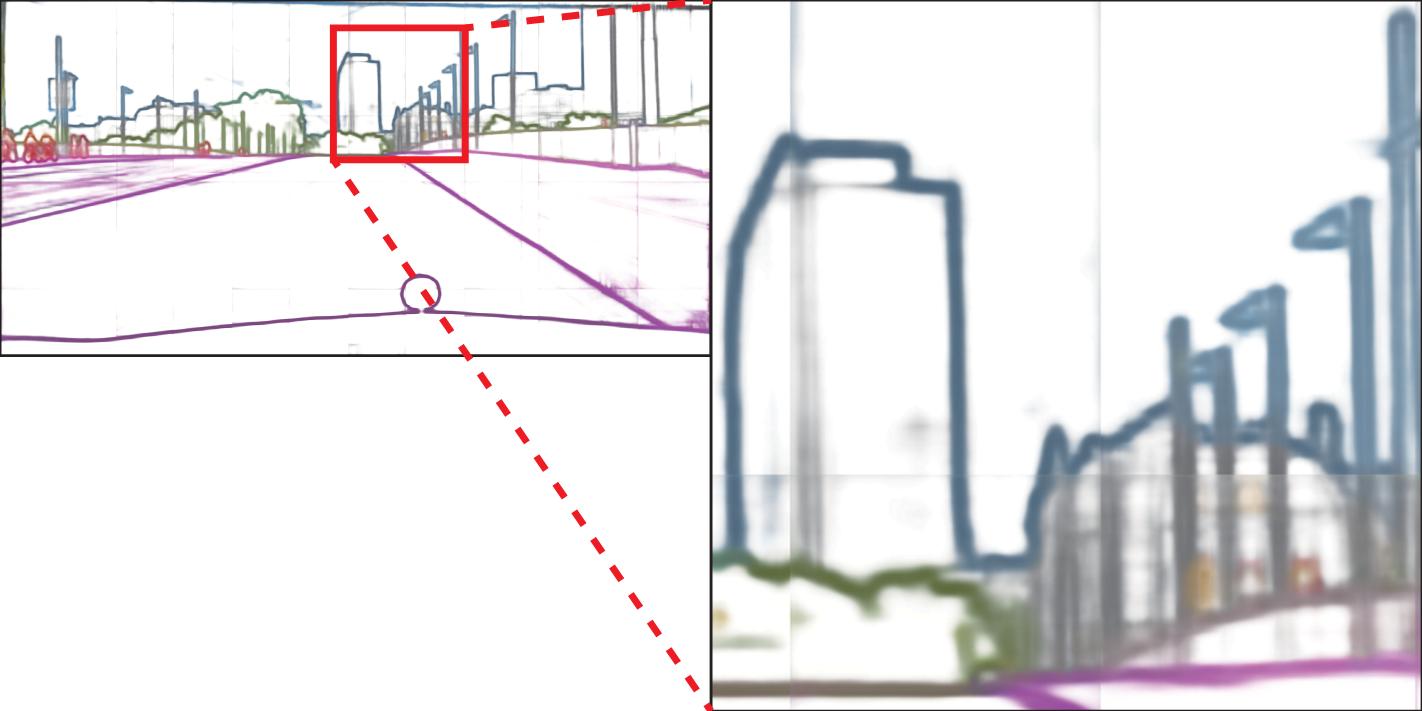}
    \AddImg{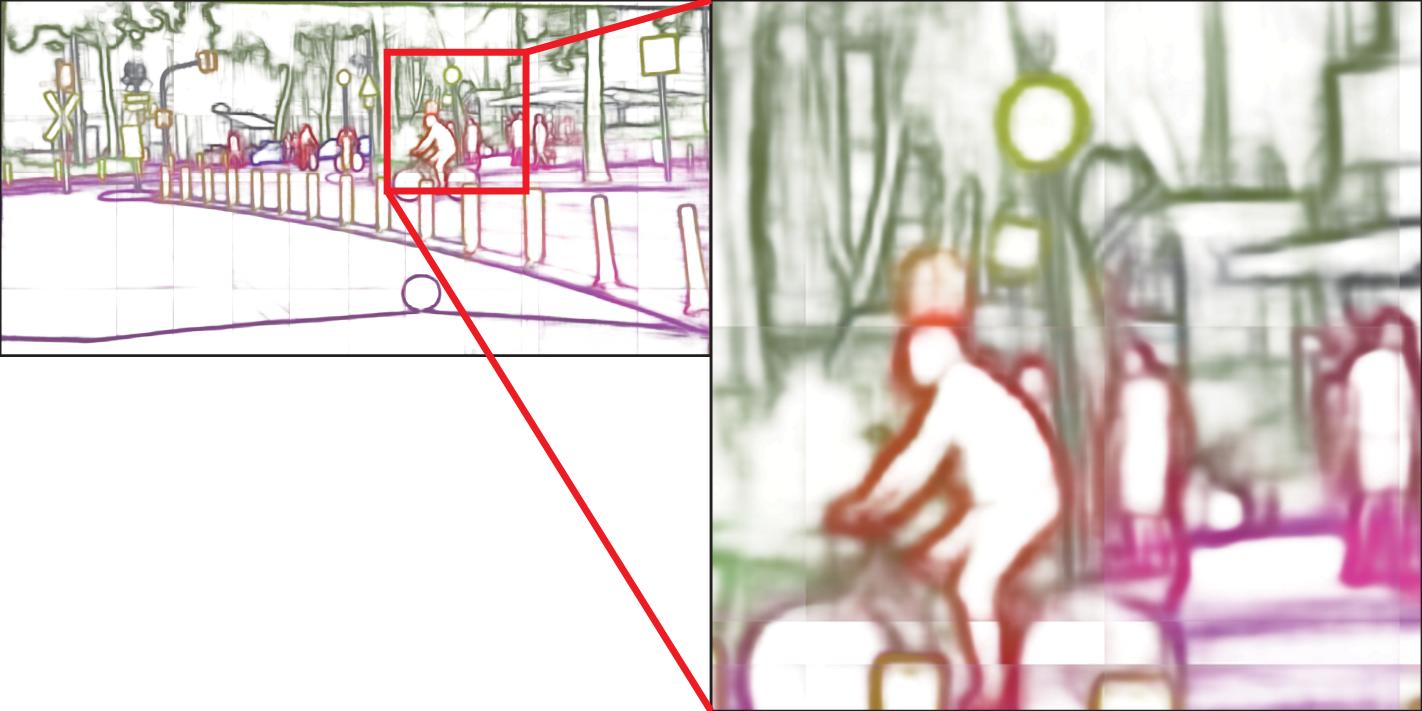}
    \AddImg{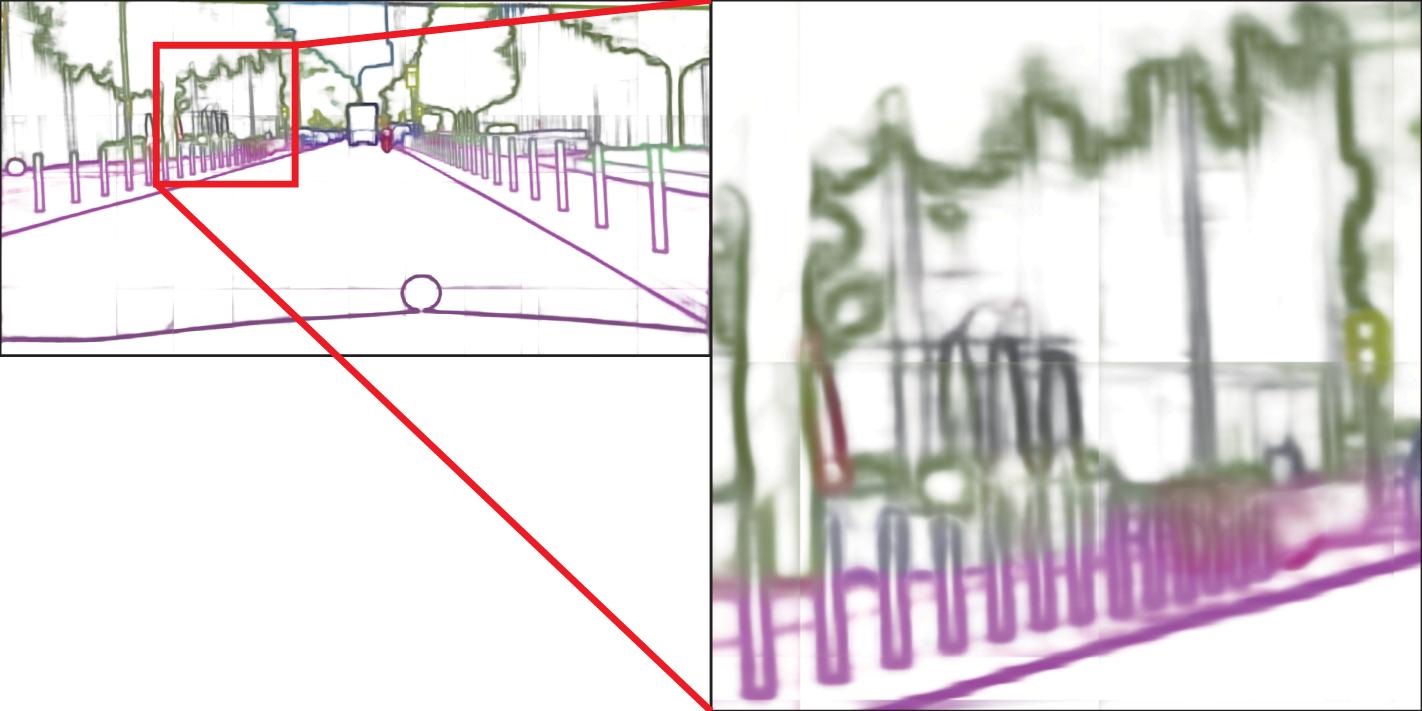}
    \\ \vspace{0.02in}
    \begin{sideways} \footnotesize \hspace{0.260in} DDS-U \end{sideways}
    \AddImg{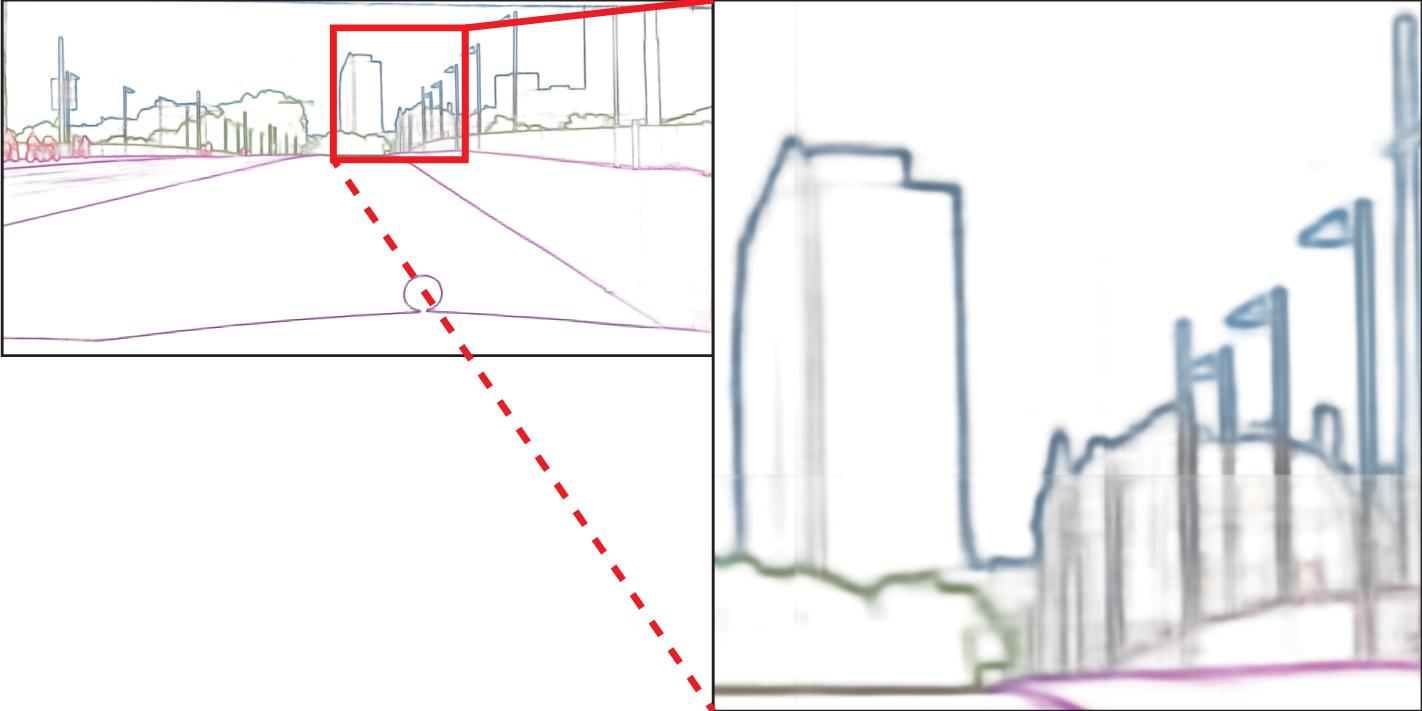}
    \AddImg{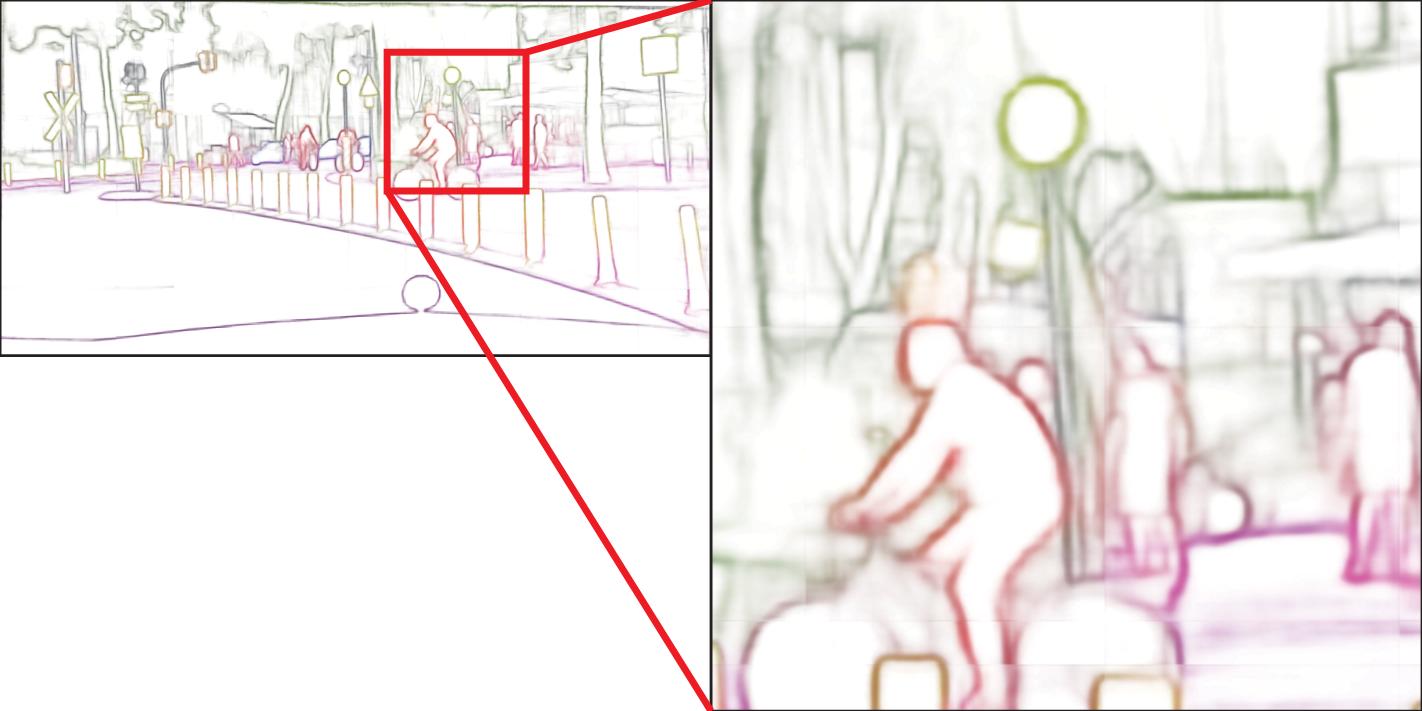}
    \AddImg{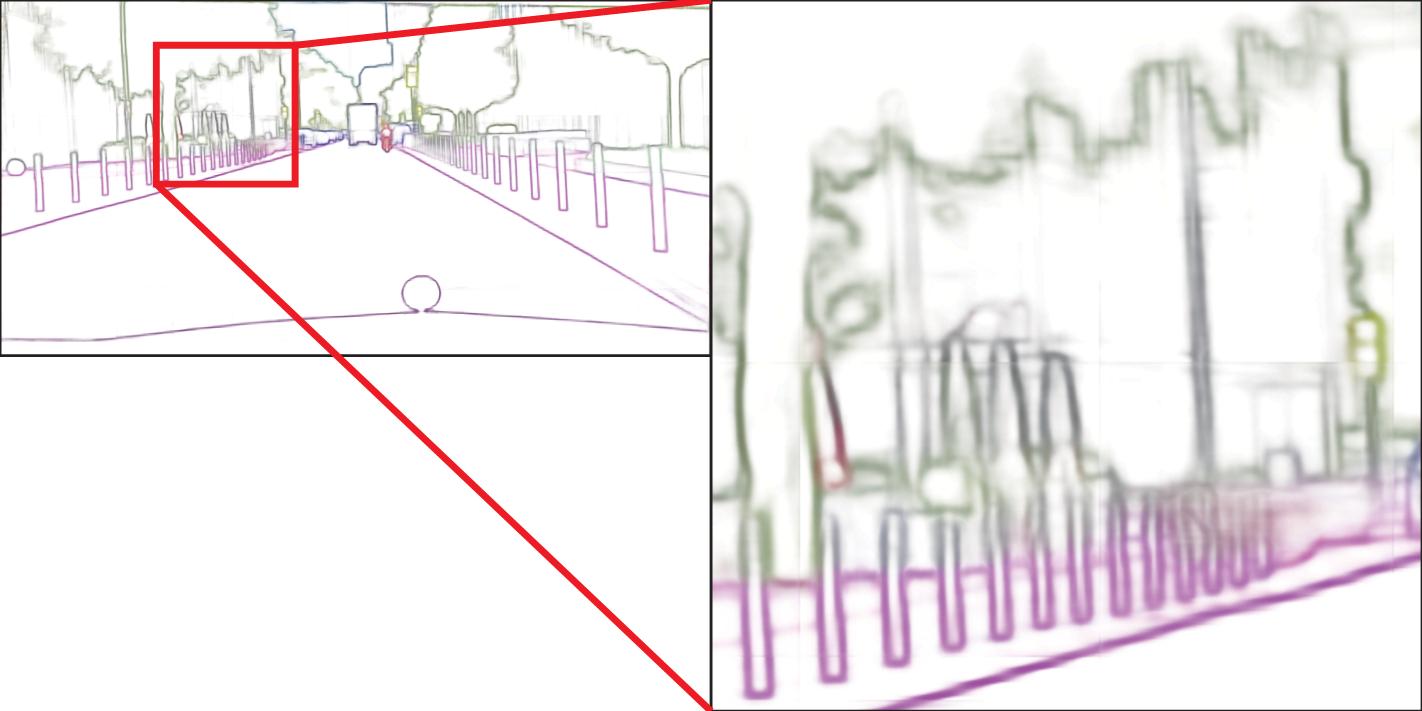}
    \\
    \caption{\revise{Some examples from Cityscapes dataset \cite{cordts2016cityscapes}. 
    \textbf{From top to bottom:} color codes, original images, 
    ground truth, CASENet \cite{yu2017casenet}, 
    SEAL \cite{yu2018simultaneous}, STEAL \cite{acuna2019devil}, 
    DFF \cite{hu2019dynamic}, our DDS-R and DDS-U. 
    We follow the color coding protocol in \cite{yu2018simultaneous}.
    The produced edges of DDS are smoother and clearer.}}
    \label{fig:city_samples}
\end{figure*}

Yu \etal \shortcite{yu2018simultaneous} discovered that some of the 
original SBD labels are a little noisy, so they re-annotated 1059 
images from the test set to form a new test set. 
We compare our method with CASENet \cite{yu2017casenet}, 
SEAL \cite{yu2018simultaneous}, STEAL \cite{acuna2019devil},
Gated-SCNN \cite{takikawa2019gated}, and DFF \cite{hu2019dynamic}
on this new dataset. 
The results are shown in \tabref{tab:eval_reanno_sbd}.
DDS can improve the performance for both CASENet and DFF 
in terms of all evaluation metrics.
\revise{Specifically, the ODS F-measures of DDS-U is 3.0\% 
and 2.7\% higher than recent SEAL \cite{yu2018simultaneous} 
in terms of the ``Thin'' and ``Raw'' metrics 
in \cite{yu2018simultaneous}, respectively. 
Note that SEAL retrains CASENet with a new training strategy: 
\ie, simultaneous alignment and learning. 
With the same training strategy, DDS-R obtains a 4.3\% and 8.0\% 
higher ODS F-measure than CASENet in terms of the ``Thin'' 
and ``Raw'' metrics in \cite{yu2018simultaneous}, respectively.}

To better visualize the edge prediction results, an example 
is shown in \figref{fig:sbd_samples}.
We also show the normalized images of side activation 
in \figref{fig:sbd_activation}.
All activations are obtained before \textit{sigmoid} 
non-linearization.
For a simple arrangement of figures, we do not display 
Side-4 activation of DDS-R. 
From Side-1 to Side-3, one can see that the feature maps 
of DDS-R are significantly clearer than those of DSN and CASENet. 
Clear category-agnostic edges can be found with DDS-R, while DSN 
and CASENet suffer from noisy activation. 
For example, in CASENet, without imposing deep supervision on Side-1 
$\sim$ Side-3, edge activation can barely be found.
For category classification activation, DDS-R can separate horse 
and person clearly, while DSN and CASENet can not.
Therefore, the \textit{information converters} also help to 
better optimize Side-5 for category-specific classification.
This further verifies the feasibility of the proposed DDS architecture.

More qualitative examples are displayed in \figref{fig:sbd_samps}. 
DDS-R/DDS-U can produce clearer and smother edges than other detectors.
In the second column, it is interesting to note that most 
detectors can recognize the boundaries of the objects with 
missing annotations, \ie, the obscured dining table and human arm. 
In the third column, DDS-R/DDS-U can generate strong responses 
at the boundaries of the small cat, while all other detectors only 
have weak or noisy responses.
This demonstrates that DDS is more robust for detecting small objects.
We also find that DDS-U and SEAL can generate thinner edges, 
suggesting that training with regular unweighted \textit{sigmoid} 
cross entropy loss and refined ground-truth edges is helpful for 
accurately locating thin boundaries.

\subsection{Evaluation on Cityscapes}
The Cityscapes dataset \cite{cordts2016cityscapes} is more 
challenging than SBD \cite{hariharan2011semantic}.
The images in Cityscapes are captured in more complicated scenes, 
usually in urban street scenes in different cities.
There are more objects, especially overlapping objects, 
in each image.
\revise{Hence, we also adopt Cityscapes for evaluating semantic edge detectors
using the ``Thin'' and ``Raw'' metrics in \cite{yu2018simultaneous}.}
%
%
\revise{We compare DDS with CASENet \cite{yu2017casenet},
SEAL \cite{yu2018simultaneous}, STEAL \cite{acuna2019devil}, 
DFF \cite{hu2019dynamic}, and Gated-SCNN \cite{takikawa2019gated}.}

\revise{The evaluation results are reported in \tabref{tab:eval_cityscapes}.
Both DDS-R and DDS-U significantly outperform other methods
in terms of both ``Thin'' and ``Raw'' metrics.
With the same loss function, \revise{the ODS} F-measure of DDS-R 
is 2.8\% higher than CASENet in terms of the ``Thin'' metric 
in \cite{yu2018simultaneous}, and DDS-U is 4.7\% higher than SEAL correspondingly.
STEAL and Gated-SCNN achieve similar performance, and both are much
worse than DDS-R and DDS-U.
Note that Gated-SCNN is the \sArt semantic segmentation model, 
suggesting that it is necessary to study semantic edge detection
rather than directly applying existing related techniques.
Some qualitative comparisons are shown in \figref{fig:city_samples}.
We can see that DDS-R/DDS-U produces smoother and clearer edges
in various complicated scenarios, which is brought by 
the low-level binary edge supervision of DDS.}

\section{Conclusion}
In this paper, we study the SED problem. 
Previous methods suggest that deep supervision is not necessary 
\cite{yu2017casenet,yu2018simultaneous,hu2019dynamic} for SED. 
Here, we show that this is false and, with proper architecture 
re-design, that the network can be deeply supervised to improve 
detection results. 
The core of our approach is the introduction of the novel 
\textit{information converter}, which plays a central role 
in resolving the distinct supervision targets by successfully 
applying category-aware edges at the top side 
and the category-agnostic edges at bottom sides.
The proposed DDS achieves \sArt performance on the popular 
SBD \cite{hariharan2011semantic} 
and Cityscapes \cite{cordts2016cityscapes} datasets. 
Our idea to leverage deep supervision for training a deep network 
opens up a new path towards putting more emphasis utilizing rich feature 
hierarchies from deep networks for SED as well as other high-level 
tasks such as semantic segmentation \cite{maninis2017convolutional,chen2016semantic}, 
object detection \cite{ferrari2008groups,maninis2017convolutional}, 
and instance segmentation \cite{kirillov2017instancecut,hayder2017boundary}.

\myPara{Future Work.}
Besides category-agnostic edge detection and SED, 
relevant tasks commonly exist in computer vision 
\cite{zamir2018taskonomy}, such as 
segmentation and saliency detection, object detection 
and keypoint detection, edge detection and skeleton extraction.
Building multi-task networks to solve relevant tasks is a good 
way to save computational resources in practical applications 
\cite{hou2018three}.
However, distinct supervision targets usually prevent this goal,
as shown in this paper. 
From this point of view, the proposed DDS provides a new 
perspective to multi-task learning. 
In the future, we plan to leverage the idea of 
\textit{information converter} for more relevant tasks.

\bibliographystyle{ijcv}
{\footnotesize\bibliography{reference}}

\begin{thebibliography}{65}
\providecommand{\natexlab}[1]{#1}
\providecommand{\url}[1]{\texttt{#1}}
\providecommand{\urlprefix}{URL }
\providecommand{\eprint}[2][]{\url{#2}}

\bibitem[{Acuna et~al.(2019)Acuna, Kar, \& Fidler}]{acuna2019devil}
Acuna, D., Kar, A., \& Fidler, S. (2019).
\newblock Devil is in the edges: Learning semantic boundaries from noisy
  annotations.
\newblock In \emph{IEEE Conf. Comput. Vis. Pattern Recog.} (pp. 11075--11083).

\bibitem[{Amer et~al.(2015)Amer, Yousefi, Raich, \&
  Todorovic}]{amer2015monocular}
Amer, M.~R., Yousefi, S., Raich, R., \& Todorovic, S. (2015).
\newblock Monocular extraction of 2.1 d sketch using constrained convex
  optimization.
\newblock \emph{Int. J. Comput. Vis.}, \emph{112}(1), pp. 23--42.

\bibitem[{Arbel{\'a}ez et~al.(2011)Arbel{\'a}ez, Maire, Fowlkes, \&
  Malik}]{arbelaez2011contour}
Arbel{\'a}ez, P., Maire, M., Fowlkes, C., \& Malik, J. (2011).
\newblock Contour detection and hierarchical image segmentation.
\newblock \emph{IEEE Trans. Pattern Anal. Mach. Intell.}, \emph{33}(5), pp.
  898--916.

\bibitem[{Bertasius et~al.(2015{\natexlab{a}})Bertasius, Shi, \&
  Torresani}]{bertasius2015deepedge}
Bertasius, G., Shi, J., \& Torresani, L. (2015{\natexlab{a}}).
\newblock {DeepEdge}: A multi-scale bifurcated deep network for top-down
  contour detection.
\newblock In \emph{IEEE Conf. Comput. Vis. Pattern Recog.} (pp. 4380--4389).

\bibitem[{Bertasius et~al.(2015{\natexlab{b}})Bertasius, Shi, \&
  Torresani}]{bertasius2015high}
Bertasius, G., Shi, J., \& Torresani, L. (2015{\natexlab{b}}).
\newblock High-for-low and low-for-high: Efficient boundary detection from deep
  object features and its applications to high-level vision.
\newblock In \emph{Int. Conf. Comput. Vis.} (pp. 504--512).

\bibitem[{Bertasius et~al.(2016)Bertasius, Shi, \&
  Torresani}]{bertasius2016semantic}
Bertasius, G., Shi, J., \& Torresani, L. (2016).
\newblock Semantic segmentation with boundary neural fields.
\newblock In \emph{IEEE Conf. Comput. Vis. Pattern Recog.} (pp. 3602--3610).

\bibitem[{Bian et~al.(2021)Bian, Zhan, Wang, Li, Zhang, Shen
  et~al.}]{bian2021depth}
Bian, J.-W., Zhan, H., Wang, N., Li, Z., Zhang, L., Shen, C., et~al. (2021).
\newblock Unsupervised scale-consistent depth learning from video.
\newblock \emph{Int. J. Comput. Vis.}, \emph{129}, pp. 2548--2564.

\bibitem[{Canny(1986)}]{canny1986computational}
Canny, J. (1986).
\newblock A computational approach to edge detection.
\newblock \emph{IEEE Trans. Pattern Anal. Mach. Intell.}, \emph{8}(6), pp.
  679--698.

\bibitem[{Chan et~al.(2015)Chan, Jia, Gao, Lu, Zeng, \& Ma}]{chan2015pcanet}
Chan, T.-H., Jia, K., Gao, S., Lu, J., Zeng, Z., \& Ma, Y. (2015).
\newblock {PCANet}: A simple deep learning baseline for image classification?
\newblock \emph{IEEE Trans. Image Process.}, \emph{24}(12), pp. 5017--5032.

\bibitem[{Chen et~al.(2016)Chen, Barron, Papandreou, Murphy, \&
  Yuille}]{chen2016semantic}
Chen, L.-C., Barron, J.~T., Papandreou, G., Murphy, K., \& Yuille, A.~L.
  (2016).
\newblock Semantic image segmentation with task-specific edge detection using
  {CNNs} and a discriminatively trained domain transform.
\newblock In \emph{IEEE Conf. Comput. Vis. Pattern Recog.} (pp. 4545--4554).

\bibitem[{Cordts et~al.(2016)Cordts, Omran, Ramos, Rehfeld, Enzweiler, Benenson
  et~al.}]{cordts2016cityscapes}
Cordts, M., Omran, M., Ramos, S., Rehfeld, T., Enzweiler, M., Benenson, R.,
  et~al. (2016).
\newblock The cityscapes dataset for semantic urban scene understanding.
\newblock In \emph{IEEE Conf. Comput. Vis. Pattern Recog.} (pp. 3213--3223).

\bibitem[{Deng et~al.(2018)Deng, Shen, Liu, Wang, \& Liu}]{deng2018learning}
Deng, R., Shen, C., Liu, S., Wang, H., \& Liu, X. (2018).
\newblock Learning to predict crisp boundaries.
\newblock In \emph{Eur. Conf. Comput. Vis.} (pp. 570--586).

\bibitem[{Doll{\'a}r \& Zitnick(2015)}]{dollar2015fast}
Doll{\'a}r, P., \& Zitnick, C.~L. (2015).
\newblock Fast edge detection using structured forests.
\newblock \emph{IEEE Trans. Pattern Anal. Mach. Intell.}, \emph{37}(8), pp.
  1558--1570.

\bibitem[{Ferrari et~al.(2008)Ferrari, Fevrier, Jurie, \&
  Schmid}]{ferrari2008groups}
Ferrari, V., Fevrier, L., Jurie, F., \& Schmid, C. (2008).
\newblock Groups of adjacent contour segments for object detection.
\newblock \emph{IEEE Trans. Pattern Anal. Mach. Intell.}, \emph{30}(1), pp.
  36--51.

\bibitem[{Ferrari et~al.(2010)Ferrari, Jurie, \& Schmid}]{ferrari2010images}
Ferrari, V., Jurie, F., \& Schmid, C. (2010).
\newblock From images to shape models for object detection.
\newblock \emph{Int. J. Comput. Vis.}, \emph{87}(3), pp. 284--303.

\bibitem[{Ganin \& Lempitsky(2014)}]{ganin2014n}
Ganin, Y., \& Lempitsky, V. (2014).
\newblock N$^4$-{Fields}: Neural network nearest neighbor fields for image
  transforms.
\newblock In \emph{Asian Conf. Comput. Vis.} (pp. 536--551).

\bibitem[{Hardie \& Boncelet(1995)}]{hardie1995gradient}
Hardie, R.~C., \& Boncelet, C.~G. (1995).
\newblock Gradient-based edge detection using nonlinear edge enhancing
  prefilters.
\newblock \emph{IEEE Trans. Image Process.}, \emph{4}(11), pp. 1572--1577.

\bibitem[{Hariharan et~al.(2011)Hariharan, Arbel{\'a}ez, Bourdev, Maji, \&
  Malik}]{hariharan2011semantic}
Hariharan, B., Arbel{\'a}ez, P., Bourdev, L., Maji, S., \& Malik, J. (2011).
\newblock Semantic contours from inverse detectors.
\newblock In \emph{Int. Conf. Comput. Vis.} (pp. 991--998).

\bibitem[{Hayder et~al.(2017)Hayder, He, \& Salzmann}]{hayder2017boundary}
Hayder, Z., He, X., \& Salzmann, M. (2017).
\newblock Boundary-aware instance segmentation.
\newblock In \emph{IEEE Conf. Comput. Vis. Pattern Recog.} (pp. 5696--5704).

\bibitem[{He et~al.(2016)He, Zhang, Ren, \& Sun}]{he2016deep}
He, K., Zhang, X., Ren, S., \& Sun, J. (2016).
\newblock Deep residual learning for image recognition.
\newblock In \emph{IEEE Conf. Comput. Vis. Pattern Recog.} (pp. 770--778).

\bibitem[{Henstock \& Chelberg(1996)}]{henstock1996automatic}
Henstock, P.~V., \& Chelberg, D.~M. (1996).
\newblock Automatic gradient threshold determination for edge detection.
\newblock \emph{IEEE Trans. Image Process.}, \emph{5}(5), pp. 784--787.

\bibitem[{Hinton et~al.(2006)Hinton, Osindero, \& Teh}]{hinton2006fast}
Hinton, G.~E., Osindero, S., \& Teh, Y.-W. (2006).
\newblock A fast learning algorithm for deep belief nets.
\newblock \emph{Neural Computation}, \emph{18}(7), pp. 1527--1554.

\bibitem[{Hou et~al.(2019)Hou, Cheng, Hu, Borji, Tu, \& Torr}]{hou2019deeply}
Hou, Q., Cheng, M.-M., Hu, X., Borji, A., Tu, Z., \& Torr, P. (2019).
\newblock Deeply supervised salient object detection with short connections.
\newblock \emph{IEEE Trans. Pattern Anal. Mach. Intell.}, \emph{41}(4), pp.
  815--828.

\bibitem[{Hou et~al.(2018)Hou, Liu, Cheng, Borji, \& Torr}]{hou2018three}
Hou, Q., Liu, J., Cheng, M.-M., Borji, A., \& Torr, P.~H. (2018).
\newblock Three birds one stone: A unified framework for salient object
  segmentation, edge detection and skeleton extraction.
\newblock \emph{arXiv preprint arXiv:1803.09860}.

\bibitem[{Hu et~al.(2018)Hu, Liu, Wang, \& Ren}]{hu2018learning}
Hu, X., Liu, Y., Wang, K., \& Ren, B. (2018).
\newblock Learning hybrid convolutional features for edge detection.
\newblock \emph{Neurocomputing}.

\bibitem[{Hu et~al.(2019)Hu, Chen, Li, \& Feng}]{hu2019dynamic}
Hu, Y., Chen, Y., Li, X., \& Feng, J. (2019).
\newblock Dynamic feature fusion for semantic edge detection.
\newblock In \emph{Int. Joint Conf. Artif. Intell.} (pp. 782--788).

\bibitem[{Ioffe \& Szegedy(2015)}]{ioffe2015batch}
Ioffe, S., \& Szegedy, C. (2015).
\newblock Batch normalization: Accelerating deep network training by reducing
  internal covariate shift.
\newblock In \emph{Int. Conf. Mach. Learn.} (pp. 448--456).

\bibitem[{Jia et~al.(2014)Jia, Shelhamer, Donahue, Karayev, Long, Girshick
  et~al.}]{jia2014caffe}
Jia, Y., Shelhamer, E., Donahue, J., Karayev, S., Long, J., Girshick, R.,
  et~al. (2014).
\newblock Caffe: Convolutional architecture for fast feature embedding.
\newblock In \emph{ACM Int. Conf. Multimedia}. (pp. 675--678).

\bibitem[{Khoreva et~al.(2016)Khoreva, Benenson, Omran, Hein, \&
  Schiele}]{khoreva2016weakly}
Khoreva, A., Benenson, R., Omran, M., Hein, M., \& Schiele, B. (2016).
\newblock Weakly supervised object boundaries.
\newblock In \emph{IEEE Conf. Comput. Vis. Pattern Recog.} (pp. 183--192).

\bibitem[{Kirillov et~al.(2017)Kirillov, Levinkov, Andres, Savchynskyy, \&
  Rother}]{kirillov2017instancecut}
Kirillov, A., Levinkov, E., Andres, B., Savchynskyy, B., \& Rother, C. (2017).
\newblock Instancecut: from edges to instances with multicut.
\newblock In \emph{IEEE Conf. Comput. Vis. Pattern Recog.} (pp. 5008--5017).

\bibitem[{Kokkinos(2016)}]{kokkinos2016pushing}
Kokkinos, I. (2016).
\newblock Pushing the boundaries of boundary detection using deep learning.
\newblock In \emph{Int. Conf. Learn. Represent.} (pp. 1--12).

\bibitem[{Konishi et~al.(2003)Konishi, Yuille, Coughlan, \&
  Zhu}]{konishi2003statistical}
Konishi, S., Yuille, A.~L., Coughlan, J.~M., \& Zhu, S.~C. (2003).
\newblock Statistical edge detection: Learning and evaluating edge cues.
\newblock \emph{IEEE Trans. Pattern Anal. Mach. Intell.}, \emph{25}(1), pp.
  57--74.

\bibitem[{Lee et~al.(2015)Lee, Xie, Gallagher, Zhang, \& Tu}]{lee2015deeply}
Lee, C.-Y., Xie, S., Gallagher, P., Zhang, Z., \& Tu, Z. (2015).
\newblock Deeply-supervised nets.
\newblock In \emph{Artificial Intelligence and Statistics}. (pp. 562--570).

\bibitem[{Lim et~al.(2013)Lim, Zitnick, \& Doll{\'a}r}]{lim2013sketch}
Lim, J.~J., Zitnick, C.~L., \& Doll{\'a}r, P. (2013).
\newblock Sketch tokens: A learned mid-level representation for contour and
  object detection.
\newblock In \emph{IEEE Conf. Comput. Vis. Pattern Recog.} (pp. 3158--3165).

\bibitem[{Lin et~al.(2017)Lin, Doll{\'a}r, Girshick, He, Hariharan, \&
  Belongie}]{lin2017feature}
Lin, T.-Y., Doll{\'a}r, P., Girshick, R., He, K., Hariharan, B., \& Belongie,
  S. (2017).
\newblock Feature pyramid networks for object detection.
\newblock In \emph{IEEE Conf. Comput. Vis. Pattern Recog.} (pp. 2117--2125).

\bibitem[{Lin et~al.(2020)Lin, Goyal, Girshick, He, \&
  Doll{\'a}r}]{lin2020focal}
Lin, T.-Y., Goyal, P., Girshick, R., He, K., \& Doll{\'a}r, P. (2020).
\newblock Focal loss for dense object detection.
\newblock \emph{IEEE Trans. Pattern Anal. Mach. Intell.}, \emph{42}(2), pp.
  318--327.

\bibitem[{Lin et~al.(2014)Lin, Maire, Belongie, Hays, Perona, Ramanan
  et~al.}]{lin2014microsoft}
Lin, T.-Y., Maire, M., Belongie, S., Hays, J., Perona, P., Ramanan, D., et~al.
  (2014).
\newblock Microsoft {COCO}: Common objects in context.
\newblock In \emph{Eur. Conf. Comput. Vis.} (pp. 740--755).

\bibitem[{Liu et~al.(2016)Liu, Anguelov, Erhan, Szegedy, Reed, Fu
  et~al.}]{liu2016ssd}
Liu, W., Anguelov, D., Erhan, D., Szegedy, C., Reed, S., Fu, C.-Y., et~al.
  (2016).
\newblock {SSD}: Single shot multibox detector.
\newblock In \emph{Eur. Conf. Comput. Vis.} (pp. 21--37).

\bibitem[{Liu et~al.(2019)Liu, Cheng, Hu, Bian, Zhang, Bai
  et~al.}]{liu2019richer}
Liu, Y., Cheng, M.-M., Hu, X., Bian, J.-W., Zhang, L., Bai, X., et~al. (2019).
\newblock Richer convolutional features for edge detection.
\newblock \emph{IEEE Trans. Pattern Anal. Mach. Intell.}, \emph{41}(8), pp.
  1939--1946.

\bibitem[{Liu et~al.(2017)Liu, Cheng, Hu, Wang, \& Bai}]{liu2017richer}
Liu, Y., Cheng, M.-M., Hu, X., Wang, K., \& Bai, X. (2017).
\newblock Richer convolutional features for edge detection.
\newblock In \emph{IEEE Conf. Comput. Vis. Pattern Recog.} (pp. 3000--3009).

\bibitem[{Liu et~al.(2018)Liu, Jiang, Petrosyan, Li, Bian, Zhang
  et~al.}]{liu2018deep}
Liu, Y., Jiang, P.-T., Petrosyan, V., Li, S.-J., Bian, J., Zhang, L., et~al.
  (2018).
\newblock {DEL:} deep embedding learning for efficient image segmentation.
\newblock In \emph{Int. Joint Conf. Artif. Intell.} (pp. 864--870).

\bibitem[{Mafi et~al.(2018)Mafi, Rajaei, Cabrerizo, \&
  Adjouadi}]{mafi2018robust}
Mafi, M., Rajaei, H., Cabrerizo, M., \& Adjouadi, M. (2018).
\newblock A robust edge detection approach in the presence of high impulse
  noise intensity through switching adaptive median and fixed weighted mean
  filtering.
\newblock \emph{IEEE Trans. Image Process.}, \emph{27}(11), pp. 5475--5490.

\bibitem[{Maninis et~al.(2017)Maninis, Pont-Tuset, Arbelaez, \&
  Van~Gool}]{maninis2017convolutional}
Maninis, K.-K., Pont-Tuset, J., Arbelaez, P., \& Van~Gool, L. (2017).
\newblock Convolutional oriented boundaries: From image segmentation to
  high-level tasks.
\newblock \emph{IEEE Trans. Pattern Anal. Mach. Intell.}, \emph{40}(4), pp.
  819--833.

\bibitem[{Martin et~al.(2004)Martin, Fowlkes, \& Malik}]{martin2004learning}
Martin, D.~R., Fowlkes, C.~C., \& Malik, J. (2004).
\newblock Learning to detect natural image boundaries using local brightness,
  color, and texture cues.
\newblock \emph{IEEE Trans. Pattern Anal. Mach. Intell.}, \emph{26}(5), pp.
  530--549.

\bibitem[{Nair \& Hinton(2010)}]{nair2010rectified}
Nair, V., \& Hinton, G.~E. (2010).
\newblock Rectified linear units improve restricted {B}oltzmann machines.
\newblock In \emph{Int. Conf. Mach. Learn.} (pp. 807--814).

\bibitem[{Ramalingam et~al.(2010)Ramalingam, Bouaziz, Sturm, \&
  Brand}]{ramalingam2010skyline2gps}
Ramalingam, S., Bouaziz, S., Sturm, P., \& Brand, M. (2010).
\newblock Skyline2gps: Localization in urban canyons using omni-skylines.
\newblock In \emph{IEEE$\backslash$RSJ Int. Conf. Intell. Robot. Syst.} (pp.
  3816--3823).

\bibitem[{Shan et~al.(2014)Shan, Curless, Furukawa, Hernandez, \&
  Seitz}]{shan2014occluding}
Shan, Q., Curless, B., Furukawa, Y., Hernandez, C., \& Seitz, S.~M. (2014).
\newblock Occluding contours for multi-view stereo.
\newblock In \emph{IEEE Conf. Comput. Vis. Pattern Recog.} (pp. 4002--4009).

\bibitem[{Shen et~al.(2015)Shen, Wang, Wang, Bai, \&
  Zhang}]{shen2015deepcontour}
Shen, W., Wang, X., Wang, Y., Bai, X., \& Zhang, Z. (2015).
\newblock {DeepContour}: A deep convolutional feature learned by
  positive-sharing loss for contour detection.
\newblock In \emph{IEEE Conf. Comput. Vis. Pattern Recog.} (pp. 3982--3991).

\bibitem[{Shui \& Wang(2017)}]{shui2017anti}
Shui, P.-L., \& Wang, F.-P. (2017).
\newblock Anti-impulse-noise edge detection via anisotropic morphological
  directional derivatives.
\newblock \emph{IEEE Trans. Image Process.}, \emph{26}(10), pp. 4962--4977.

\bibitem[{Sobel(1970)}]{sobel1970camera}
Sobel, I. (1970).
\newblock Camera models and machine perception.
\newblock \emph{Technical report}, Stanford Univ Calif Dept of Computer
  Science.

\bibitem[{Srivastava et~al.(2014)Srivastava, Hinton, Krizhevsky, Sutskever, \&
  Salakhutdinov}]{srivastava2014dropout}
Srivastava, N., Hinton, G., Krizhevsky, A., Sutskever, I., \& Salakhutdinov, R.
  (2014).
\newblock Dropout: A simple way to prevent neural networks from overfitting.
\newblock \emph{J. Mach. Learn. Res.}, \emph{15}(1), pp. 1929--1958.

\bibitem[{Szegedy et~al.(2015)Szegedy, Liu, Jia, Sermanet, Reed, Anguelov
  et~al.}]{szegedy2015going}
Szegedy, C., Liu, W., Jia, Y., Sermanet, P., Reed, S., Anguelov, D., et~al.
  (2015).
\newblock Going deeper with convolutions.
\newblock In \emph{IEEE Conf. Comput. Vis. Pattern Recog.} (pp. 1--9).

\bibitem[{Takikawa et~al.(2019)Takikawa, Acuna, Jampani, \&
  Fidler}]{takikawa2019gated}
Takikawa, T., Acuna, D., Jampani, V., \& Fidler, S. (2019).
\newblock Gated-{SCNN}: Gated shape {CNN}s for semantic segmentation.
\newblock In \emph{Int. Conf. Comput. Vis.} (pp. 5229--5238).

\bibitem[{Tang et~al.(2017)Tang, Wang, Feng, \& Liu}]{tang2017learning}
Tang, P., Wang, X., Feng, B., \& Liu, W. (2017).
\newblock Learning multi-instance deep discriminative patterns for image
  classification.
\newblock \emph{IEEE Trans. Image Process.}, \emph{26}(7), pp. 3385--3396.

\bibitem[{Trahanias \& Venetsanopoulos(1993)}]{trahanias1993color}
Trahanias, P.~E., \& Venetsanopoulos, A.~N. (1993).
\newblock Color edge detection using vector order statistics.
\newblock \emph{IEEE Trans. Image Process.}, \emph{2}(2), pp. 259--264.

\bibitem[{Wang et~al.(2015)Wang, Ouyang, Wang, \& Lu}]{wang2015visual}
Wang, L., Ouyang, W., Wang, X., \& Lu, H. (2015).
\newblock Visual tracking with fully convolutional networks.
\newblock In \emph{Int. Conf. Comput. Vis.} (pp. 3119--3127).

\bibitem[{Wang et~al.(2019)Wang, Zhao, Li, \& Huang}]{wang2019deep}
Wang, Y., Zhao, X., Li, Y., \& Huang, K. (2019).
\newblock Deep crisp boundaries: From boundaries to higher-level tasks.
\newblock \emph{IEEE Trans. Image Process.}, \emph{28}(3), pp. 1285--1298.

\bibitem[{Xie \& Tu(2015)}]{xie2015holistically}
Xie, S., \& Tu, Z. (2015).
\newblock Holistically-nested edge detection.
\newblock In \emph{Int. Conf. Comput. Vis.} (pp. 1395--1403).

\bibitem[{Xie \& Tu(2017)}]{xie2017holistically}
Xie, S., \& Tu, Z. (2017).
\newblock Holistically-nested edge detection.
\newblock \emph{Int. J. Comput. Vis.}, \emph{125}(1-3), pp. 3--18.

\bibitem[{Yang et~al.(2016)Yang, Price, Cohen, Lee, \& Yang}]{yang2016object}
Yang, J., Price, B., Cohen, S., Lee, H., \& Yang, M.-H. (2016).
\newblock Object contour detection with a fully convolutional encoder-decoder
  network.
\newblock In \emph{IEEE Conf. Comput. Vis. Pattern Recog.} (pp. 193--202).

\bibitem[{Yang et~al.(2017)Yang, Feng, Yang, Zhao, Liu, Guo
  et~al.}]{yang2017deep}
Yang, W., Feng, J., Yang, J., Zhao, F., Liu, J., Guo, Z., et~al. (2017).
\newblock Deep edge guided recurrent residual learning for image
  super-resolution.
\newblock \emph{IEEE Trans. Image Process.}, \emph{26}(12), pp. 5895--5907.

\bibitem[{Yu \& Koltun(2016)}]{yu2015multi}
Yu, F., \& Koltun, V. (2016).
\newblock Multi-scale context aggregation by dilated convolutions.
\newblock In \emph{Int. Conf. Learn. Represent.} (pp. 1--13).

\bibitem[{Yu et~al.(2017)Yu, Feng, Liu, \& Ramalingam}]{yu2017casenet}
Yu, Z., Feng, C., Liu, M.-Y., \& Ramalingam, S. (2017).
\newblock {CASENet}: Deep category-aware semantic edge detection.
\newblock In \emph{IEEE Conf. Comput. Vis. Pattern Recog.} (pp. 5964--5973).

\bibitem[{Yu et~al.(2018)Yu, Liu, Zou, Feng, Ramalingam, Kumar
  et~al.}]{yu2018simultaneous}
Yu, Z., Liu, W., Zou, Y., Feng, C., Ramalingam, S., Kumar, B., et~al. (2018).
\newblock Simultaneous edge alignment and learning.
\newblock In \emph{Eur. Conf. Comput. Vis.} (pp. 400--417).

\bibitem[{Zamir et~al.(2018)Zamir, Sax, Shen, Guibas, Malik, \&
  Savarese}]{zamir2018taskonomy}
Zamir, A.~R., Sax, A., Shen, W., Guibas, L., Malik, J., \& Savarese, S. (2018).
\newblock Taskonomy: Disentangling task transfer learning.
\newblock In \emph{IEEE Conf. Comput. Vis. Pattern Recog.} (pp. 3712--3722).

\end{thebibliography}

\end{document}